\RequirePackage[2025-02-01]{latexrelease}
\documentclass[11pt]{article}

\usepackage[final]{acl}

\usepackage{times}
\usepackage{latexsym}
\usepackage{booktabs}
\usepackage{placeins}
\usepackage{amsmath}

\usepackage[T1]{fontenc}

\usepackage[utf8]{inputenc}
\usepackage{csquotes}
\usepackage{microtype}

\usepackage{inconsolata}
\usepackage{subcaption}

\usepackage{graphicx}
\usepackage{listings}
\usepackage{xcolor}

\title{When Does Continual Pre-Training Help? An Empirical Study of Decoupled Knowledge and Generation Trajectories in Clinical LLMs}

\author{
  \textbf{Niclas Doll}\textsuperscript{\mdseries 1,2},
  \textbf{Jasper Schulze Buschhoff}\textsuperscript{\mdseries 1},
  \textbf{Shalaka Satheesh}\textsuperscript{\mdseries 1},
  \\
  \textbf{H\'ector Allende-Cid}\textsuperscript{\mdseries 1,2},
  \textbf{Katrin Klug}\textsuperscript{\mdseries 1}
  \\
  \\
  \textsuperscript{\mdseries 1}Fraunhofer IAIS,
  \textsuperscript{\mdseries 2}Lamarr Institute
  \\
  \{niclas.doll, johann.jasper.schulze.buschoff, shalaka.satheesh\}@iais.fraunhofer.de
}

\begin{document}


\maketitle

\begin{abstract}
Continual pre-training (\textit{CPT}) is a standard route for domain adaptation, yet existing studies typically track pre-training loss or multiple-choice question answering (\textit{MCQA}) accuracy alone, offering only a narrow view of how adaptation unfolds. We jointly track factual knowledge via \textit{MCQA} and open-ended behavior via pairwise preference rankings across $18$ checkpoints for each of eight instruction-tuned \textit{Qwen} models ($0.5B$ to $8B$), adapted to German medicine with \textit{CPT} and \textit{SLERP} merging. The two axes converge at consistently different rates, with preference rankings reaching $90\%$ of their total gain after a median of $1.7B$ tokens versus $3.5B$ for \textit{MCQA}. A judge-free token-distribution measure reaches this threshold at $1.9B$ tokens. Per-aspect scores indicate that these early preference gains largely reflect surface alignment, meaning models can sound domain-appropriate before their factual knowledge catches up. Larger adapted models close more of the gap to unadapted next-tier generalists, whereas preference gain per FLOP falls by about an order of magnitude across our size range. We therefore argue that judge-based preference scores, which can plateau while \textit{MCQA} knowledge is still improving, should not be used alone to infer adaptation convergence or domain readiness.
\end{abstract}

\begin{figure*}[th]
    \centering

    \begin{subfigure}{\linewidth}
        \centering
        \begin{subfigure}{0.49\linewidth}
            \centering
            \includegraphics[width=\linewidth]{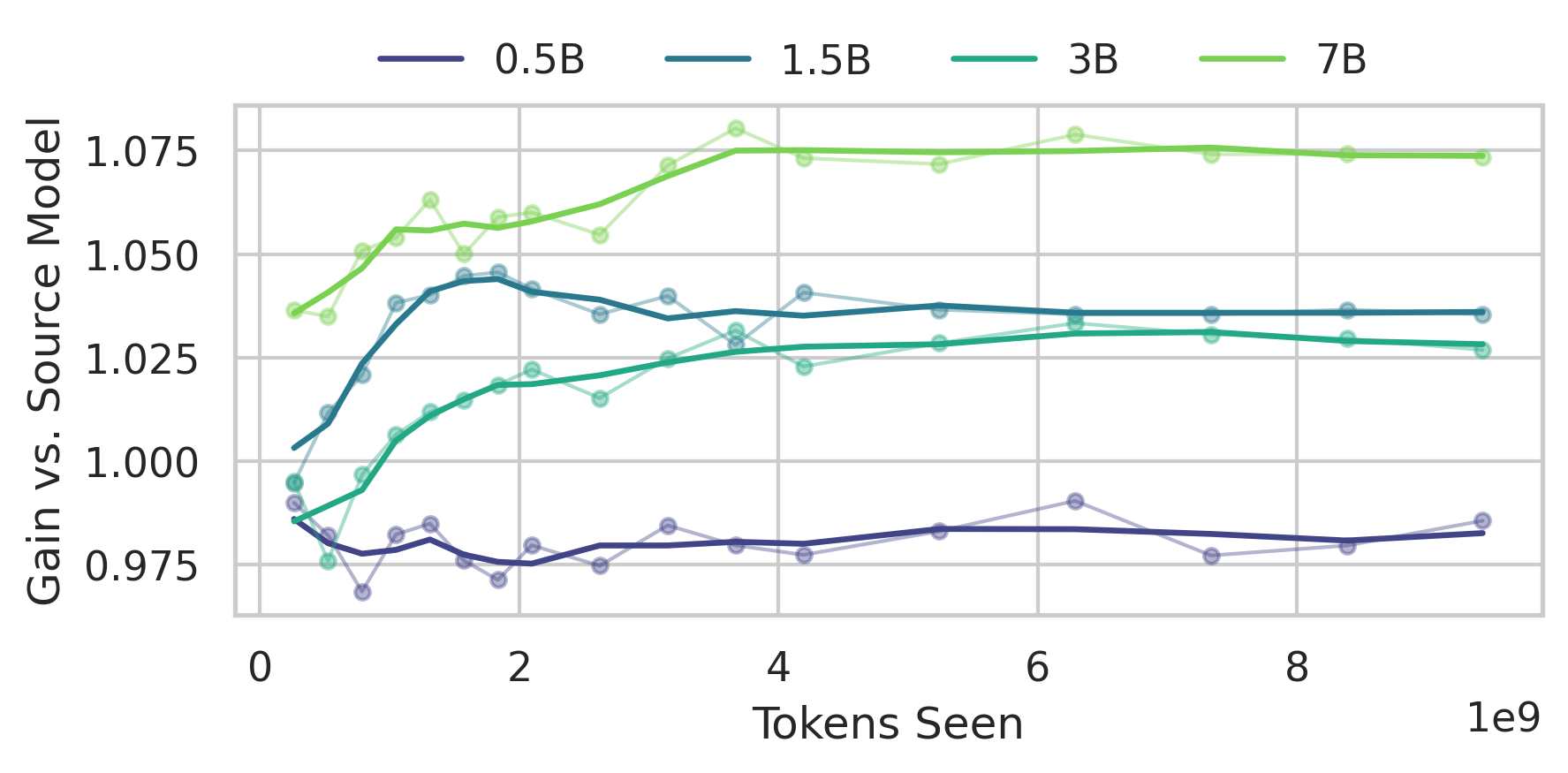}
            \caption{\textit{MCQA}: against source model}
            \label{fig:qwen2.5-mcqa-vs-source-tokens}
        \end{subfigure}
        \hfill
        \begin{subfigure}{0.49\linewidth}
            \centering
            \includegraphics[width=\linewidth]{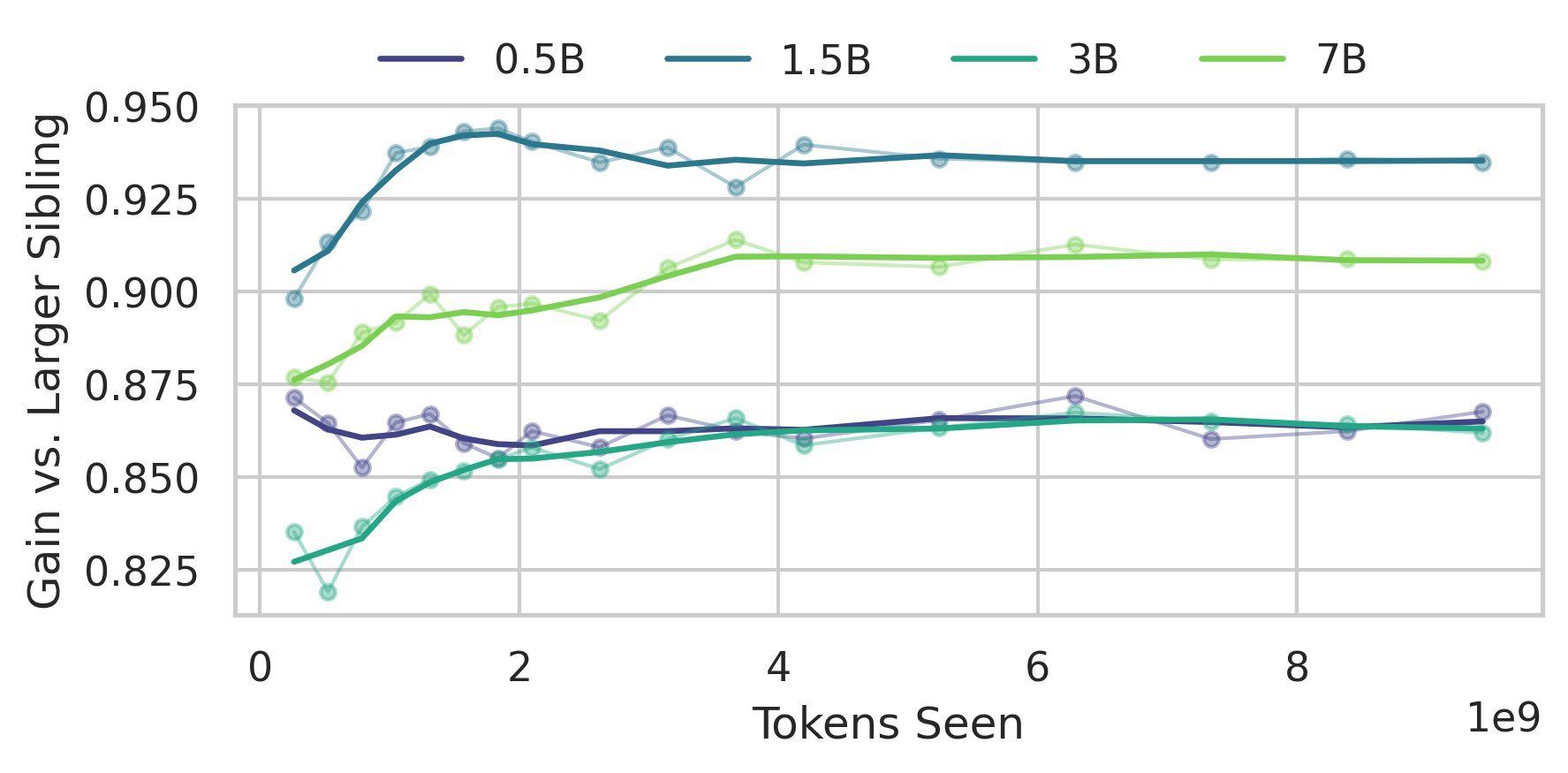}
            \caption{\textit{MCQA}: against next larger model}
            \label{fig:qwen2.5-mcqa-vs-larger-tokens}
        \end{subfigure}
    \end{subfigure}

    \begin{subfigure}{\linewidth}
        \centering
        \begin{subfigure}{0.49\linewidth}
            \centering
            \includegraphics[width=\linewidth]{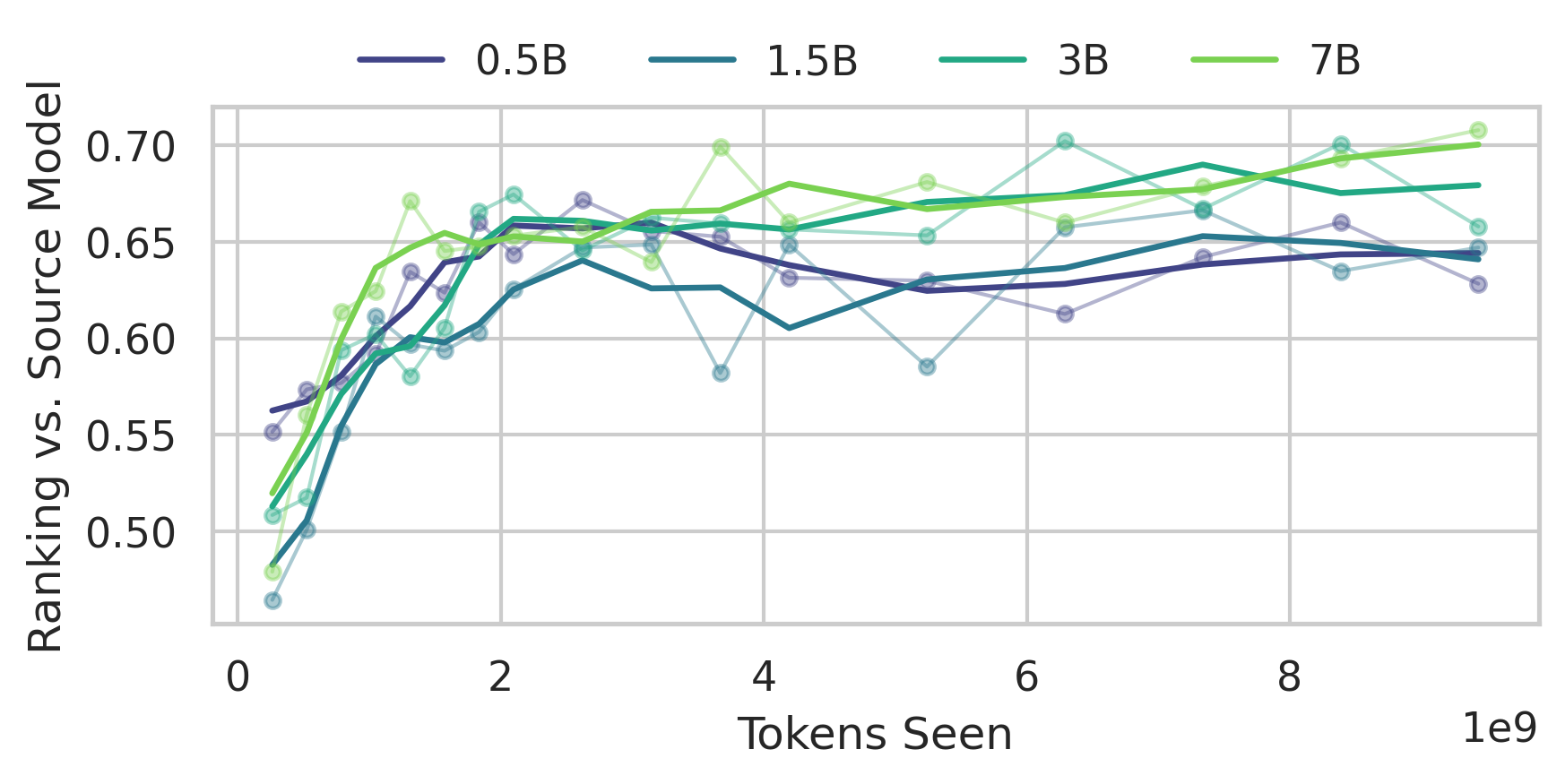}
            \caption{Generative rankings: against source model}
            \label{fig:qwen2.5-ranking-vs-source-tokens}
        \end{subfigure}
        \hfill
        \begin{subfigure}{0.49\linewidth}
            \centering
            \includegraphics[width=\linewidth]{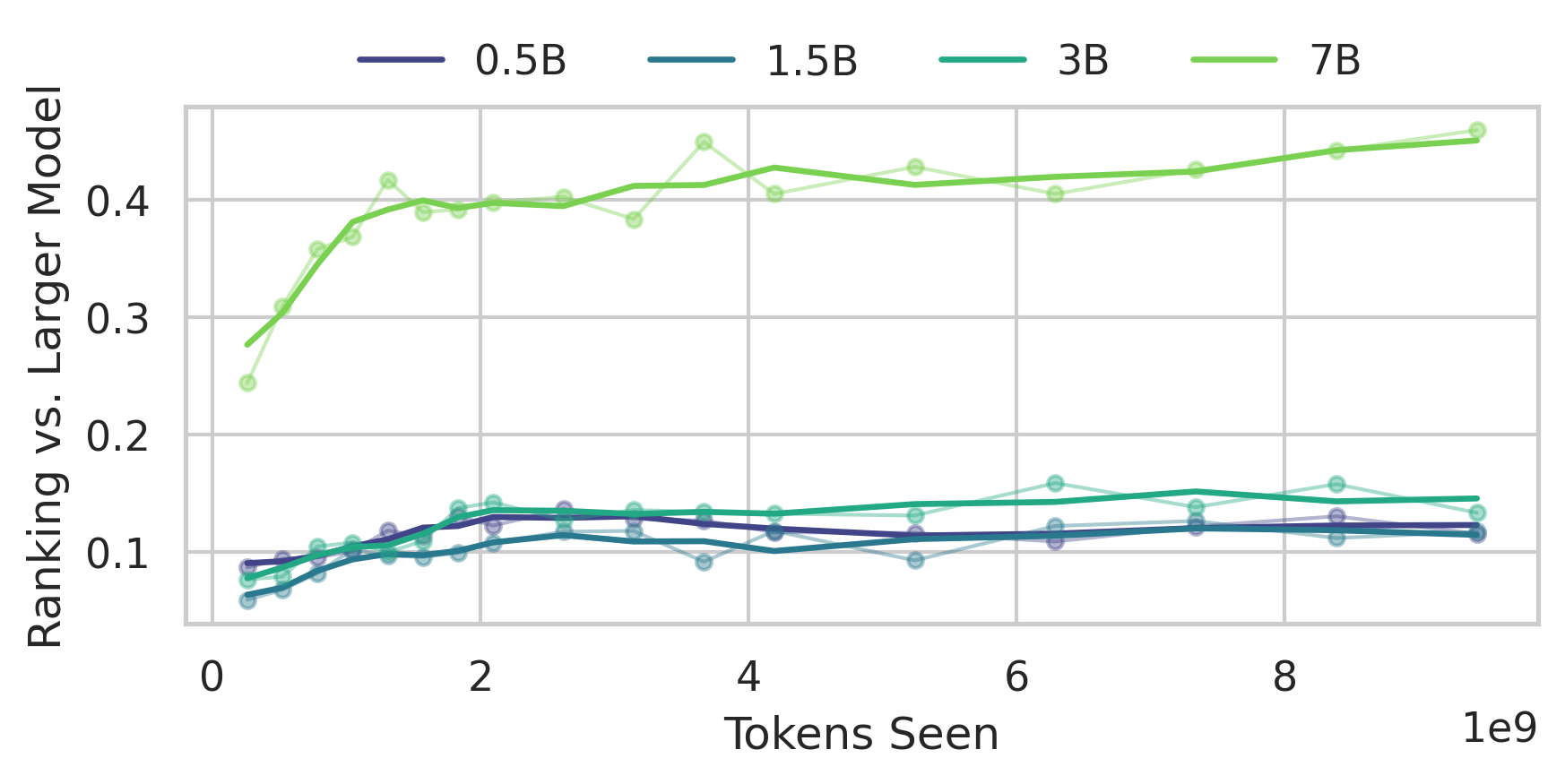}
            \caption{Generative rankings: against next larger model}
            \label{fig:qwen2.5-ranking-vs-larger-tokens}
        \end{subfigure}
    \end{subfigure}

    \caption{
    Scaling trajectories for the \textit{Qwen2.5} family across domain-specific token exposure. The top row tracks factual knowledge (\textit{MCQA}) and the bottom row tracks open-ended response quality via Davidson preference rankings. These plots illustrate adaptation dynamics relative to the original source models and relative parity achieved against the next-tier general-purpose variants (e.g., $7B$ vs. $14B$).
    }
    \label{fig:qwen2.5-results-combined-tokens}
\end{figure*}

\section{Introduction}

Large Language Models (LLMs) have rapidly become central to innovation in data-driven fields, with healthcare representing a particularly dynamic and challenging area for their deployment \cite{naveed2025comprehensive, thirunavukarasu2023large}. These models are increasingly used to interpret complex clinical information and support a range of medical tasks, offering the potential to enhance both the efficiency and quality of care. However, translating the broad capabilities of LLMs into reliable, domain-specific performance is challenging. Specialized domains, such as medicine, demand not only expertise of technical terminology and structured documentation, but also reliable adaptation to local linguistic and operational contexts \cite{meyer2024}. Furthermore, the need for secure, efficient, and compliant AI solutions makes it important to develop small, efficient models that can be tailored to diverse regulatory and hardware environments \cite{belcak2025}.

To address these limitations, domain adaptation through continual pre-training (\textit{CPT}) has emerged as a standard methodology for internalizing specialized knowledge \cite{shi2025continual}. Yet, current research in this area remains limited by two critical gaps. First, many studies evaluate domain adaptation as a static, ``before-and-after'' comparison of the source and the final adapted model \cite{chen2025towards}, overlooking the evolution of model performance as a function of computational investment. The few studies that do track \textit{CPT} across checkpoints or model scales \cite{christophe2024beyond, guo2025efficient, gu-2024-cmr, junior-2025-interplay} measure progress through loss- or knowledge-based metrics alone. This points to a second, broader evaluation gap. The majority of domain-adaptation literature relies on objective Multiple Choice Question Answering (\textit{MCQA}) benchmarks to measure success \cite{christophe2024beyond, guo2025efficient}. This narrow focus overlooks the quality of open-ended generation, as strong performance on \textit{MCQA} benchmarks does not necessarily translate into clinically reliable open‑ended generation~\cite{singhal2025toward, li-etal-2024-multiple, gong2025knowledge}.

This evaluation gap is not specific to domain-adaptation research, it is shared by the broader scaling-law literature. Foundational compute-optimal scaling analyses such as \citet{hoffmann2022training} optimize pre-training loss, while subsequent work on the interplay between scale and specialization \cite{junior-2025-interplay} relies on \textit{MCQA} perplexity. Neither evaluates open-ended generative quality, despite evidence that this axis is not interchangeable with loss-based or knowledge-based metrics~\cite{holtzman2020curious, lyu-etal-2024-beyond, wang-etal-2024-answer, zheng2023judging}. These analyses provide the foundation for our compute-based view of adaptation, but because they follow a single evaluation axis, they cannot show whether different capabilities evolve at different rates, a question that is both scientifically open and clinically consequential.

This paper bridges these gaps by providing a systematic, comparative analysis of two state-of-the-art model families, \textit{Qwen2.5} and \textit{Qwen3}, across parameter scales ranging from $0.5B$ to $8B$. By tracking the adaptation process across billions of tokens and utilizing weight-merging techniques to preserve conversational alignment, we move beyond factual recall to investigate the compute-performance frontier of open-ended generation quality. Our work provides a dual-metric analysis, in which we utilize traditional \textit{MCQA} benchmarks to track knowledge acquisition and an open-ended ``LLM-as-a-judge'' framework to approximate real-world usability.

Our findings offer three primary contributions:

\begin{enumerate}
    \item \textbf{Decoupled adaptation dynamics.} We reveal that factual recall and surface alignment follow decoupled evolutionary paths during domain adaptation, with surface alignment converging rapidly while factual knowledge acquisition progresses more gradually.
    \item \textbf{Scale-dependent gap-closing.} We demonstrate that domain-specific compute progressively narrows the performance gap to larger generalist counterparts, with the degree of gap-closing increasing as a function of the adapted model's capacity.
    \item \textbf{Compute-efficiency trade-off.} We show that while larger models achieve higher absolute performance ceilings, smaller models obtain a large share of the gain at a fraction of the computational cost. Preference gain per FLOP falls by about an order of magnitude across our size range, a pattern that resembles the compute-optimal trade-offs \citet{hoffmann2022training} describe for pre-training from scratch.
\end{enumerate}

The remainder of this paper is organized as follows. Section \ref{sec:related} reviews related work in domain adaptation. Sections \ref{sec:method} and \ref{sec:setup} describe the methodology, adaptation pipeline, and experimental configuration. Our findings are reported in Section \ref{sec:findings} and discussed in Section \ref{sec:discussion}. Finally, Section \ref{sec:conclusion} provides concluding remarks and directions for future research.

\section{Related Work}
\label{sec:related}

Our study draws on three lines of prior work. We first review scaling analyses that relate model size and compute to performance, then approaches for adapting LLMs to specialized domains, and finally model merging, which our pipeline uses to retain instruction-following abilities.

\subsection{Model Scale and Specialization}

The relationship between model scale, training compute, and downstream performance has been formalized through a series of scaling-law studies. \citet{kaplan2020scaling} established that pre-training loss follows predictable power-law relationships with model size, dataset size, and compute. \citet{hoffmann2022training} refined this picture by deriving the compute-optimal allocation between parameters and training tokens, establishing that smaller models trained on more tokens are more compute-efficient than larger under-trained models. These analyses, however, operate exclusively on a single evaluation axis, pre-training loss, and characterize only the final model rather than the trajectory through which it acquires capabilities. \citet{junior-2025-interplay} extended this line of work to domain specialization by continually pre-training \textit{Qwen2.5} base models from $1.5B$ to $14B$ parameters on Portuguese legal, medical, and accounting data. Contrary to their initial hypothesis that larger models would benefit less from specialization, they found that specialized models outperform their general counterparts at every size under a fixed compute budget, and that this compute advantage grows with model size. Their evaluation, however, is restricted to \textit{MCQA} perplexity, which inherits the single-axis limitation of the broader scaling literature.

Our work differs from these analyses by evaluating two complementary axes (\textit{MCQA} and open-ended preference rankings) over $18$ checkpoints per run, in a regime that combines continual pre-training with \textit{SLERP} merging on already instruction-tuned models rather than from-scratch pre-training. The dual-axis trajectory analysis exposes the decoupled-dynamics finding that constitutes our primary contribution, and the compute-efficiency result (Section~\ref{sec:compute_efficiency}) provides empirical evidence that Chinchilla-style compute-optimal behavior extends into the continual-pre-training regime.

\subsection{Domain Adaptation of LLMs}

Continual pre-training (CPT) has emerged as a common approach for adapting general-purpose LLMs to specialized domains. Its effectiveness depends heavily on corpus quality and alignment with the model’s prior knowledge. Additional pre-training does not always help, and can even degrade perplexity on the target domain \citep{oncel2024}, while combining CPT with model merging can recover general capabilities while retaining the newly acquired domain or language knowledge \citep{siriwardhana2024, alexandrov2024mitigating}. Domain-specific pre-training has also been applied to the legal domain and to Traditional Chinese Medicine \citep{niyogi2024, Yang2024b}. Complementary work emphasizes balancing general and domain-specific knowledge. \citet{colombo-2024-saullm7b} mix a small proportion of general data into their domain corpus to reduce forgetting, \citet{ibrahim2024simple} show empirically that such replay preserves earlier capabilities during continual pre-training, and \citet{gu-2024-cmr} formalize this trade-off via the Critical Mixture Ratio. Together, these studies demonstrate that domain specialization is not solely determined by model scale, but also by careful data selection, training design, and compute allocation.

A parallel line of work approaches medical specialization through supervised fine-tuning on curated reasoning data. \citet{wu2025medreason} use a medical knowledge graph to generate and verify reasoning chains, a sample of which was assessed by medical experts, and \citet{zhuang2025efficient} show that principled data selection allows medical \textit{LLMs} to match full-dataset performance with $1\%$ of the data. These post-training approaches operate on a different lever than \textit{CPT}, but their shared argument that principled choices outperform brute-force scaling parallels our model-size finding.

\subsection{Merging Large Language Models}

Resource-efficient adaptation strategies, such as model merging, offer a practical alternative to full-scale fine-tuning. Surveys and recent experiments \citep{yang2024modelmergingllmsmllms, nobari2025} report that model merging can yield substantial performance gains with reduced computational cost. Larger models generally facilitate more successful merging, while different merging strategies perform comparably at scale \citep{yadav2024}. In the medical domain, merging a continually pre-trained model back into its instruction-tuned source with Spherical Linear Interpolation (SLERP) \citep{shoemake-1985-slerp} has been shown to achieve strong empirical results \citep{labrak2024biomistralcollectionopensourcepretrained}, and toolkits such as \textit{mergekit} \citep{goddard-etal-2024-arcees} make these techniques practical for creating high-performing, compute-efficient LLMs.

\begin{figure*}[ht]
    \centering

    \begin{subfigure}{\linewidth}
        \centering
        \begin{subfigure}{0.49\linewidth}
            \centering
            \includegraphics[width=\linewidth]{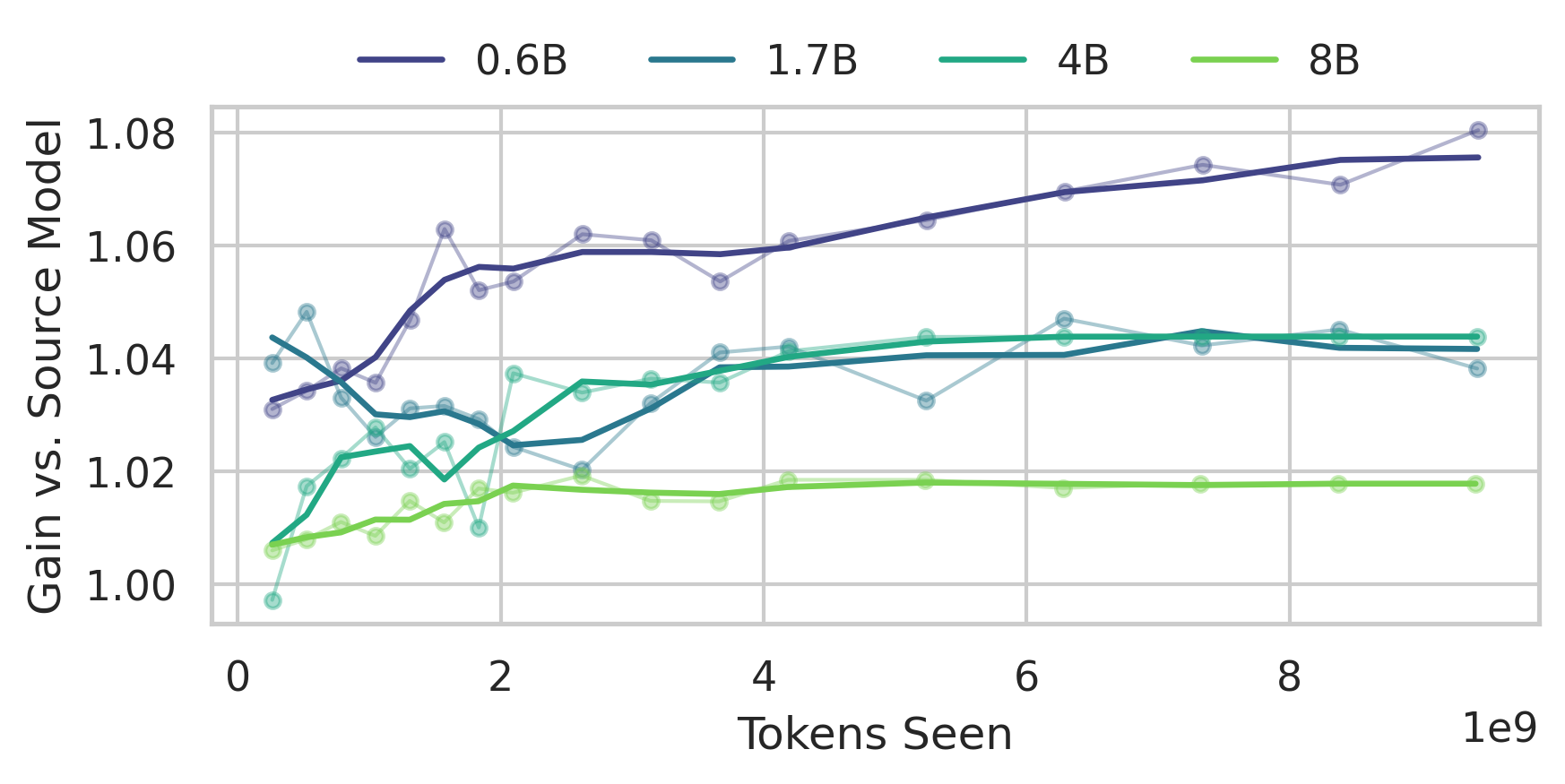}
            \caption{\textit{MCQA}: against source model}
            \label{fig:qwen3-mcqa-vs-source-tokens}
        \end{subfigure}
        \hfill
        \begin{subfigure}{0.49\linewidth}
            \centering
            \includegraphics[width=\linewidth]{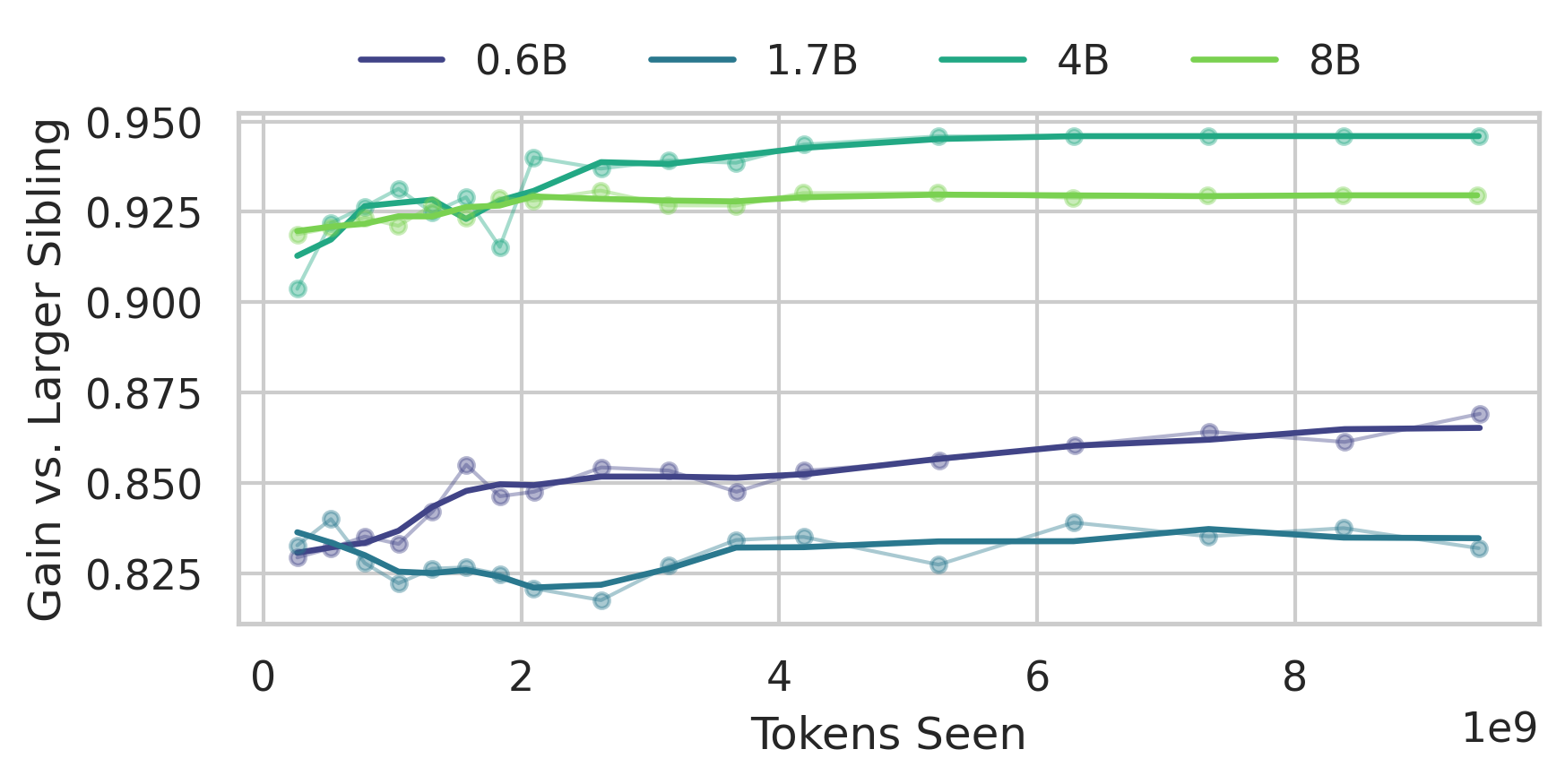}
            \caption{\textit{MCQA}: against next larger model}
            \label{fig:qwen3-mcqa-vs-larger-tokens}
        \end{subfigure}
    \end{subfigure}

    \begin{subfigure}{\linewidth}
        \centering
        \begin{subfigure}{0.49\linewidth}
            \centering
            \includegraphics[width=\linewidth]{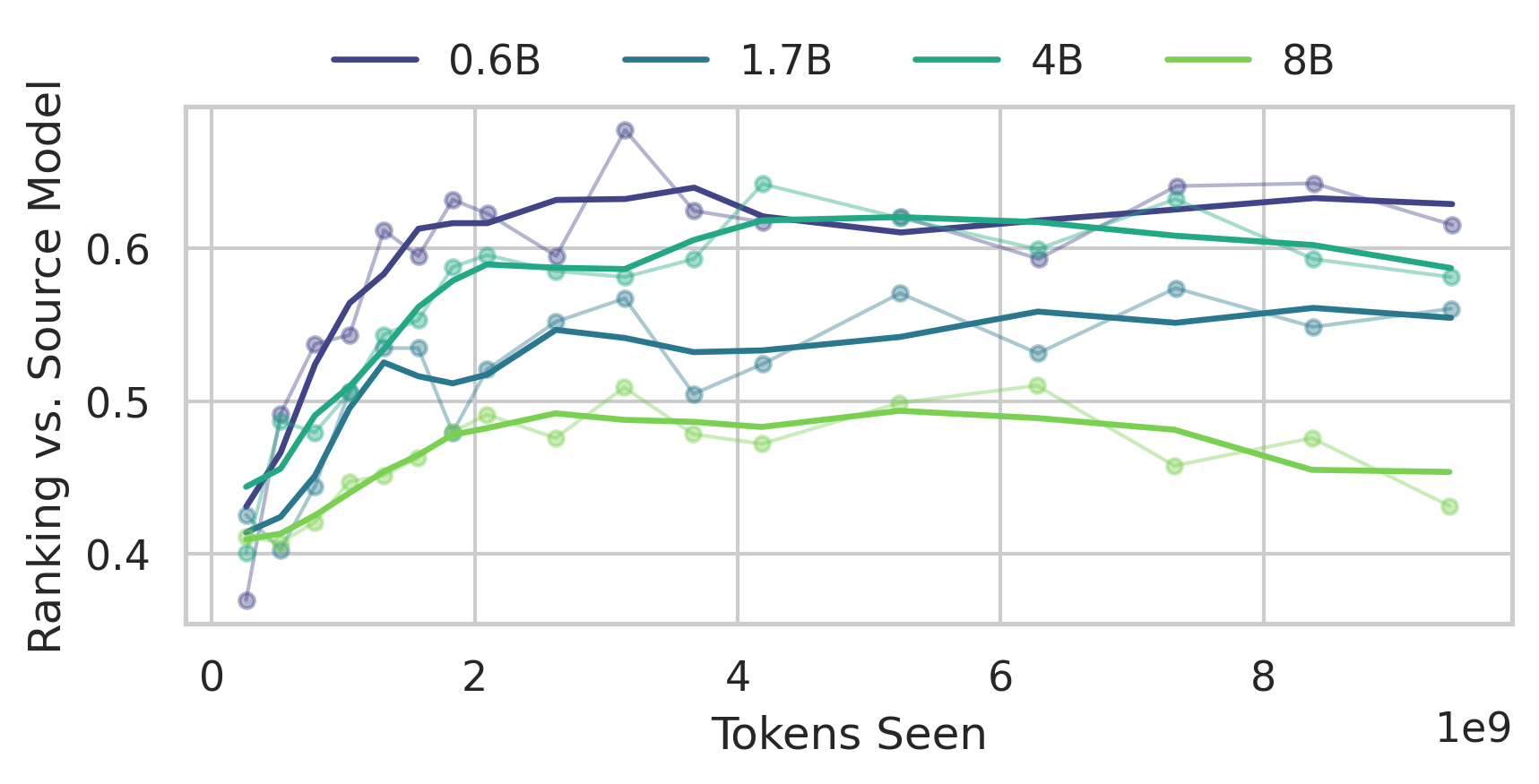}
            \caption{Generative rankings: against source model}
            \label{fig:qwen3-ranking-vs-source-tokens}
        \end{subfigure}
        \hfill
        \begin{subfigure}{0.49\linewidth}
            \centering
            \includegraphics[width=\linewidth]{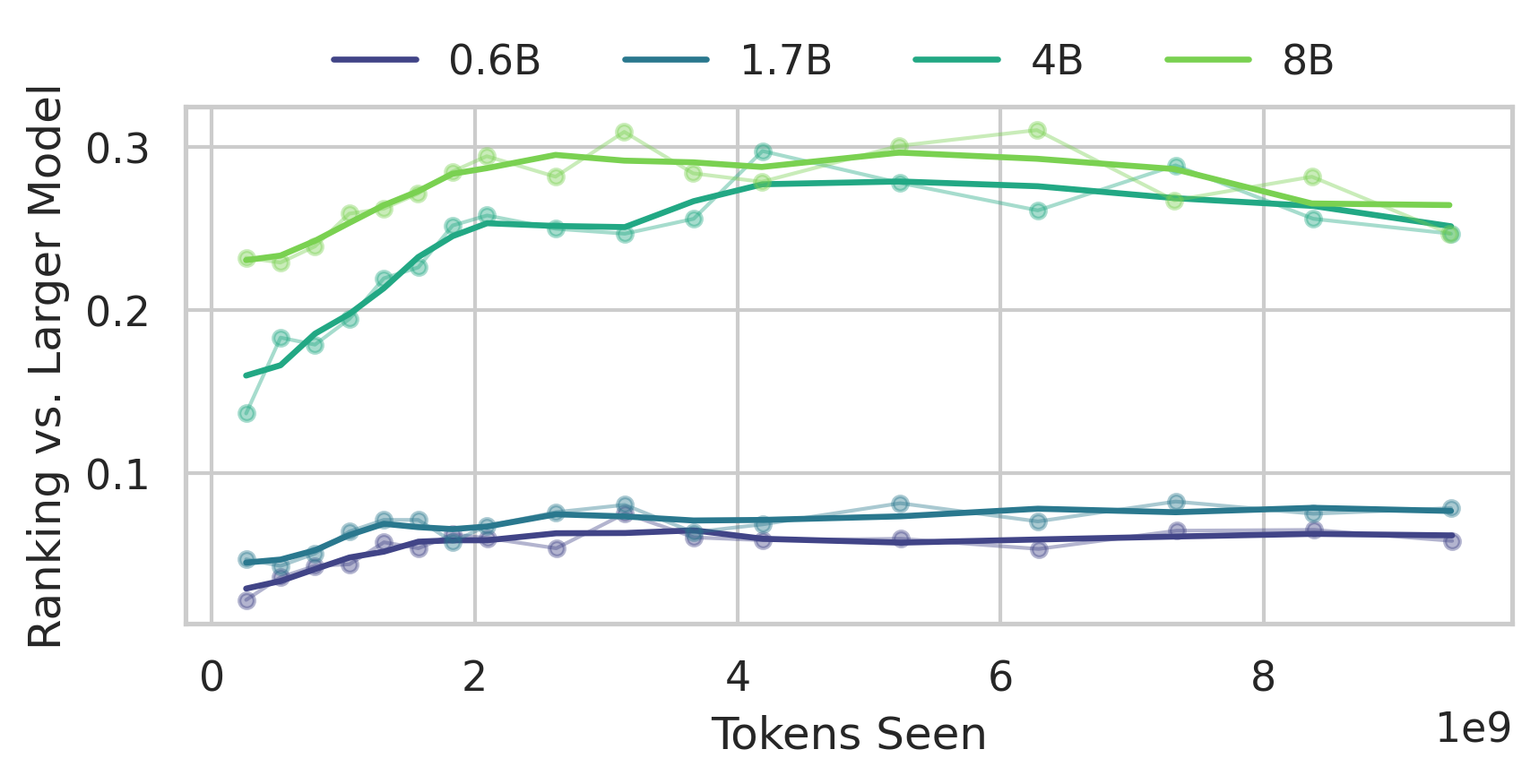}
            \caption{Generative rankings: against next larger model}
            \label{fig:qwen3-ranking-vs-larger-tokens}
        \end{subfigure}
    \end{subfigure}

    \caption{
    Scaling trajectories for the \textit{Qwen3} family across domain-specific token exposure, analogous to Figure~\ref{fig:qwen2.5-results-combined-tokens}. The top row tracks factual knowledge while the bottom row tracks preference rankings, comparing adapted variants against their original source models and the next-tier general-purpose baselines (e.g., $8B$ vs. $14B$).
    }
    \label{fig:qwen3-results-combined-tokens}
\end{figure*}

\section{Methodology}
\label{sec:method}

This section describes the adaptation pipeline applied to every model and the two evaluation axes along which we track each checkpoint. The concrete corpus, models, and benchmarks follow in Section~\ref{sec:setup}.

\subsection{Domain Adaptation Pipeline}
\label{sec:method:pipeline}

We follow the multi-stage pipeline of \citet{Lu2025}, starting from an instruction-tuned base model. The first phase applies \textit{CPT} on a curated, domain-specific corpus with a standard causal language modeling objective. To preserve instruction-following capabilities, we then merge the continually pre-trained model with its original instruction-tuned counterpart using Spherical Linear Interpolation (SLERP) \cite{Lu2025, labrak2024biomistralcollectionopensourcepretrained}, which accounts for the high-dimensional spherical geometry of the weight space. The layer-specific interpolation configuration is given in Appendix~\ref{sec:merge-config}.

\subsection{Evaluation}
\label{sec:method:eval}

We evaluate checkpoints along two complementary axes, objective knowledge via knowledge-based benchmarks, and open-ended response quality via preference rankings.

\subsubsection{Knowledge-Based Benchmarks}
\label{sec:method:eval:mcqa}

Objective performance is measured through \textit{MCQA} using standardized benchmarks relevant to the target domain. These benchmarks provide a stable and reproducible measure of factual knowledge independent of generative formatting or linguistic style. We report one-shot accuracy across all saved checkpoints. To isolate the impact of continual pre-training, we report the relative performance gain $G_{rel}(C_i) = Acc(C_i) / Acc(M_{bl})$, where $C_i$ is an adapted checkpoint and $M_{bl}$ is the baseline, typically the original instruction-tuned source model.

\subsubsection{Preference Ranking}
\label{sec:method:eval:ranking}

We evaluate open-ended response quality with an \textit{LLM-as-a-Judge} approach \cite{zheng2023judging, dubois2025lengthcontrolledalpacaevalsimpleway}. For each prompt, an adapted checkpoint and a baseline (typically the unadapted source model) generate free-text responses, and a high-capacity judge assigns a win, loss, or tie. To mitigate position bias, the assignment to positions A and B is randomized per comparison \cite{zheng2023judging}.

The pairwise data is aggregated into global rankings using the \textit{Davidson} model \cite{davidson1970extending}, an extension of \textit{Bradley-Terry} \cite{bradley1952rank} that accounts for ties. This yields latent quality scores on a linear scale, providing a normalized measure of response quality across the training trajectory.

\section{Experimental Setup}
\label{sec:setup}

This section describes the technical implementation of our domain adaptation pipeline.

\subsection{Domain-Specific Corpus}

We target the German medical domain, a high-resource non-English language with distinct clinical terminology that serves as a representative test case for cross-lingual domain adaptation while addressing the practical need for locally compliant AI in European healthcare.
The continual pre-training phase utilizes a specialized subset of the \textit{FineWeb-2} dataset \cite{penedo2025fineweb2}. To ensure high domain relevance, we employ a filtered version of the corpus specifically tailored for the German medical domain, introduced by \citet{doll-etal-2026-continual}. The filtered dataset consists of $7.3M$ unique documents, which amount to $9.5B$ tokens with respect to the \textit{Qwen2.5} tokenizer.

\subsection{Model Selection}

We adapt the instruction-tuned variants of two consecutive \textit{Qwen} generations, \textit{Qwen2.5} \citep{qwen2025qwen25technicalreport} and \textit{Qwen3} \citep{yang2025qwen3technicalreport}, each with four model sizes ranging from $0.5B$ to $8B$. This choice follows from two requirements of our pipeline. First, our study compares adaptation across model sizes and therefore needs instruction-tuned checkpoints at several sizes from a single release line, and second, the models must have strong multilingual capability that covers German. Among the families we surveyed, only \textit{Qwen} meets both requirements, offering four instruction-tuned sizes per generation in this range, with German among the supported languages \citep{qwen2024qwen25card, yang2025qwen3technicalreport}. \textit{LLaMA} \citep{grattafiori2024llama3herdmodels, meta2024llama32} mixes the $3.1$ and $3.2$ release lines below $8B$, \textit{Ministral} \citep{mistral2024ministraux} and \textit{Gemma~3} \citep{gemmateam2025gemma3} offer only two sizes in this range, and \textit{Pythia} \citep{biderman2023pythia} has no instruction-tuned variants. Using two generations of the same family further allows us to test whether our findings hold across pre-training generations while keeping the architectural lineage fixed. \textit{Qwen2.5} also matches the family used by \citet{junior-2025-interplay}, our closest related work, which eases comparison with their results.

\subsection{Evaluation Setup}

Our objective evaluation focuses on the German medical domain.
Following the methodology of \citet{doll-etal-2026-continual}, we utilize the German subset of \textit{MMMLU} \cite{openaiMMMLU}, a professionally human-translated multilingual version of the MMLU (Massive Multitask Language Understanding) benchmark \cite{hendryckstest2021}. Specifically, we select the medical subtasks with the highest share of error-free questions as reported by \citet{gema-etal-2025-done}, ensuring the evaluation is grounded in high-signal data. These include \textit{Clinical Knowledge}, \textit{Anatomy} and \textit{College Medicine}. Additionally, we include the machine-translated German version of the \textit{MedQA} benchmark \cite{medqa} introduced by \citet{doll-etal-2026-continual}. We use the Eval-Harness framework \cite{eval-harness} for all \textit{MCQA} evaluations.

For open-ended evaluation, we utilize a machine-translated version of the \textit{MedInstruct} dataset \cite{zhang2023alpacare} as the source of prompts, consisting of $216$ entries. Responses are generated using \textit{vLLM} \cite{vllm} and evaluated using a head-to-head comparison against the selected baseline. We employ \textit{GPT-5-mini} as our judge model. A preliminary annotator agreement study (Appendix~\ref{sec:judge-selection}) found that \textit{GPT-5-mini} achieves Cohen's $\kappa$ agreement with the strongest available model comparable to substantially more expensive alternatives, making it the most suitable choice given the large number of pairwise comparisons required across all checkpoints and model scales. The exact prompt template used by the judge model is presented in Figure \ref{fig:judge-model-prompt-template} in the Appendix.

To diagnose which dimensions of the response drive the pairwise ranking signal, we additionally score the same \textit{MedInstruct} responses along five of the six aspects defined in the rubric of \citet{lin2024the}, namely helpfulness, clarity, factuality, depth, and engagement (omitting safety), with the same \textit{GPT-5-mini} judge. This per-aspect decomposition is used in support of the interpretation of the ranking trajectories rather than as an independent evaluation. The full procedure and results are reported in Appendix~\ref{sec:per-aspect}.

We additionally evaluate all checkpoints on a German subset of \textit{HealthBench} \cite{arora2025healthbench}, a multilingual benchmark of single- and multi-turn health conversations scored by physician-authored rubrics. The subset consists of $180$ samples and includes the $29$ samples that are originally German in the benchmark, while the remaining $151$ were machine-translated. For more information see Appendix \ref{sec:healthbench-results}.

\begin{figure*}[ht]
    \centering

    \begin{subfigure}{0.49\linewidth}
        \centering
        \includegraphics[width=\linewidth]{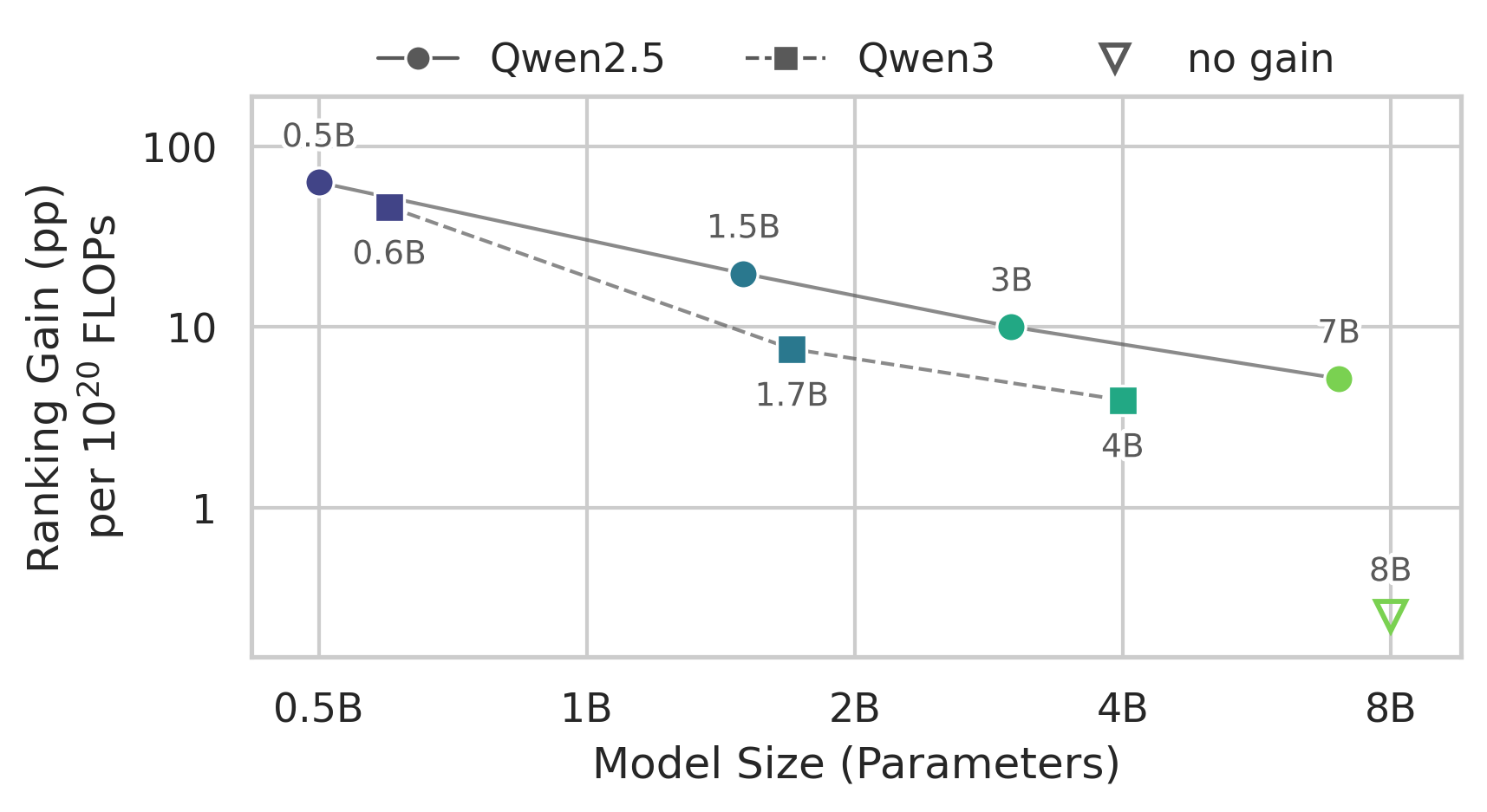}
    \end{subfigure}
    \hfill
    \begin{subfigure}{0.49\linewidth}
        \centering
        \includegraphics[width=\linewidth]{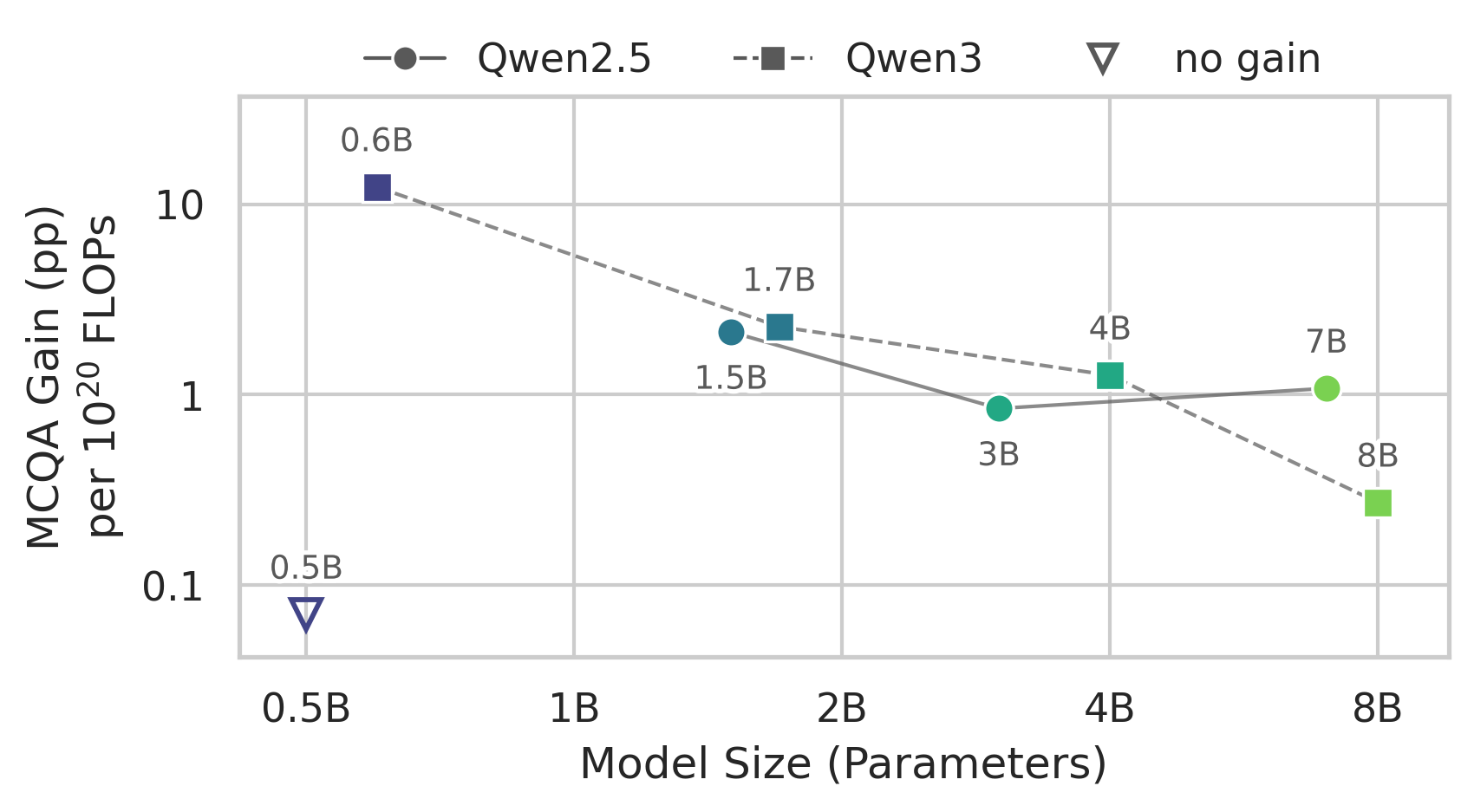}
    \end{subfigure}
    \caption{Adaptation efficiency as a function of model size for preference rankings (left) and \textit{MCQA} (right). Efficiency is the gain of the final checkpoint over its source model in percentage points per $10^{20}$ training FLOPs, shown on a log scale. For rankings, the gain is the win rate against the source model above $50\%$. Hollow triangles mark models whose final checkpoint shows no gain. For preference rankings, efficiency drops by more than an order of magnitude from the smallest to the largest model with a positive gain in both families.}
    \label{fig:adaptation-efficiency}
\end{figure*}

\section{Findings}
\label{sec:findings}

All reported metrics correspond to the merged checkpoints from the protocol in Section~\ref{sec:method:pipeline}. We organize the findings around three insights from the joint analysis of factual knowledge and open-ended generation, rather than by evaluation metric.

\subsection{Adaptation Trajectories and Convergence}
\label{sec:trajectories}

A central finding of this study is that factual knowledge and open-ended response quality converge at different rates during domain adaptation, consistently across both model families and all parameter scales. We first describe the trajectories of both axes, then quantify their convergence speed, and finally examine what the early gains in response quality consist of.

As illustrated in Figures \ref{fig:qwen2.5-mcqa-vs-source-tokens} and \ref{fig:qwen3-mcqa-vs-source-tokens}, \textit{MCQA} performance improves gradually, with the largest relative gains within the first $4B$ tokens, followed by a plateau. The absolute accuracy trajectories in Figures \ref{fig:qwen2.5-mcqa} and \ref{fig:qwen3-mcqa} in the Appendix confirm that accuracy remains primarily a function of model scale, with \textit{Qwen2.5} $7B$ and \textit{Qwen3} $0.6B$ achieving the strongest relative gains in their respective families.

The preference rankings in Figures \ref{fig:qwen2.5-ranking-vs-source-tokens} and \ref{fig:qwen3-ranking-vs-source-tokens} follow a faster course. All models rise steeply against their respective source models and typically plateau within $2B$ to $3B$ tokens, at ranking scores between $0.61$ and $0.70$ for \textit{Qwen2.5} and between $0.53$ and $0.64$ for \textit{Qwen3}. At the smallest sizes, ties dominate the pairwise judgments, and their share shrinks as model size grows, as shown in Appendix~\ref{sec:win-tie-loss}.

To compare the convergence speed of the two axes beyond visual inspection, we fit a saturating exponential to each trajectory over training tokens and report $c_{90}$, the number of tokens required to reach $90\%$ of the fitted total gain. We keep only trajectories that clearly saturate and refer to Appendix~\ref{sec:convergence-analysis} for more detail. As summarized in Table~\ref{tab:c90-summary}, preference rankings reach this point after a median of $1.7B$ tokens, whereas \textit{MCQA} accuracy requires $3.5B$ tokens, and rankings converge earlier than \textit{MCQA} in $5$ of the $6$ models for which both fits are retained. The per-model confidence intervals on $c_{90}$ are wide, and with six models the difference does not reach conventional significance with a Wilcoxon signed-rank $p = 0.06$. We therefore read it as a consistent tendency rather than a precise speed-up factor.

\begin{table}[t]
    \centering
    \small
    \begin{tabular}{lcc}
        \toprule
        Metric & Median $c_{90}$ & Fits retained \\
        \midrule
        Preference rankings      & $1.7B$ & $8/8$ \\
        Token-distribution shift & $1.9B$ & $5/8$ \\
        \textit{MCQA} accuracy   & $3.5B$ & $6/8$ \\
        \bottomrule
    \end{tabular}
    \caption{Training tokens required to reach $90\%$ of the fitted total gain ($c_{90}$, lower is faster), as the median over all models across both families whose trajectories clearly saturate. Per-model values and confidence intervals are reported in Appendix~\ref{sec:convergence-analysis}.}
    \label{tab:c90-summary}
\end{table}

The faster convergence of the rankings raises the question of what the early gains consist of. The per-aspect scores introduced in Section~\ref{sec:setup}, with full results in Appendix~\ref{sec:per-aspect}, show that all five aspects improve over the course of training, but the stylistic aspects helpfulness, clarity and engagement reach their ceilings early. Factuality, in contrast, keeps improving until the end of training and closely tracks the corresponding \textit{MCQA} curves at every size, most visibly for \textit{Qwen2.5} $3B$ and $7B$. In line with this, the rubric-based \textit{HealthBench} scores in Appendix~\ref{sec:healthbench-results} end above the source baselines only for \textit{Qwen2.5} $7B$ and \textit{Qwen3} $0.6B$, the two models with the clearest \textit{MCQA} gains, which suggests that physician-written rubrics reward factual content more than surface form.

Because both the rankings and the per-aspect scores rely on an LLM judge, we add a judge-free check, namely the Jensen--Shannon divergence between the output token distribution of each checkpoint and that of its source model on the same \textit{MedInstruct} generations, as detailed in Appendix~\ref{sec:token-shift}. This shift is strongly front-loaded at every model size, with smaller magnitudes for larger models, and reaches its $c_{90}$ after a median of $1.9B$ tokens, close to the preference rankings in Table~\ref{tab:c90-summary}. The surface form of the outputs therefore settles early without any judge involved. Together, these results indicate that the early preference gains largely reflect surface alignment from \textit{CPT}, while later gains increasingly reflect factual knowledge, which constitutes the first key finding of this study.

\subsection{Scale-Dependent Gap-Closing}
\label{sec:gap_closing}

The previous section characterised how adapted models evolve relative to their own source baselines. A separate, deployment-oriented question is how a domain-adapted smaller model compares to a larger but unadapted generalist of the same family. We therefore evaluate how much of the next-tier performance gap is closed by \textit{CPT} on the smaller model. This comparison reflects a practical deployment trade-off rather than a pure measure of adaptation gain. The gap our adapted models close mixes the capacity-dependent benefit of adaptation with the distributional mismatch of the unadapted larger baseline, which has not been exposed to German medical text. The numbers below should be read as practical deployment outcomes, not as estimates of the adaptation benefit in isolation.

As illustrated in Figures \ref{fig:qwen2.5-mcqa-vs-larger-tokens} and \ref{fig:qwen3-mcqa-vs-larger-tokens}, domain adaptation narrows the practical \textit{MCQA} gap across both model families, with a clear capacity-scaling effect. In the \textit{Qwen3} family, the $0.6B$ and $1.7B$ models reach approximately $85\%$ of the accuracy of their respective larger counterparts, while the $4B$ and $8B$ models reach upwards of $92.5\%$. The \textit{Qwen2.5} $1.5B$ model is a notable exception, reaching over $93\%$ of the $3B$ variant's performance, the highest relative parity observed in the \textit{Qwen2.5} series.

The preference rankings against the respective larger models in Figures \ref{fig:qwen2.5-ranking-vs-larger-tokens} and \ref{fig:qwen3-ranking-vs-larger-tokens} exhibit the same scale-dependent pattern. In the lower parameter regimes, the larger generalist models maintain a substantial lead. For example, the \textit{Qwen2.5} $3B$ model's ranking against the $7B$ version improves from $0.07$ to $0.16$ through adaptation. The gap narrows but remains wide, indicating that for smaller architectures \textit{CPT} primarily reduces the ranking gap rather than achieving parity.

As model size increases, the gap closes substantially, mirroring the \textit{MCQA} observations. The \textit{Qwen2.5} $7B$ model exhibits the most pronounced narrowing, improving its preference score against the $14B$ generalist from $0.26$ to $0.46$. Subject to the deployment-oriented framing stated above, an adapted $7B$ in this regime is a plausible alternative to the unadapted $14B$ generalist when domain register matters and additional hardware for the larger model is not available. In the \textit{Qwen3} family, the $4B$ model improves from $0.19$ to $0.30$. The \textit{HealthBench} trajectories (Appendix \ref{sec:healthbench-results}) point in the same direction within the \textit{Qwen2.5} family, where the $7B$ variant is the only model that durably surpasses its source baseline, with the gain again concentrated in the late training stage.

\subsection{Compute Efficiency Across Scales}
\label{sec:compute_efficiency}

The preceding sections establish that larger models achieve higher absolute performance and close more of the gap to the next parameter tier. However, this analysis does not account for the computational cost of these gains. When the adaptation trajectories are normalized by cumulative compute rather than tokens, a fundamentally different picture emerges.

As shown in the FLOPs-normalized trajectories (Appendix Figures \ref{fig:qwen2.5-mcqa-vs-source-flops}, \ref{fig:qwen3-mcqa-vs-source-flops}, \ref{fig:qwen2.5-ranking-vs-source-flops}, and \ref{fig:qwen3-ranking-vs-source-flops}), models across all scales improve within a similar number of tokens, but diverge strongly in compute efficiency. Smaller models reach their respective performance ceilings at a fraction of the total FLOPs investment. For example, the \textit{Qwen2.5} $0.5B$ model achieves its full relative gain at approximately $10^{18}$ to $10^{19}$ FLOPs, while the $7B$ model requires approximately $10^{20}$ FLOPs to reach comparable relative improvements. This represents an order-of-magnitude difference in compute cost, while the absolute gains differ far less across sizes.

To quantify this relationship, Figure~\ref{fig:adaptation-efficiency} plots the adaptation efficiency against model size, defined as the gain of the final checkpoint over its source model in percentage points per $10^{20}$ training FLOPs. For preference rankings, efficiency falls steadily with size in both families, from $63$ to $5$ for \textit{Qwen2.5} and from $46$ to $4$ up to \textit{Qwen3}~$4B$, while \textit{Qwen3}~$8B$ ends slightly below its source model. Larger models therefore pay considerably more compute for each point of ranking gain. For \textit{MCQA}, efficiency also falls with size for \textit{Qwen3}, whereas \textit{Qwen2.5} flattens beyond $1.5B$ and its $0.5B$ model ends slightly below its source model. Because \textit{Qwen3}~$0.6B$ reaches the highest \textit{MCQA} efficiency of all models, this exception more likely reflects the individual model than a general capacity threshold for acquiring factual knowledge.

Together, these results reveal a clear trade-off. For a fixed compute budget, adapting a smaller model yields more gain per FLOP, but the absolute performance ceiling remains lower. Which size offers the best balance therefore depends on whether the compute budget or the absolute performance level is the binding constraint.

\section{Discussion}
\label{sec:discussion}

The decoupled adaptation dynamics observed in Section~\ref{sec:trajectories} have both methodological and practical implications. Preference rankings and the surface form of the outputs settle within roughly the first $2B$ tokens, while \textit{MCQA} accuracy keeps improving beyond that point. Prior scaling studies were designed around a single target quantity, such as pre-training loss \cite{hoffmann2022training} or \textit{MCQA} perplexity \cite{junior-2025-interplay}, which suits the questions they address. For the domain adaptation of instruction-tuned models, our results suggest that tracking open-ended response quality alongside knowledge adds a complementary view, because the two axes need not converge together. Methodologically, judge-based preference scores can plateau while knowledge is still being acquired, so they should not be used on their own to decide when adaptation has converged or whether a model is ready for domain use. Practically, the risk is most acute in the short-\textit{CPT} regime, where a model that has been adapted only long enough to acquire the domain register, but not long enough to accumulate factual content, can sound authoritative without the corresponding factual grounding.

Our findings also bear on reinforcement learning from AI feedback (\textit{RLAIF}), in which an LLM judge provides the preference or reward signal \citep{lee2024rlaif}, for example through rubric-based rewards \citep{gunjal2025rubrics}. In our setting, judge-based preference scores saturated within roughly $2B$ tokens, largely through surface alignment rather than added factual knowledge. Used as a reward, the same signal would credit a policy for adopting the domain register well before its factual content improves. This supports recent concerns that LLM judges favor style over factuality \citep{feuer2025style} and that \textit{RLAIF} policies learn to exploit systematic judge errors as training progresses \citep{kenton2026debate}. In contrast, the rubric-based \textit{HealthBench} scores in our study followed \textit{MCQA} rather than the early preference gains, which favors rubrics that tie reward to specific content \citep{gunjal2025rubrics}. For clinical \textit{RLAIF}, we therefore suggest pairing judge-based rewards with knowledge-oriented checks and tracking both axes separately during training.

The practical question for institutional deployment is model selection under resource constraints. Larger adapted models close progressively more of the performance gap to the next parameter tier (Section~\ref{sec:gap_closing}), but at disproportionate compute cost (Section~\ref{sec:compute_efficiency}). The shape of this trade-off resembles the compute-optimal behavior that \citet{hoffmann2022training} describe for pre-training from scratch, although our setting differs in starting from instruction-tuned models adapted with \textit{SLERP} merging and in measuring downstream domain performance rather than pre-training loss. For environments where hardware limitations or inference costs are primary concerns, adapting a smaller or mid-sized model is therefore a more viable strategy than simply scaling up. Further, by evaluating on general-domain benchmarks in Appendix \ref{sec:forgetting}, we observe no catastrophic forgetting in the adapted models.

Read together, our findings suggest the following orientation for resource-constrained clinical deployments. If the goal is to obtain a model that responds in a domain-appropriate register, the bulk of the ranking gain is realised within the first $2B$ to $3B$ tokens of \textit{CPT}, and continued training yields diminishing returns on this axis. If factual capability is the priority, longer training is required, but the marginal gain remains modest in absolute terms and larger models need considerably more compute per point of gain, which in the families we evaluated points to mid-sized models ($3B$ to $7B$) as a reasonable trade-off rather than a validated optimum. Where factual precision is the binding constraint, \textit{CPT} on its own is unlikely to be the most effective lever. Retrieval-augmented generation against curated medical sources has been shown to improve the factuality of clinical answers \citep{zakka2024almanac} and to raise medical question-answering accuracy for the same model without changing its size \citep{xiong-etal-2024-benchmarking}, and is a natural complement to the surface alignment that \textit{CPT} provides.

\section{Conclusion}
\label{sec:conclusion}

This study empirically characterizes domain adaptation dynamics in the German medical domain across the \textit{Qwen2.5} and \textit{Qwen3} families through a dual-metric evaluation. The decoupled trajectories of knowledge and generation, the scale-dependent gap-closing, and the steeply rising compute cost of adaptation with model size together indicate that domain proficiency is a multi-dimensional process rather than a single curve to optimize. Methodologically, judge-based preference scores can plateau while \textit{MCQA} knowledge is still improving, so they should not be used alone to judge adaptation convergence. Future work could extend the analysis to architecturally independent model families, to other specialised domains and languages, and integrate post-training and retrieval-augmented strategies as complementary levers.

\section*{Limitations}

This study is subject to several constraints that define the scope of its conclusions. The empirical setting is bounded to the German medical domain, to two related \textit{Qwen} model families that share an architectural lineage, and to parameter counts up to $8B$. Cross-family validation against architecturally independent models such as \textit{LLaMA} or \textit{Mistral} and the behaviour of larger architectures therefore remain open, as does replication in domains other than medicine. A second constraint concerns data contamination. Because the pre-training data of the \textit{Qwen} models is not publicly disclosed, we cannot rule out overlap between our evaluation benchmarks and the base training of either family \cite{xu2024benchmarkdatacontaminationlarge}. Using translated versions of established English benchmarks (human-translated for \textit{MMMLU}, machine-translated for the remaining evaluation data) removes exact string overlap with the English originals, but contamination can carry across languages \cite{yao2024crosslingual}, so translation only partly reduces this risk, and the same translation-based construction introduces its own limitation. The \textit{HealthBench} subset on which we report (Appendix \ref{sec:healthbench-results}) is mostly machine-translated rather than authored in German, and the per-aspect factuality scoring (Appendix \ref{sec:per-aspect}) is bounded by the judge model's own medical knowledge, which we cross-check against the independent \textit{MCQA} measurements but cannot fully separate. More broadly, our reliance on \textit{MCQA}, preference rankings, and the rubric-based \textit{HealthBench} subset provides a proxy for clinical proficiency rather than a direct measure of it, because multi-step diagnostic reasoning, refusal behaviour, and long-term utility in a clinical workflow are not evaluated. Although our judge model was selected following best practices via an annotator agreement study against the strongest available reference model (Appendix \ref{sec:judge-selection}), the evaluation ultimately relies on automated \textit{LLM}-as-a-Judge assessments without human expert verification. A human evaluation by qualified medical professionals would strengthen the reliability of the preference rankings but was not conducted due to the unavailability of domain experts for data annotation. Finally, the preference ranking trajectories exhibit substantially higher variability across training checkpoints than the \textit{MCQA} measurements, particularly for the smallest model variants where high tie rates compress the effective scoring range, so fine-grained trajectory comparisons in the ranking signal should be interpreted at a coarse temporal resolution rather than at the level of individual checkpoints. For the same reason, the per-model convergence estimates (Appendix~\ref{sec:convergence-analysis}) carry wide confidence intervals, and the faster convergence of preference rankings relative to \textit{MCQA} is a consistent directional tendency across families rather than a precisely estimated speed-up.

\bibliography{custom}

@inproceedings{loshchilov2019decoupledweightdecayregularization,
  author       = {Ilya Loshchilov and
                  Frank Hutter},
  title        = {Decoupled Weight Decay Regularization},
  booktitle    = {7th International Conference on Learning Representations, {ICLR} 2019,
                  New Orleans, LA, USA, May 6-9, 2019},
  publisher    = {OpenReview.net},
  year         = {2019},
  url          = {https://openreview.net/forum?id=Bkg6RiCqY7},
  bibsource    = {dblp computer science bibliography, https://dblp.org}
}

@misc{axolotl,
  author       = {{Axolotl AI}},
  title        = {Axolotl},
  year         = {2023},
  publisher    = {GitHub},
  howpublished = {\url{https://github.com/axolotl-ai-cloud/axolotl}}
}

@misc{qwen2025qwen25technicalreport,
      title={Qwen2.5 Technical Report}, 
      author={{Qwen} and An Yang and Baosong Yang and Beichen Zhang and Binyuan Hui and Bo Zheng and Bowen Yu and Chengyuan Li and Dayiheng Liu and Fei Huang and Haoran Wei and Huan Lin and Jian Yang and Jianhong Tu and Jianwei Zhang and Jianxin Yang and Jiaxi Yang and Jingren Zhou and Junyang Lin and Kai Dang and Keming Lu and Keqin Bao and Kexin Yang and Le Yu and Mei Li and Mingfeng Xue and Pei Zhang and Qin Zhu and Rui Men and Runji Lin and Tianhao Li and Tianyi Tang and Tingyu Xia and Xingzhang Ren and Xuancheng Ren and Yang Fan and Yang Su and Yichang Zhang and Yu Wan and Yuqiong Liu and Zeyu Cui and Zhenru Zhang and Zihan Qiu},
      year={2025},
      eprint={2412.15115},
      archivePrefix={arXiv},
      primaryClass={cs.CL},
      url={https://arxiv.org/abs/2412.15115}, 
}

@inproceedings{wolf-etal-2020-transformers,
    title = "Transformers: State-of-the-Art Natural Language Processing",
    author = "Wolf, Thomas  and
      Debut, Lysandre  and
      Sanh, Victor  and
      Chaumond, Julien  and
      Delangue, Clement  and
      Moi, Anthony  and
      Cistac, Pierric  and
      Rault, Tim  and
      Louf, Remi  and
      Funtowicz, Morgan  and
      Davison, Joe  and
      Shleifer, Sam  and
      von Platen, Patrick  and
      Ma, Clara  and
      Jernite, Yacine  and
      Plu, Julien  and
      Xu, Canwen  and
      Le Scao, Teven  and
      Gugger, Sylvain  and
      Drame, Mariama  and
      Lhoest, Quentin  and
      Rush, Alexander",
    editor = "Liu, Qun  and
      Schlangen, David",
    booktitle = "Proceedings of the 2020 Conference on Empirical Methods in Natural Language Processing: System Demonstrations",
    month = oct,
    year = "2020",
    address = "Online",
    publisher = "Association for Computational Linguistics",
    url = "https://aclanthology.org/2020.emnlp-demos.6/",
    doi = "10.18653/v1/2020.emnlp-demos.6",
    pages = "38--45"
}

@article{zhao2023pytorchfsdpexperiencesscaling,
  title    = {PyTorch FSDP: Experiences on Scaling Fully Sharded Data Parallel},
  author   = {Zhao, Yanli and Gu, Andrew and Varma, Rohan and Luo, Liang and Huang, Chien-Chin and Xu, Min and Wright, Less and Shojanazeri, Hamid and Ott, Myle and Shleifer, Sam and Desmaison, Alban and Balioglu, Can and Damania, Pritam and Nguyen, Bernard and Chauhan, Geeta and Hao, Yuchen and Mathews, Ajit and Li, Shen},
  journal  = {Proceedings of the VLDB Endowment},
  volume   = {16},
  number   = {12},
  pages    = {3848--3860},
  year     = {2023},
  publisher= {Association for Computing Machinery (ACM)},
  doi      = {10.14778/3611540.3611569},
  issn     = {2150-8097}
}

@inproceedings{labrak2024biomistralcollectionopensourcepretrained,
    title = "{B}io{M}istral: A Collection of Open-Source Pretrained Large Language Models for Medical Domains",
    author = "Labrak, Yanis  and
      Bazoge, Adrien  and
      Morin, Emmanuel  and
      Gourraud, Pierre-Antoine  and
      Rouvier, Mickael  and
      Dufour, Richard",
    editor = "Ku, Lun-Wei  and
      Martins, Andre  and
      Srikumar, Vivek",
    booktitle = "Findings of the {A}ssociation for {C}omputational {L}inguistics: {ACL} 2024",
    month = aug,
    year = "2024",
    address = "Bangkok, Thailand",
    publisher = "Association for Computational Linguistics",
    url = "https://aclanthology.org/2024.findings-acl.348/",
    doi = "10.18653/v1/2024.findings-acl.348",
    pages = "5848--5864"
}

@misc{yang2024modelmergingllmsmllms,
      title={Model Merging in {LLMs}, {MLLMs}, and {Beyond}: {Methods, Theories, Applications and Opportunities}}, 
      author={Enneng Yang and Li Shen and Guibing Guo and Xingwei Wang and Xiaochun Cao and Jie Zhang and Dacheng Tao},
      year={2024},
      eprint={2408.07666},
      archivePrefix={arXiv},
      primaryClass={cs.LG},
      url={https://arxiv.org/abs/2408.07666}, 
}

@inproceedings{goddard-etal-2024-arcees,
    title = "Arcee{'}s {M}erge{K}it: A Toolkit for Merging Large Language Models",
    author = "Goddard, Charles  and
      Siriwardhana, Shamane  and
      Ehghaghi, Malikeh  and
      Meyers, Luke  and
      Karpukhin, Vladimir  and
      Benedict, Brian  and
      McQuade, Mark  and
      Solawetz, Jacob",
    editor = "Dernoncourt, Franck  and
      Preo{\c{t}}iuc-Pietro, Daniel  and
      Shimorina, Anastasia",
    booktitle = "Proceedings of the 2024 Conference on Empirical Methods in Natural Language Processing: Industry Track",
    month = nov,
    year = "2024",
    address = "Miami, Florida, US",
    publisher = "Association for Computational Linguistics",
    url = "https://aclanthology.org/2024.emnlp-industry.36",
    doi = "10.18653/v1/2024.emnlp-industry.36",
    pages = "477--485",
}

@inproceedings{shoemake-1985-slerp,
  title    = {Animating rotation with quaternion curves},
  author   = {Shoemake, Ken},
  booktitle= {Proceedings of the 12th Annual Conference on Computer Graphics and Interactive Techniques (SIGGRAPH '85)},
  pages    = {245--254},
  year     = {1985},
  publisher= {ACM Press},
  doi      = {10.1145/325334.325242}
}

@inproceedings{nobari2025,
 author = {Heyrani Nobari, Amin and Alimohammadi, Kaveh and ArjomandBigdeli, Ali and Srivastava, Akash and Ahmed, Faez and Azizan, Navid},
 booktitle = {Advances in Neural Information Processing Systems},
 doi = {10.52202/085713-2446},
 editor = {D. Belgrave and C. Zhang and H. Lin and R. Pascanu and P. Koniusz and M. Ghassemi and N. Chen},
 pages = {72975--72998},
 publisher = {Curran Associates, Inc.},
 title = {Activation-Informed Merging of Large Language Models},
 url = {https://proceedings.neurips.cc/paper_files/paper/2025/file/69b576be5219a18195f2dba1ae1ffceb-Paper-Conference.pdf},
 volume = {38, Main Conference},
 year = {2025}
}

@article{yadav2024,
title={What Matters for Model Merging at Scale?},
author={Prateek Yadav and Tu Vu and Jonathan Lai and Alexandra Chronopoulou and Manaal Faruqui and Mohit Bansal and Tsendsuren Munkhdalai},
journal={Transactions on Machine Learning Research},
issn={2835-8856},
year={2025},
url={https://openreview.net/forum?id=9sbetmvNpW},
note={}
}

@inproceedings{gema-etal-2025-done,
    title = "Are We Done with {MMLU}?",
    author = "Gema, Aryo Pradipta  and
      Leang, Joshua Ong Jun  and
      Hong, Giwon  and
      Devoto, Alessio  and
      Mancino, Alberto Carlo Maria  and
      Saxena, Rohit  and
      He, Xuanli  and
      Zhao, Yu  and
      Du, Xiaotang  and
      Ghasemi Madani, Mohammad Reza  and
      Barale, Claire  and
      McHardy, Robert  and
      Harris, Joshua  and
      Kaddour, Jean  and
      Van Krieken, Emile  and
      Minervini, Pasquale",
    editor = "Chiruzzo, Luis  and
      Ritter, Alan  and
      Wang, Lu",
    booktitle = "Proceedings of the 2025 Conference of the Nations of the {A}mericas Chapter of the {A}ssociation for {C}omputational {L}inguistics: Human Language Technologies (Volume 1: Long Papers)",
    month = apr,
    year = "2025",
    address = "Albuquerque, New Mexico",
    publisher = "Association for Computational Linguistics",
    url = "https://aclanthology.org/2025.naacl-long.262/",
    doi = "10.18653/v1/2025.naacl-long.262",
    pages = "5069--5096",
    ISBN = "979-8-89176-189-6"
}

@article{meyer2024,
    author="Meyer, Annika
    and Riese, Janik
    and Streichert, Thomas",
    title="Comparison of the Performance of GPT-3.5 and GPT-4 With That of Medical Students on the Written German Medical Licensing Examination: Observational Study",
    journal="JMIR Med Educ",
    year="2024",
    month="Feb",
    day="8",
    volume="10",
    pages="e50965",
    issn="2369-3762",
    doi="10.2196/50965",
    url="https://doi.org/10.2196/50965",
}

@misc{eval-harness,
  author       = {Gao, Leo and Tow, Jonathan and Abbasi, Baber and Biderman, Stella and Black, Sid and DiPofi, Anthony and Foster, Charles and Golding, Laurence and Hsu, Jeffrey and Le Noac'h, Alain and Li, Haonan and McDonell, Kyle and Muennighoff, Niklas and Ociepa, Chris and Phang, Jason and Reynolds, Laria and Schoelkopf, Hailey and Skowron, Aviya and Sutawika, Lintang and Tang, Eric and Thite, Anish and Wang, Ben and Wang, Kevin and Zou, Andy},
  title        = {A framework for few-shot language model evaluation},
  month        = 07,
  year         = 2024,
  publisher    = {Zenodo},
  version      = {v0.4.3},
  doi          = {10.5281/zenodo.12608602},
  url          = {https://zenodo.org/records/12608602}
}

@Article{medqa,
AUTHOR = {Jin, Di and Pan, Eileen and Oufattole, Nassim and Weng, Wei-Hung and Fang, Hanyi and Szolovits, Peter},
TITLE = {What Disease Does This Patient Have? A Large-Scale Open Domain Question Answering Dataset from Medical Exams},
JOURNAL = {Applied Sciences},
VOLUME = {11},
YEAR = {2021},
NUMBER = {14},
ARTICLE-NUMBER = {6421},
URL = {https://www.mdpi.com/2076-3417/11/14/6421},
ISSN = {2076-3417},
DOI = {10.3390/app11146421}
}

@inproceedings{hendryckstest2021,
title={Measuring Massive Multitask Language Understanding},
author={Dan Hendrycks and Collin Burns and Steven Basart and Andy Zou and Mantas Mazeika and Dawn Song and Jacob Steinhardt},
booktitle={International Conference on Learning Representations},
year={2021},
url={https://openreview.net/forum?id=d7KBjmI3GmQ}
}

@misc{openaiMMMLU,
  author = {OpenAI},
  title = {Multilingual Massive Multitask Language Understanding (MMMLU)},
  publisher = {Hugging Face},
  journal = {Hugging Face repository},
  year = {2024},
  url = {https://huggingface.co/datasets/openai/MMMLU}
}

@misc{GPT-4.1-mini,
  author       = {OpenAI},
  title        = {{Introducing GPT-4.1 in the API}},
  howpublished = {\url{https://openai.com/index/gpt-4-1/}},
  note         = {Accessed: 2025-10-07},
  year         = {2025},
  month        = {Apr},
}

@inproceedings{oncel2024,
    title = "Adaptation Odyssey in {LLM}s: Why Does Additional Pretraining Sometimes Fail to Improve?",
    author = {{\"O}ncel, F{\i}rat  and
      Bethge, Matthias  and
      Ermis, Beyza  and
      Ravanelli, Mirco  and
      Subakan, Cem  and
      Y{\i}ld{\i}z, {\c{C}}a{\u{g}}atay},
    editor = "Al-Onaizan, Yaser  and
      Bansal, Mohit  and
      Chen, Yun-Nung",
    booktitle = "Proceedings of the 2024 Conference on Empirical Methods in Natural Language Processing",
    month = nov,
    year = "2024",
    address = "Miami, Florida, USA",
    publisher = "Association for Computational Linguistics",
    url = "https://aclanthology.org/2024.emnlp-main.1108/",
    doi = "10.18653/v1/2024.emnlp-main.1108",
    pages = "19834--19843"
}

@inproceedings{niyogi2024,
    title = "{AYN}: A Tiny Yet Competitive {I}ndian Legal Language Model Pretrained from Scratch",
    author = "Niyogi, Mitodru  and
      Gaussier, Eric  and
      Bhattacharya, Arnab",
    editor = "Piperidis, Stelios  and
      Bel, N{\'u}ria  and
      van den Heuvel, Henk  and
      Ide, Nancy  and
      Krek, Simon  and
      Toral, Antonio",
    booktitle = "Proceedings of the Fifteenth Language Resources and Evaluation Conference",
    month = may,
    year = "2026",
    address = "Palma de Mallorca, Spain",
    publisher = "ELRA Language Resource Association",
    url = "https://aclanthology.org/2026.lrec-1.839/",
    doi = "10.63317/3c62f5f9g7q2",
    pages = "10699--10722"
}

@misc{siriwardhana2024,
      title={Domain Adaptation of Llama3-70B-Instruct through Continual Pre-Training and Model Merging: A Comprehensive Evaluation}, 
      author={Shamane Siriwardhana and Mark McQuade and Thomas Gauthier and Lucas Atkins and Fernando Fernandes Neto and Luke Meyers and Anneketh Vij and Tyler Odenthal and Charles Goddard and Mary MacCarthy and Jacob Solawetz},
      year={2024},
      eprint={2406.14971},
      archivePrefix={arXiv},
      primaryClass={cs.CL},
      url={https://arxiv.org/abs/2406.14971}, 
}

@article{Yang2024b,
  title    = {TCM-GPT: Efficient pre-training of large language models for domain adaptation in Traditional Chinese Medicine},
  author   = {Yang, Guoxing and Liu, Xiaohong and Shi, Jianyu and Wang, Zan and Wang, Guangyu},
  journal  = {Computer Methods and Programs in Biomedicine Update},
  volume   = {6},
  pages    = {100158},
  year     = {2024},
  publisher= {Elsevier BV},
  doi      = {10.1016/j.cmpbup.2024.100158},
  issn     = {2666-9900}
}

@misc{belcak2025,
      title={Small Language Models are the Future of Agentic AI}, 
      author={Peter Belcak and Greg Heinrich and Shizhe Diao and Yonggan Fu and Xin Dong and Saurav Muralidharan and Yingyan Celine Lin and Pavlo Molchanov},
      year={2025},
      eprint={2506.02153},
      archivePrefix={arXiv},
      primaryClass={cs.AI},
      url={https://arxiv.org/abs/2506.02153}, 
}

@inproceedings{zellers2019hellaswag,
    title = "{H}ella{S}wag: Can a Machine Really Finish Your Sentence?",
    author = "Zellers, Rowan  and
      Holtzman, Ari  and
      Bisk, Yonatan  and
      Farhadi, Ali  and
      Choi, Yejin",
    editor = "Korhonen, Anna  and
      Traum, David  and
      M{\`a}rquez, Llu{\'i}s",
    booktitle = "Proceedings of the 57th Annual Meeting of the {A}ssociation for {C}omputational {L}inguistics",
    month = jul,
    year = "2019",
    address = "Florence, Italy",
    publisher = "Association for Computational Linguistics",
    url = "https://aclanthology.org/P19-1472/",
    doi = "10.18653/v1/P19-1472",
    pages = "4791--4800"
}

@misc{cobbe2021gsm8k,
      title={Training Verifiers to Solve Math Word Problems}, 
      author={Karl Cobbe and Vineet Kosaraju and Mohammad Bavarian and Mark Chen and Heewoo Jun and Lukasz Kaiser and Matthias Plappert and Jerry Tworek and Jacob Hilton and Reiichiro Nakano and Christopher Hesse and John Schulman},
      year={2021},
      eprint={2110.14168},
      archivePrefix={arXiv},
      primaryClass={cs.LG},
      url={https://arxiv.org/abs/2110.14168}, 
}

@misc{clark2018arc,
      title={Think you have Solved Question Answering? Try ARC, the AI2 Reasoning Challenge}, 
      author={Peter Clark and Isaac Cowhey and Oren Etzioni and Tushar Khot and Ashish Sabharwal and Carissa Schoenick and Oyvind Tafjord},
      year={2018},
      eprint={1803.05457},
      archivePrefix={arXiv},
      primaryClass={cs.AI},
      url={https://arxiv.org/abs/1803.05457}, 
}

@misc{junior-2025-interplay,
  title         = {The interplay between domain specialization and model size},
  author        = {Malaquias Junior, Roseval and Pires, Ramon and Almeida, Thales Sales and Sakiyama, Kenzo and Romero, Roseli A. F. and Nogueira, Rodrigo},
  year          = {2025},
  eprint        = {2501.02068},
  archivePrefix = {arXiv},
  primaryClass  = {cs.CL},
  url           = {https://arxiv.org/abs/2501.02068}
}

@misc{colombo-2024-saullm7b,
      title={SaulLM-7B: A pioneering Large Language Model for Law}, 
      author={Pierre Colombo and Telmo Pessoa Pires and Malik Boudiaf and Dominic Culver and Rui Melo and Caio Corro and Andre F. T. Martins and Fabrizio Esposito and Vera Lúcia Raposo and Sofia Morgado and Michael Desa},
      year={2024},
      eprint={2403.03883},
      archivePrefix={arXiv},
      primaryClass={cs.CL},
      url={https://arxiv.org/abs/2403.03883}, 
}

@inproceedings{gu-2024-cmr,
    title = "{CMR} Scaling Law: Predicting Critical Mixture Ratios for Continual Pre-training of Language Models",
    author = "Gu, Jiawei  and
      Yang, Zacc  and
      Ding, Chuanghao  and
      Zhao, Rui  and
      Tan, Fei",
    editor = "Al-Onaizan, Yaser  and
      Bansal, Mohit  and
      Chen, Yun-Nung",
    booktitle = "Proceedings of the 2024 Conference on Empirical Methods in Natural Language Processing",
    month = nov,
    year = "2024",
    address = "Miami, Florida, USA",
    publisher = "Association for Computational Linguistics",
    url = "https://aclanthology.org/2024.emnlp-main.903/",
    doi = "10.18653/v1/2024.emnlp-main.903",
    pages = "16143--16162",
}

@article{kaplan2020scaling,
  title={Scaling laws for neural language models},
  author={Kaplan, Jared and McCandlish, Sam and Henighan, Tom and Brown, Tom B and Chess, Benjamin and Child, Rewon and Gray, Scott and Radford, Alec and Wu, Jeffrey and Amodei, Dario},
  journal={arXiv preprint arXiv:2001.08361},
  year={2020}
}

@inproceedings{hoffmann2022training,
  title={Training Compute-Optimal Large Language Models},
  author={Hoffmann, Jordan and Borgeaud, Sebastian and Mensch, Arthur and Buchatskaya, Elena and Cai, Trevor and Rutherford, Eliza and de Las Casas, Diego and Hendricks, Lisa Anne and Welbl, Johannes and Clark, Aidan and Hennigan, Tom and Noland, Eric and Millican, Katie and van den Driessche, George and Damoc, Bogdan and Guy, Aurelia and Osindero, Simon and Simonyan, Karen and Elsen, Erich and Vinyals, Oriol and Rae, Jack W. and Sifre, Laurent},
  booktitle={Advances in Neural Information Processing Systems (NeurIPS)},
  year={2022},
  url={https://arxiv.org/abs/2203.15556}
}

@inproceedings{holtzman2020curious,
  title={The Curious Case of Neural Text Degeneration},
  author={Holtzman, Ari and Buys, Jan and Du, Li and Forbes, Maxwell and Choi, Yejin},
  booktitle={International Conference on Learning Representations (ICLR)},
  year={2020},
  url={https://arxiv.org/abs/1904.09751}
}

@article{gong2025knowledge,
  title    = {Knowledge-Practice Performance Gap in Clinical Large Language Models: Systematic Review of 39 Benchmarks},
  author   = {Gong, Eun Jeong and Bang, Chang Seok and Lee, Jae Jun and Baik, Gwang Ho},
  journal  = {Journal of Medical Internet Research},
  volume   = {27},
  pages    = {e84120},
  year     = {2025},
  publisher= {JMIR Publications Inc.},
  doi      = {10.2196/84120},
  issn     = {1438-8871}
}

@article{naveed2025comprehensive,
  title    = {A Comprehensive Overview of Large Language Models},
  author   = {Naveed, Humza and Khan, Asad Ullah and Qiu, Shi and Saqib, Muhammad and Anwar, Saeed and Usman, Muhammad and Akhtar, Naveed and Barnes, Nick and Mian, Ajmal},
  journal  = {ACM Transactions on Intelligent Systems and Technology},
  volume   = {16},
  number   = {5},
  pages    = {1--72},
  year     = {2025},
  publisher= {Association for Computing Machinery (ACM)},
  doi      = {10.1145/3744746},
  issn     = {2157-6912}
}

@article{shi2025continual,
  title    = {Continual Learning of Large Language Models: A Comprehensive Survey},
  author   = {Shi, Haizhou and Xu, Zihao and Wang, Hengyi and Qin, Weiyi and Wang, Wenyuan and Wang, Yibin and Wang, Zifeng and Ebrahimi, Sayna and Wang, Hao},
  journal  = {ACM Computing Surveys},
  volume   = {58},
  number   = {5},
  pages    = {1--42},
  year     = {2025},
  publisher= {Association for Computing Machinery (ACM)},
  doi      = {10.1145/3735633},
  issn     = {1557-7341}
}

@inproceedings{chen2025towards,
    title = "Towards Effective and Efficient Continual Pre-training of Large Language Models",
    author = "Chen, Jie  and
      Chen, Zhipeng  and
      Wang, Jiapeng  and
      Zhou, Kun  and
      Zhu, Yutao  and
      Jiang, Jinhao  and
      Min, Yingqian  and
      Zhao, Wayne Xin  and
      Dou, Zhicheng  and
      Mao, Jiaxin  and
      Lin, Yankai  and
      Song, Ruihua  and
      Xu, Jun  and
      Chen, Xu  and
      Yan, Rui  and
      Wei, Zhewei  and
      Hu, Di  and
      Huang, Wenbing  and
      Wen, Ji-Rong",
    editor = "Che, Wanxiang  and
      Nabende, Joyce  and
      Shutova, Ekaterina  and
      Pilehvar, Mohammad Taher",
    booktitle = "Proceedings of the 63rd Annual Meeting of the {A}ssociation for {C}omputational {L}inguistics (Volume 1: Long Papers)",
    month = jul,
    year = "2025",
    address = "Vienna, Austria",
    publisher = "Association for Computational Linguistics",
    url = "https://aclanthology.org/2025.acl-long.289/",
    doi = "10.18653/v1/2025.acl-long.289",
    pages = "5779--5795",
    ISBN = "979-8-89176-251-0"
}

@inproceedings{christophe2024beyond,
    title = "Beyond Fine-tuning: Unleashing the Potential of Continuous Pretraining for Clinical {LLM}s",
    author = "Christophe, Clement  and
      Raha, Tathagata  and
      Maslenkova, Svetlana  and
      Salman, Muhammad Umar  and
      Kanithi, Praveenkumar  and
      Pimentel, Marco AF  and
      Khan, Shadab",
    editor = "Al-Onaizan, Yaser  and
      Bansal, Mohit  and
      Chen, Yun-Nung",
    booktitle = "Findings of the {A}ssociation for {C}omputational {L}inguistics: {EMNLP} 2024",
    month = nov,
    year = "2024",
    address = "Miami, Florida, USA",
    publisher = "Association for Computational Linguistics",
    url = "https://aclanthology.org/2024.findings-emnlp.618/",
    doi = "10.18653/v1/2024.findings-emnlp.618",
    pages = "10549--10561"
}

@inproceedings{guo2025efficient,
    title = "Efficient Domain Continual pretraining by Mitigating the Stability Gap",
    author = "Guo, Yiduo  and
      Fu, Jie  and
      Zhang, Huishuai  and
      Zhao, Dongyan",
    editor = "Che, Wanxiang  and
      Nabende, Joyce  and
      Shutova, Ekaterina  and
      Pilehvar, Mohammad Taher",
    booktitle = "Proceedings of the 63rd Annual Meeting of the {A}ssociation for {C}omputational {L}inguistics (Volume 1: Long Papers)",
    month = jul,
    year = "2025",
    address = "Vienna, Austria",
    publisher = "Association for Computational Linguistics",
    url = "https://aclanthology.org/2025.acl-long.1578/",
    doi = "10.18653/v1/2025.acl-long.1578",
    pages = "32850--32870",
    ISBN = "979-8-89176-251-0"
}

@article{singhal2025toward,
  title    = {Toward expert-level medical question answering with large language models},
  author   = {Singhal, Karan and Tu, Tao and Gottweis, Juraj and Sayres, Rory and Wulczyn, Ellery and Amin, Mohamed and Hou, Le and Clark, Kevin and Pfohl, Stephen R. and Cole-Lewis, Heather and Neal, Darlene and Rashid, Qazi Mamunur and Schaekermann, Mike and Wang, Amy and Dash, Dev and Chen, Jonathan H. and Shah, Nigam H. and Lachgar, Sami and Mansfield, Philip Andrew and Prakash, Sushant and Green, Bradley and Dominowska, Ewa and Agüera y Arcas, Blaise and Tomašev, Nenad and Liu, Yun and Wong, Renee and Semturs, Christopher and Mahdavi, S. Sara and Barral, Joelle K. and Webster, Dale R. and Corrado, Greg S. and Matias, Yossi and Azizi, Shekoofeh and Karthikesalingam, Alan and Natarajan, Vivek},
  journal  = {Nature Medicine},
  volume   = {31},
  number   = {3},
  pages    = {943--950},
  year     = {2025},
  publisher= {Springer Science and Business Media LLC},
  doi      = {10.1038/s41591-024-03423-7},
  issn     = {1546-170X}
}

@inproceedings{zheng2023judging,
 author = {Zheng, Lianmin and Chiang, Wei-Lin and Sheng, Ying and Zhuang, Siyuan and Wu, Zhanghao and Zhuang, Yonghao and Lin, Zi and Li, Zhuohan and Li, Dacheng and Xing, Eric and Zhang, Hao and Gonzalez, Joseph and Stoica, Ion},
 booktitle = {Advances in Neural Information Processing Systems},
 doi = {10.52202/075280-2020},
 editor = {A. Oh and T. Naumann and A. Globerson and K. Saenko and M. Hardt and S. Levine},
 pages = {46595--46623},
 publisher = {Curran Associates, Inc.},
 title = {Judging LLM-as-a-Judge with MT-Bench and Chatbot Arena},
 url = {https://proceedings.neurips.cc/paper_files/paper/2023/file/91f18a1287b398d378ef22505bf41832-Paper-Datasets_and_Benchmarks.pdf},
 volume = {36},
 year = {2023}
}

@article{davidson1970extending,
  title    = {On Extending the Bradley-Terry Model to Accommodate Ties in Paired Comparison Experiments},
  author   = {Davidson, Roger R.},
  journal  = {Journal of the American Statistical Association},
  volume   = {65},
  number   = {329},
  pages    = {317--328},
  year     = {1970},
  publisher= {Informa UK Limited},
  doi      = {10.1080/01621459.1970.10481082},
  issn     = {1537-274X}
}

@article{bradley1952rank,
  title    = {Rank Analysis of Incomplete Block Designs: I. The Method of Paired Comparisons},
  author   = {Bradley, Ralph Allan and Terry, Milton E.},
  journal  = {Biometrika},
  volume   = {39},
  number   = {3/4},
  pages    = {324--345},
  year     = {1952},
  doi      = {10.2307/2334029},
  issn     = {0006-3444}
}

@inproceedings{penedo2025fineweb2,
title={FineWeb2: One Pipeline to Scale Them All {\textemdash} Adapting Pre-Training Data Processing to Every Language},
author={Guilherme Penedo and Hynek Kydl{\'\i}{\v{c}}ek and Vinko Sabol{\v{c}}ec and Bettina Messmer and Negar Foroutan and Amir Hossein Kargaran and Colin Raffel and Martin Jaggi and Leandro Von Werra and Thomas Wolf},
booktitle={Second Conference on Language Modeling},
year={2025},
url={https://openreview.net/forum?id=jnRBe6zatP}
}

@article{zhang2023alpacare,
  title={Alpacare: Instruction-tuned large language models for medical application},
  author={Zhang, Xinlu and Tian, Chenxin and Yang, Xianjun and Chen, Lichang and Li, Zekun and Petzold, Linda Ruth},
  journal={arXiv preprint arXiv:2310.14558},
  year={2023}
}

@article{zakka2024almanac,
  title    = {Almanac {\textemdash} Retrieval-Augmented Language Models for Clinical Medicine},
  author   = {Zakka, Cyril and Shad, Rohan and Chaurasia, Akash and Dalal, Alex R. and Kim, Jennifer L. and Moor, Michael and Fong, Robyn and Phillips, Curran and Alexander, Kevin and Ashley, Euan and Boyd, Jack and Boyd, Kathleen and Hirsch, Karen and Langlotz, Curt and Lee, Rita and Melia, Joanna and Nelson, Joanna and Sallam, Karim and Tullis, Stacey and Vogelsong, Melissa Ann and Cunningham, John Patrick and Hiesinger, William},
  journal  = {NEJM AI},
  volume   = {1},
  number   = {2},
  year     = {2024},
  publisher= {Massachusetts Medical Society},
  doi      = {10.1056/aioa2300068},
  issn     = {2836-9386}
}

@inproceedings{xiong-etal-2024-benchmarking,
  title = "Benchmarking Retrieval-Augmented Generation for Medicine",
  author = "Xiong, Guangzhi and Jin, Qiao and Lu, Zhiyong and Zhang, Aidong",
  editor = "Ku, Lun-Wei and Martins, Andre and Srikumar, Vivek",
  booktitle = "Findings of the Association for Computational Linguistics: ACL 2024",
  month = aug,
  year = "2024",
  address = "Bangkok, Thailand",
  publisher = "Association for Computational Linguistics",
  url = "https://aclanthology.org/2024.findings-acl.372/",
  doi = "10.18653/v1/2024.findings-acl.372",
  pages = "6233--6251"
}

@inproceedings{wu2025medreason,
    title = "{M}ed{R}eason: Eliciting Factual Medical Reasoning Steps in {LLM}s via Knowledge Graphs",
    author = "Wu, Juncheng  and
      Deng, Wenlong  and
      Li, Xingxuan  and
      Liu, Sheng  and
      Mi, Taomian  and
      Peng, Yifan  and
      Xu, Ziyang  and
      Liu, Yi  and
      Cho, Hyunjin  and
      Choi, Chang-In  and
      Cao, Yihan  and
      Ren, Hui  and
      Li, Xiang  and
      Li, Xiaoxiao  and
      Zhou, Yuyin",
    editor = "Wang, Yimu  and
      Choi, Yee Man  and
      Wu, Di  and
      Ouyang, Siqi  and
      Azadani, Mozhgan Nasr  and
      Fung, Yi R. (May)",
    booktitle = "Proceedings of the 1st Workshop on Grounding Language Models: Learning Faithfully and Efficiently ({G}round{LM} 2026)",
    month = oct,
    year = "2026",
    address = "Budapest, Hungary",
    publisher = "Association for Computational Linguistics",
    url = "https://aclanthology.org/2026.groundlm-1.1/",
    doi = "10.18653/2026.groundlm-1.1",
    pages = "1--17",
    ISBN = "979-8-89176-478-1"
}

@InProceedings{zhuang2025efficient,
    author    = {Zhuang, Xinlin and Tang, Feilong and Yang, Haolin and Liu, Xiwei and Hu, Ming and Li, Huifa and Xue, Haochen and He, Junjun and Ge, Zongyuan and Li, Yichen and Qian, Ying and Razzak, Imran},
    title     = {Towards Efficient Medical Reasoning with Minimal Fine-Tuning Data},
    booktitle = {Proceedings of the IEEE/CVF Conference on Computer Vision and Pattern Recognition (CVPR)},
    month     = {June},
    year      = {2026},
    pages     = {20223-20232}
}

@misc{arora2025healthbench,
  title={HealthBench: Evaluating Large Language Models Towards Improved Human Health},
  author={Arora, Rahul K. and Wei, Jason and Soskin Hicks, Rebecca and Bowman, Preston and Quiñonero-Candela, Joaquin and Tsimpourlas, Foivos and Sharman, Michael and Shah, Meghan and Vallone, Andrea and Beutel, Alex and Heidecke, Johannes and Singhal, Karan},
  year={2025},
  eprint={2505.08775},
  archivePrefix={arXiv},
  primaryClass={cs.CL},
  url={https://arxiv.org/abs/2505.08775}
}

@inproceedings{vllm,
  title    = {Efficient Memory Management for Large Language Model Serving with PagedAttention},
  author   = {Kwon, Woosuk and Li, Zhuohan and Zhuang, Siyuan and Sheng, Ying and Zheng, Lianmin and Yu, Cody Hao and Gonzalez, Joseph and Zhang, Hao and Stoica, Ion},
  booktitle= {Proceedings of the 29th Symposium on Operating Systems Principles (SOSP '23)},
  pages    = {611--626},
  year     = {2023},
  publisher= {ACM},
  doi      = {10.1145/3600006.3613165}
}

@article{Lu2025,
  author  = {Lu, Wei and Luu, Rachel K. and Buehler, Markus J.},
  title   = {Fine-tuning large language models for domain adaptation: exploration of training strategies, scaling, model merging and synergistic capabilities},
  journal = {npj Computational Materials},
  year    = {2025},
  volume  = {11},
  number  = {1},
  pages   = {84},
  doi     = {10.1038/s41524-025-01564-y},
  url     = {https://doi.org/10.1038/s41524-025-01564-y},
  issn    = {2057-3960}
}

@inproceedings{dubois2025lengthcontrolledalpacaevalsimpleway,
title={Length-Controlled AlpacaEval: A Simple Debiasing of Automatic Evaluators},
author={Yann Dubois and Percy Liang and Tatsunori Hashimoto},
booktitle={First Conference on Language Modeling},
year={2024},
url={https://openreview.net/forum?id=CybBmzWBX0}
}

@inproceedings{doll-etal-2026-continual,
    title = "Can Continual Pretraining Bridge the Performance Gap between General-purpose and Specialized Language Models in the Medical Domain?",
    author = "Doll, Niclas  and
      Buschhoff, Jasper Schulze  and
      Satheesh, Shalaka  and
      Abdelwahab, Hammam  and
      Allende-Cid, H{\'e}ctor  and
      Klug, Katrin",
    editor = "Liakata, Maria  and
      Moreira, Viviane P.  and
      Zhang, Jiajun  and
      Jurgens, David",
    booktitle = "Proceedings of the 64th Annual Meeting of the {A}ssociation for {C}omputational {L}inguistics (Volume 1: Long Papers)",
    month = jul,
    year = "2026",
    address = "San Diego, California, United States",
    publisher = "Association for Computational Linguistics",
    url = "https://aclanthology.org/2026.acl-long.17/",
    doi = "10.18653/v1/2026.acl-long.17",
    pages = "427--444",
    ISBN = "979-8-89176-390-6"
}

@misc{xu2024benchmarkdatacontaminationlarge,
      title={Benchmark Data Contamination of Large Language Models: A Survey}, 
      author={Cheng Xu and Shuhao Guan and Derek Greene and M-Tahar Kechadi},
      year={2024},
      eprint={2406.04244},
      archivePrefix={arXiv},
      primaryClass={cs.CL},
      url={https://arxiv.org/abs/2406.04244}, 
}

@inproceedings{lin2024the,
  title={The Unlocking Spell on Base {LLM}s:  Rethinking Alignment via In-Context Learning},
  author={Bill Yuchen Lin and Abhilasha Ravichander and Ximing Lu and Nouha Dziri and Melanie Sclar and Khyathi Chandu and Chandra Bhagavatula and Yejin Choi},
  booktitle={The Twelfth International Conference on Learning Representations},
  year={2024},
  url={https://openreview.net/forum?id=wxJ0eXwwda}
}

@article{lin1991divergence,
  title    = {Divergence measures based on the Shannon entropy},
  author   = {Lin, Jianhua},
  journal  = {IEEE Transactions on Information Theory},
  volume   = {37},
  number   = {1},
  pages    = {145--151},
  year     = {1991},
  publisher= {Institute of Electrical and Electronics Engineers (IEEE)},
  doi      = {10.1109/18.61115},
  issn     = {0018-9448}
}

@article{kullback1951information,
  title    = {On Information and Sufficiency},
  author   = {Kullback, Solomon and Leibler, Richard A.},
  journal  = {The Annals of Mathematical Statistics},
  volume   = {22},
  number   = {1},
  pages    = {79--86},
  year     = {1951},
  publisher= {Institute of Mathematical Statistics},
  doi      = {10.1214/aoms/1177729694},
  issn     = {0003-4851}
}

@inproceedings{li-etal-2024-multiple,
    title = "Can Multiple-choice Questions Really Be Useful in Detecting the Abilities of {LLM}s?",
    author = "Li, Wangyue and Li, Liangzhi and Xiang, Tong and Liu, Xiao and Deng, Wei and Garcia, Noa",
    booktitle = "Proceedings of the 2024 Joint International Conference on Computational Linguistics, Language Resources and Evaluation (LREC-COLING 2024)",
    year = "2024",
    address = "Torino, Italia",
    publisher = "ELRA and ICCL",
    url = "https://aclanthology.org/2024.lrec-main.251",
    pages = "2819--2834"
}

@inproceedings{lyu-etal-2024-beyond,
    title = "Beyond Probabilities: Unveiling the Misalignment in Evaluating Large Language Models",
    author = "Lyu, Chenyang and Wu, Minghao and Aji, Alham Fikri",
    booktitle = "Proceedings of the 1st Workshop on Towards Knowledgeable Language Models (KnowLLM 2024)",
    year = "2024",
    address = "Bangkok, Thailand",
    publisher = "Association for Computational Linguistics",
    url = "https://aclanthology.org/2024.knowllm-1.10",
    doi = "10.18653/v1/2024.knowllm-1.10",
    pages = "109--131"
}

@inproceedings{wang-etal-2024-answer,
    title = "{``}My Answer is {C}{''}: First-Token Probabilities Do Not Match Text Answers in Instruction-Tuned Language Models",
    author = "Wang, Xinpeng and Ma, Bolei and Hu, Chengzhi and Weber-Genzel, Leon and R{\"o}ttger, Paul and Kreuter, Frauke and Hovy, Dirk and Plank, Barbara",
    booktitle = "Findings of the Association for Computational Linguistics: ACL 2024",
    year = "2024",
    address = "Bangkok, Thailand",
    publisher = "Association for Computational Linguistics",
    url = "https://aclanthology.org/2024.findings-acl.441",
    doi = "10.18653/v1/2024.findings-acl.441",
    pages = "7407--7416"
}

@misc{grattafiori2024llama3herdmodels,
      title={The {Llama} 3 Herd of Models},
      author={Aaron Grattafiori and Abhimanyu Dubey and Abhinav Jauhri and Abhinav Pandey and Abhishek Kadian and Ahmad Al-Dahle and Aiesha Letman and Akhil Mathur and Alan Schelten and Alex Vaughan and Amy Yang and Angela Fan and Anirudh Goyal and Anthony Hartshorn and Aobo Yang and Archi Mitra and Archie Sravankumar and Artem Korenev and Arthur Hinsvark and Arun Rao and Aston Zhang and Aurelien Rodriguez and Austen Gregerson and Ava Spataru and Baptiste Roziere and Bethany Biron and Binh Tang and Bobbie Chern and Charlotte Caucheteux and Chaya Nayak and Chloe Bi and Chris Marra and Chris McConnell and Christian Keller and Christophe Touret and Chunyang Wu and Corinne Wong and Cristian Canton Ferrer and Cyrus Nikolaidis and Damien Allonsius and Daniel Song and Danielle Pintz and Danny Livshits and Danny Wyatt and David Esiobu and Dhruv Choudhary and Dhruv Mahajan and Diego Garcia-Olano and Diego Perino and Dieuwke Hupkes and Egor Lakomkin and Ehab AlBadawy and Elina Lobanova and Emily Dinan and Eric Michael Smith and Filip Radenovic and Francisco Guzmán and Frank Zhang and Gabriel Synnaeve and Gabrielle Lee and Georgia Lewis Anderson and Govind Thattai and Graeme Nail and Gregoire Mialon and Guan Pang and Guillem Cucurell and Hailey Nguyen and Hannah Korevaar and Hu Xu and Hugo Touvron and Iliyan Zarov and Imanol Arrieta Ibarra and Isabel Kloumann and Ishan Misra and Ivan Evtimov and Jack Zhang and Jade Copet and Jaewon Lee and Jan Geffert and Jana Vranes and Jason Park and Jay Mahadeokar and Jeet Shah and Jelmer van der Linde and Jennifer Billock and Jenny Hong and Jenya Lee and Jeremy Fu and Jianfeng Chi and Jianyu Huang and Jiawen Liu and Jie Wang and Jiecao Yu and Joanna Bitton and Joe Spisak and Jongsoo Park and Joseph Rocca and Joshua Johnstun and Joshua Saxe and Junteng Jia and Kalyan Vasuden Alwala and Karthik Prasad and Kartikeya Upasani and Kate Plawiak and Ke Li and Kenneth Heafield and Kevin Stone and Khalid El-Arini and Krithika Iyer and Kshitiz Malik and Kuenley Chiu and Kunal Bhalla and Kushal Lakhotia and Lauren Rantala-Yeary and Laurens van der Maaten and Lawrence Chen and Liang Tan and Liz Jenkins and Louis Martin and Lovish Madaan and Lubo Malo and Lukas Blecher and Lukas Landzaat and Luke de Oliveira and Madeline Muzzi and Mahesh Pasupuleti and Mannat Singh and Manohar Paluri and Marcin Kardas and Maria Tsimpoukelli and Mathew Oldham and Mathieu Rita and Maya Pavlova and Melanie Kambadur and Mike Lewis and Min Si and Mitesh Kumar Singh and Mona Hassan and Naman Goyal and Narjes Torabi and Nikolay Bashlykov and Nikolay Bogoychev and Niladri Chatterji and Ning Zhang and Olivier Duchenne and Onur Çelebi and Patrick Alrassy and Pengchuan Zhang and Pengwei Li and Petar Vasic and Peter Weng and Prajjwal Bhargava and Pratik Dubal and Praveen Krishnan and Punit Singh Koura and Puxin Xu and Qing He and Qingxiao Dong and Ragavan Srinivasan and Raj Ganapathy and Ramon Calderer and Ricardo Silveira Cabral and Robert Stojnic and Roberta Raileanu and Rohan Maheswari and Rohit Girdhar and Rohit Patel and Romain Sauvestre and Ronnie Polidoro and Roshan Sumbaly and Ross Taylor and Ruan Silva and Rui Hou and Rui Wang and Saghar Hosseini and Sahana Chennabasappa and Sanjay Singh and Sean Bell and Seohyun Sonia Kim and Sergey Edunov and Shaoliang Nie and Sharan Narang and Sharath Raparthy and Sheng Shen and Shengye Wan and Shruti Bhosale and Shun Zhang and Simon Vandenhende and Soumya Batra and Spencer Whitman and Sten Sootla and Stephane Collot and Suchin Gururangan and Sydney Borodinsky and Tamar Herman and Tara Fowler and Tarek Sheasha and Thomas Georgiou and Thomas Scialom and Tobias Speckbacher and Todor Mihaylov and Tong Xiao and Ujjwal Karn and Vedanuj Goswami and Vibhor Gupta and Vignesh Ramanathan and Viktor Kerkez and Vincent Gonguet and Virginie Do and Vish Vogeti and Vítor Albiero and Vladan Petrovic and Weiwei Chu and Wenhan Xiong and Wenyin Fu and Whitney Meers and Xavier Martinet and Xiaodong Wang and Xiaofang Wang and Xiaoqing Ellen Tan and Xide Xia and Xinfeng Xie and Xuchao Jia and Xuewei Wang and Yaelle Goldschlag and Yashesh Gaur and Yasmine Babaei and Yi Wen and Yiwen Song and Yuchen Zhang and Yue Li and Yuning Mao and Zacharie Delpierre Coudert and Zheng Yan and Zhengxing Chen and Zoe Papakipos and Aaditya Singh and Aayushi Srivastava and Abha Jain and Adam Kelsey and Adam Shajnfeld and Adithya Gangidi and Adolfo Victoria and Ahuva Goldstand and Ajay Menon and Ajay Sharma and Alex Boesenberg and Alexei Baevski and Allie Feinstein and Amanda Kallet and Amit Sangani and Amos Teo and Anam Yunus and Andrei Lupu and Andres Alvarado and Andrew Caples and Andrew Gu and Andrew Ho and Andrew Poulton and Andrew Ryan and Ankit Ramchandani and Annie Dong and Annie Franco and Anuj Goyal and Aparajita Saraf and Arkabandhu Chowdhury and Ashley Gabriel and Ashwin Bharambe and Assaf Eisenman and Azadeh Yazdan and Beau James and Ben Maurer and Benjamin Leonhardi and Bernie Huang and Beth Loyd and Beto De Paola and Bhargavi Paranjape and Bing Liu and Bo Wu and Boyu Ni and Braden Hancock and Bram Wasti and Brandon Spence and Brani Stojkovic and Brian Gamido and Britt Montalvo and Carl Parker and Carly Burton and Catalina Mejia and Ce Liu and Changhan Wang and Changkyu Kim and Chao Zhou and Chester Hu and Ching-Hsiang Chu and Chris Cai and Chris Tindal and Christoph Feichtenhofer and Cynthia Gao and Damon Civin and Dana Beaty and Daniel Kreymer and Daniel Li and David Adkins and David Xu and Davide Testuggine and Delia David and Devi Parikh and Diana Liskovich and Didem Foss and Dingkang Wang and Duc Le and Dustin Holland and Edward Dowling and Eissa Jamil and Elaine Montgomery and Eleonora Presani and Emily Hahn and Emily Wood and Eric-Tuan Le and Erik Brinkman and Esteban Arcaute and Evan Dunbar and Evan Smothers and Fei Sun and Felix Kreuk and Feng Tian and Filippos Kokkinos and Firat Ozgenel and Francesco Caggioni and Frank Kanayet and Frank Seide and Gabriela Medina Florez and Gabriella Schwarz and Gada Badeer and Georgia Swee and Gil Halpern and Grant Herman and Grigory Sizov and Guangyi and Zhang and Guna Lakshminarayanan and Hakan Inan and Hamid Shojanazeri and Han Zou and Hannah Wang and Hanwen Zha and Haroun Habeeb and Harrison Rudolph and Helen Suk and Henry Aspegren and Hunter Goldman and Hongyuan Zhan and Ibrahim Damlaj and Igor Molybog and Igor Tufanov and Ilias Leontiadis and Irina-Elena Veliche and Itai Gat and Jake Weissman and James Geboski and James Kohli and Janice Lam and Japhet Asher and Jean-Baptiste Gaya and Jeff Marcus and Jeff Tang and Jennifer Chan and Jenny Zhen and Jeremy Reizenstein and Jeremy Teboul and Jessica Zhong and Jian Jin and Jingyi Yang and Joe Cummings and Jon Carvill and Jon Shepard and Jonathan McPhie and Jonathan Torres and Josh Ginsburg and Junjie Wang and Kai Wu and Kam Hou U and Karan Saxena and Kartikay Khandelwal and Katayoun Zand and Kathy Matosich and Kaushik Veeraraghavan and Kelly Michelena and Keqian Li and Kiran Jagadeesh and Kun Huang and Kunal Chawla and Kyle Huang and Lailin Chen and Lakshya Garg and Lavender A and Leandro Silva and Lee Bell and Lei Zhang and Liangpeng Guo and Licheng Yu and Liron Moshkovich and Luca Wehrstedt and Madian Khabsa and Manav Avalani and Manish Bhatt and Martynas Mankus and Matan Hasson and Matthew Lennie and Matthias Reso and Maxim Groshev and Maxim Naumov and Maya Lathi and Meghan Keneally and Miao Liu and Michael L. Seltzer and Michal Valko and Michelle Restrepo and Mihir Patel and Mik Vyatskov and Mikayel Samvelyan and Mike Clark and Mike Macey and Mike Wang and Miquel Jubert Hermoso and Mo Metanat and Mohammad Rastegari and Munish Bansal and Nandhini Santhanam and Natascha Parks and Natasha White and Navyata Bawa and Nayan Singhal and Nick Egebo and Nicolas Usunier and Nikhil Mehta and Nikolay Pavlovich Laptev and Ning Dong and Norman Cheng and Oleg Chernoguz and Olivia Hart and Omkar Salpekar and Ozlem Kalinli and Parkin Kent and Parth Parekh and Paul Saab and Pavan Balaji and Pedro Rittner and Philip Bontrager and Pierre Roux and Piotr Dollar and Polina Zvyagina and Prashant Ratanchandani and Pritish Yuvraj and Qian Liang and Rachad Alao and Rachel Rodriguez and Rafi Ayub and Raghotham Murthy and Raghu Nayani and Rahul Mitra and Rangaprabhu Parthasarathy and Raymond Li and Rebekkah Hogan and Robin Battey and Rocky Wang and Russ Howes and Ruty Rinott and Sachin Mehta and Sachin Siby and Sai Jayesh Bondu and Samyak Datta and Sara Chugh and Sara Hunt and Sargun Dhillon and Sasha Sidorov and Satadru Pan and Saurabh Mahajan and Saurabh Verma and Seiji Yamamoto and Sharadh Ramaswamy and Shaun Lindsay and Shaun Lindsay and Sheng Feng and Shenghao Lin and Shengxin Cindy Zha and Shishir Patil and Shiva Shankar and Shuqiang Zhang and Shuqiang Zhang and Sinong Wang and Sneha Agarwal and Soji Sajuyigbe and Soumith Chintala and Stephanie Max and Stephen Chen and Steve Kehoe and Steve Satterfield and Sudarshan Govindaprasad and Sumit Gupta and Summer Deng and Sungmin Cho and Sunny Virk and Suraj Subramanian and Sy Choudhury and Sydney Goldman and Tal Remez and Tamar Glaser and Tamara Best and Thilo Koehler and Thomas Robinson and Tianhe Li and Tianjun Zhang and Tim Matthews and Timothy Chou and Tzook Shaked and Varun Vontimitta and Victoria Ajayi and Victoria Montanez and Vijai Mohan and Vinay Satish Kumar and Vishal Mangla and Vlad Ionescu and Vlad Poenaru and Vlad Tiberiu Mihailescu and Vladimir Ivanov and Wei Li and Wenchen Wang and Wenwen Jiang and Wes Bouaziz and Will Constable and Xiaocheng Tang and Xiaojian Wu and Xiaolan Wang and Xilun Wu and Xinbo Gao and Yaniv Kleinman and Yanjun Chen and Ye Hu and Ye Jia and Ye Qi and Yenda Li and Yilin Zhang and Ying Zhang and Yossi Adi and Youngjin Nam and Yu and Wang and Yu Zhao and Yuchen Hao and Yundi Qian and Yunlu Li and Yuzi He and Zach Rait and Zachary DeVito and Zef Rosnbrick and Zhaoduo Wen and Zhenyu Yang and Zhiwei Zhao and Zhiyu Ma},
      year={2024},
      eprint={2407.21783},
      archivePrefix={arXiv},
      primaryClass={cs.AI},
      url={https://arxiv.org/abs/2407.21783}
}

@misc{mistral2024ministraux,
  author       = {{Mistral AI}},
  title        = {Un {Ministral}, des {Ministraux}},
  year         = {2024},
  howpublished = {\url{https://mistral.ai/news/ministraux}},
  note         = {Accessed: 2026-11-19}
}

@inproceedings{biderman2023pythia,
  title     = {Pythia: A Suite for Analyzing Large Language Models Across Training and Scaling},
  author    = {Biderman, Stella and Schoelkopf, Hailey and Anthony, Quentin Gregory and Bradley, Herbie and O'Brien, Kyle and Hallahan, Eric and Khan, Mohammad Aflah and Purohit, Shivanshu and Prashanth, USVSN Sai and Raff, Edward and Skowron, Aviya and Sutawika, Lintang and van der Wal, Oskar},
  booktitle = {Proceedings of the 40th International Conference on Machine Learning},
  series    = {Proceedings of Machine Learning Research},
  volume    = {202},
  pages     = {2397--2430},
  publisher = {PMLR},
  year      = {2023},
  url       = {https://proceedings.mlr.press/v202/biderman23a.html}
}

@misc{yang2025qwen3technicalreport,
      title={{Qwen3} Technical Report},
      author={An Yang and Anfeng Li and Baosong Yang and Beichen Zhang and Binyuan Hui and Bo Zheng and Bowen Yu and Chang Gao and Chengen Huang and Chenxu Lv and Chujie Zheng and Dayiheng Liu and Fan Zhou and Fei Huang and Feng Hu and Hao Ge and Haoran Wei and Huan Lin and Jialong Tang and Jian Yang and Jianhong Tu and Jianwei Zhang and Jianxin Yang and Jiaxi Yang and Jing Zhou and Jingren Zhou and Junyang Lin and Kai Dang and Keqin Bao and Kexin Yang and Le Yu and Lianghao Deng and Mei Li and Mingfeng Xue and Mingze Li and Pei Zhang and Peng Wang and Qin Zhu and Rui Men and Ruize Gao and Shixuan Liu and Shuang Luo and Tianhao Li and Tianyi Tang and Wenbiao Yin and Xingzhang Ren and Xinyu Wang and Xinyu Zhang and Xuancheng Ren and Yang Fan and Yang Su and Yichang Zhang and Yinger Zhang and Yu Wan and Yuqiong Liu and Zekun Wang and Zeyu Cui and Zhenru Zhang and Zhipeng Zhou and Zihan Qiu},
      year={2025},
      eprint={2505.09388},
      archivePrefix={arXiv},
      primaryClass={cs.CL},
      url={https://arxiv.org/abs/2505.09388}
}

@inproceedings{lee2024rlaif,
  title     = {{RLAIF} vs. {RLHF}: Scaling Reinforcement Learning from Human Feedback with {AI} Feedback},
  author    = {Lee, Harrison and Phatale, Samrat and Mansoor, Hassan and Mesnard, Thomas and Ferret, Johan and Lu, Kellie Ren and Bishop, Colton and Hall, Ethan and Carbune, Victor and Rastogi, Abhinav and Prakash, Sushant},
  booktitle = {Proceedings of the 41st International Conference on Machine Learning},
  series    = {Proceedings of Machine Learning Research},
  volume    = {235},
  pages     = {26874--26901},
  publisher = {PMLR},
  year      = {2024},
  url       = {https://proceedings.mlr.press/v235/lee24t.html}
}

@inproceedings{gunjal2025rubrics,
  title     = {Rubrics as Rewards: Reinforcement Learning Beyond Verifiable Domains},
  author    = {Gunjal, Anisha and Wang, Anthony and Lau, Elaine and Nath, Vaskar and He, Yunzhong and Liu, Bing and Hendryx, Sean},
  booktitle = {The Fourteenth International Conference on Learning Representations},
  year      = {2026},
  url       = {https://arxiv.org/abs/2507.17746}
}

@inproceedings{feuer2025style,
  title     = {Style Outweighs Substance: Failure Modes of {LLM} Judges in Alignment Benchmarking},
  author    = {Feuer, Benjamin and Goldblum, Micah and Datta, Teresa and Nambiar, Sanjana and Besaleli, Raz and Dooley, Samuel and Cembalest, Max and Dickerson, John P.},
  booktitle = {The Thirteenth International Conference on Learning Representations},
  year      = {2025},
  url       = {https://arxiv.org/abs/2409.15268}
}

@misc{kenton2026debate,
  title         = {Debate Training Reduces Reward Hacking in {RLAIF}},
  author        = {Kenton, Zachary and Janzer, Lili and Greig, Rory and Teh, Tian Huey and Tyshchuk, Kirill and Brown-Cohen, Jonah and Edwards, Harri and Rajamanoharan, Senthooran and Siegel, Noah Y. and Jaques, Natasha and Shah, Rohin},
  year          = {2026},
  eprint        = {2608.17776},
  archivePrefix = {arXiv},
  primaryClass  = {cs.LG},
  url           = {https://arxiv.org/abs/2608.17776}
}

@article{thirunavukarasu2023large,
  title    = {Large language models in medicine},
  author   = {Thirunavukarasu, Arun James and Ting, Darren Shu Jeng and Elangovan, Kabilan and Gutierrez, Laura and Tan, Ting Fang and Ting, Daniel Shu Wei},
  journal  = {Nature Medicine},
  volume   = {29},
  number   = {8},
  pages    = {1930--1940},
  year     = {2023},
  publisher= {Springer Science and Business Media LLC},
  doi      = {10.1038/s41591-023-02448-8},
  issn     = {1546-170X}
}

@inproceedings{alexandrov2024mitigating,
    title = "Mitigating Catastrophic Forgetting in Language Transfer via Model Merging",
    author = {Alexandrov, Anton  and
      Raychev, Veselin  and
      M{\"u}ller, Mark Niklas  and
      Zhang, Ce  and
      Vechev, Martin  and
      Toutanova, Kristina},
    editor = "Al-Onaizan, Yaser  and
      Bansal, Mohit  and
      Chen, Yun-Nung",
    booktitle = "Findings of the {A}ssociation for {C}omputational {L}inguistics: {EMNLP} 2024",
    month = nov,
    year = "2024",
    address = "Miami, Florida, USA",
    publisher = "Association for Computational Linguistics",
    url = "https://aclanthology.org/2024.findings-emnlp.1000/",
    doi = "10.18653/v1/2024.findings-emnlp.1000",
    pages = "17167--17186"
}

@article{ibrahim2024simple,
title={Simple and Scalable Strategies to Continually Pre-train Large Language Models},
author={Adam Ibrahim and Benjamin Th{\'e}rien and Kshitij Gupta and Mats Leon Richter and Quentin Gregory Anthony and Eugene Belilovsky and Timoth{\'e}e Lesort and Irina Rish},
journal={Transactions on Machine Learning Research},
issn={2835-8856},
year={2024},
url={https://openreview.net/forum?id=DimPeeCxKO},
note={}
}

@inproceedings{yao2024crosslingual,
    title = "Data Contamination Can Cross Language Barriers",
    author = "Yao, Feng  and
      Zhuang, Yufan  and
      Sun, Zihao  and
      Xu, Sunan  and
      Kumar, Animesh  and
      Shang, Jingbo",
    editor = "Al-Onaizan, Yaser  and
      Bansal, Mohit  and
      Chen, Yun-Nung",
    booktitle = "Proceedings of the 2024 Conference on Empirical Methods in Natural Language Processing",
    month = nov,
    year = "2024",
    address = "Miami, Florida, USA",
    publisher = "Association for Computational Linguistics",
    url = "https://aclanthology.org/2024.emnlp-main.990/",
    doi = "10.18653/v1/2024.emnlp-main.990",
    pages = "17864--17875"
}

@misc{meta2024llama32,
  author       = {{Meta AI}},
  title        = {Llama 3.2: Revolutionizing edge AI and vision with open, customizable models},
  year         = {2024},
  month        = sep,
  howpublished = {\url{https://ai.meta.com/blog/llama-3-2-connect-2024-vision-edge-mobile-devices/}},
  note         = {Accessed: 2026-11-20}
}

@misc{qwen2024qwen25card,
  author       = {{Qwen Team}},
  title        = {Qwen2.5-7B-Instruct},
  year         = {2024},
  howpublished = {\url{https://huggingface.co/Qwen/Qwen2.5-7B-Instruct}},
  note         = {Hugging Face model card. Accessed: 2026-11-20}
}

@misc{pytorch2024fsdp2,
  author       = {{PyTorch Contributors}},
  title        = {torch.distributed.fsdp.fully\_shard},
  year         = {2024},
  howpublished = {\url{https://docs.pytorch.org/docs/stable/distributed.fsdp.fully_shard.html}},
  note         = {PyTorch documentation. Accessed: 2026-11-20}
}

@misc{gemmateam2025gemma3,
  title         = {Gemma 3 Technical Report},
  author        = {{Gemma Team}},
  year          = {2025},
  eprint        = {2503.19786},
  archivePrefix = {arXiv},
  primaryClass  = {cs.CL},
  url           = {https://arxiv.org/abs/2503.19786}
}

\newpage

\appendix
\renewcommand{\dbltopfraction}{0.9}
\renewcommand{\topfraction}{0.9}
\renewcommand{\textfraction}{0.1}
\renewcommand{\floatpagefraction}{0.75}
\renewcommand{\dblfloatpagefraction}{0.75}
\setcounter{dbltopnumber}{3}
\raggedbottom

\section{Training Configuration}

The domain adaptation process is implemented using the \textit{Axolotl} framework \cite{axolotl}. For the larger model variants, we utilize \textit{FSDP2} \cite{pytorch2024fsdp2}, the redesigned version of PyTorch Fully Sharded Data Parallel \cite{zhao2023pytorchfsdpexperiencesscaling}, to optimize memory distribution across the compute cluster. All models are trained for exactly one epoch using an \textit{AdamW} optimizer \cite{loshchilov2019decoupledweightdecayregularization} and a cosine learning rate schedule. To analyze the evolution of the adaptation process and map the performance trajectory, we save 18 intermediate checkpoints per experiment, distributed at logarithmic intervals throughout the training epoch.

To maintain consistency in our scaling analysis, we apply a uniform hyperparameter configuration across all experiments, as detailed in Table \ref{tab:hyperparameters}.

\begin{table}[!h]
    \centering
    \begin{tabular}{@{}ll@{}}
        \toprule
        \textbf{Hyperparameter} & \textbf{Value} \\
        \midrule
        Effective Batch Size    & 128            \\
        Learning Rate           & $2 \times 10^{-5}$ \\
        Weight Decay            & 0.1            \\
        Warmup Ratio            & 0.1            \\
        \bottomrule
    \end{tabular}
    \caption{Core training hyperparameters for domain adaptation across all model scales.}
    \label{tab:hyperparameters}
\end{table}

\section{SLERP Merge Configuration}
\label{sec:merge-config}

Each continually pre-trained checkpoint is merged with the original instruction-tuned source model using \textit{mergekit} \cite{goddard-etal-2024-arcees}. The interpolation coefficient $t$ is varied by layer type to balance domain knowledge retention in the attention layers with preservation of general reasoning in the feed-forward layers. The configuration below is applied uniformly across all model scales and families:

\begin{lstlisting}[
  basicstyle=\ttfamily\small,
  frame=single,
  breaklines=true
]
base_model: <source-instruct-model>
models:
  - model: <source-instruct-model>
  - model: <cpt-checkpoint>
merge_method: slerp
parameters:
  t:
    - filter: self_attn
      value: [0, 0.5, 0.3, 0.7, 1]
    - filter: mlp
      value: [1, 0.5, 0.7, 0.3, 0]
    - value: 0.5
dtype: bfloat16
\end{lstlisting}

The \texttt{self\_attn} filter applies a gradient from $0$ (fully source) to $1$ (fully adapted) across the interpolation steps for attention weights, while the \texttt{mlp} filter applies the inverse gradient for feed-forward weights. All remaining parameters (e.g., layer norms, embeddings) use a uniform interpolation of $t = 0.5$. This configuration was held constant across all experiments.

\section{Computational Resources}

The experiments conducted as part of this study involved a significant allocation of high-performance computing resources. The 18 checkpoints archived for each model scale resulted in a total storage footprint of 2.2 TB of model weights. To quantify the computational investment, we calculate the total Floating Point Operations (FLOPs) and cumulative GPU hours for each training run. Table \ref{tab:compute} provides a granular breakdown of these metrics, allowing for a direct comparison of the resources required to achieve specific performance thresholds across different parameter counts.

\begin{table*}[t]
    \centering
    \begin{tabular}{lcccc}
        \toprule
        \textbf{Model Family} & \textbf{Parameters} & \textbf{Training FLOPs} & \textbf{GPU Hours} & \textbf{GPU Model} \\
        \midrule
        Qwen 2.5 & $0.5B$ & $2.03 \cdot 10^{19}$ & $3.63h \times 8$ & NVIDIA H100 (80GB) \\
        Qwen 2.5 & $1.5B$ & $7.42 \cdot 10^{19}$ & $8.40h \times 8$ & NVIDIA H100 (80GB) \\
        Qwen 2.5 & $3B$ & $1.57 \cdot 10^{20}$ & $15.71h \times 8$ & NVIDIA H100 (80GB) \\
        Qwen 2.5 & $7B$ & $4.00 \cdot 10^{20}$ & $31.35h \times 8$ & NVIDIA H100 (80GB) \\
        Qwen 3 & $0.6B$ & $2.49 \cdot 10^{19}$ & $27.31h \times 8$ & AMD MI250X  \\
        Qwen 3 & $1.7B$ & $7.98 \cdot 10^{19}$ & $17.65h \times 32$ & AMD MI250X  \\
        Qwen 3 & $4B$ & $2.06 \cdot 10^{20}$ & $33.29h \times 32$ & AMD MI250X  \\
        Qwen 3 & $8B$ & $4.29 \cdot 10^{20}$ & $30.63h \times 64$ & AMD MI250X \\
        \bottomrule
    \end{tabular}
    \caption{Computational expenditure and hardware utilization for domain adaptation. The training load is quantified via total Floating Point Operations (FLOPs) and temporal resource usage. GPU hours are reported as wall-clock time $\times$ number of accelerators to reflect the parallel scale of each training run.}
    \label{tab:compute}
\end{table*}

\section{Judge Model Selection}
\label{sec:judge-selection}

The preference ranking evaluation requires a high-capacity judge model to perform pairwise comparisons across all checkpoints and model scales, resulting in a substantial number of evaluations. To balance annotation quality against cost, we conducted a preliminary agreement study comparing six candidate judge models, namely \textit{GPT-4.1-nano}, \textit{GPT-4.1-mini}, \textit{GPT-5}, \textit{GPT-5-mini}, \textit{GPT-5.2}, and \textit{GPT-5.4}. We randomly sampled $100$ response pairs from our evaluation set and collected pairwise judgments from each candidate model. In the absence of human expert annotations, we use the most capable available model, \textit{GPT-5.4}, as the reference annotator. Prior work has shown that strong frontier LLMs can serve as reliable automatic evaluators and exhibit high agreement with human judgments \cite{zheng2023judging}. Agreement between each candidate and this reference was quantified using Cohen's $\kappa$.

\begin{figure}[t]
    \centering
    \includegraphics[width=\columnwidth]{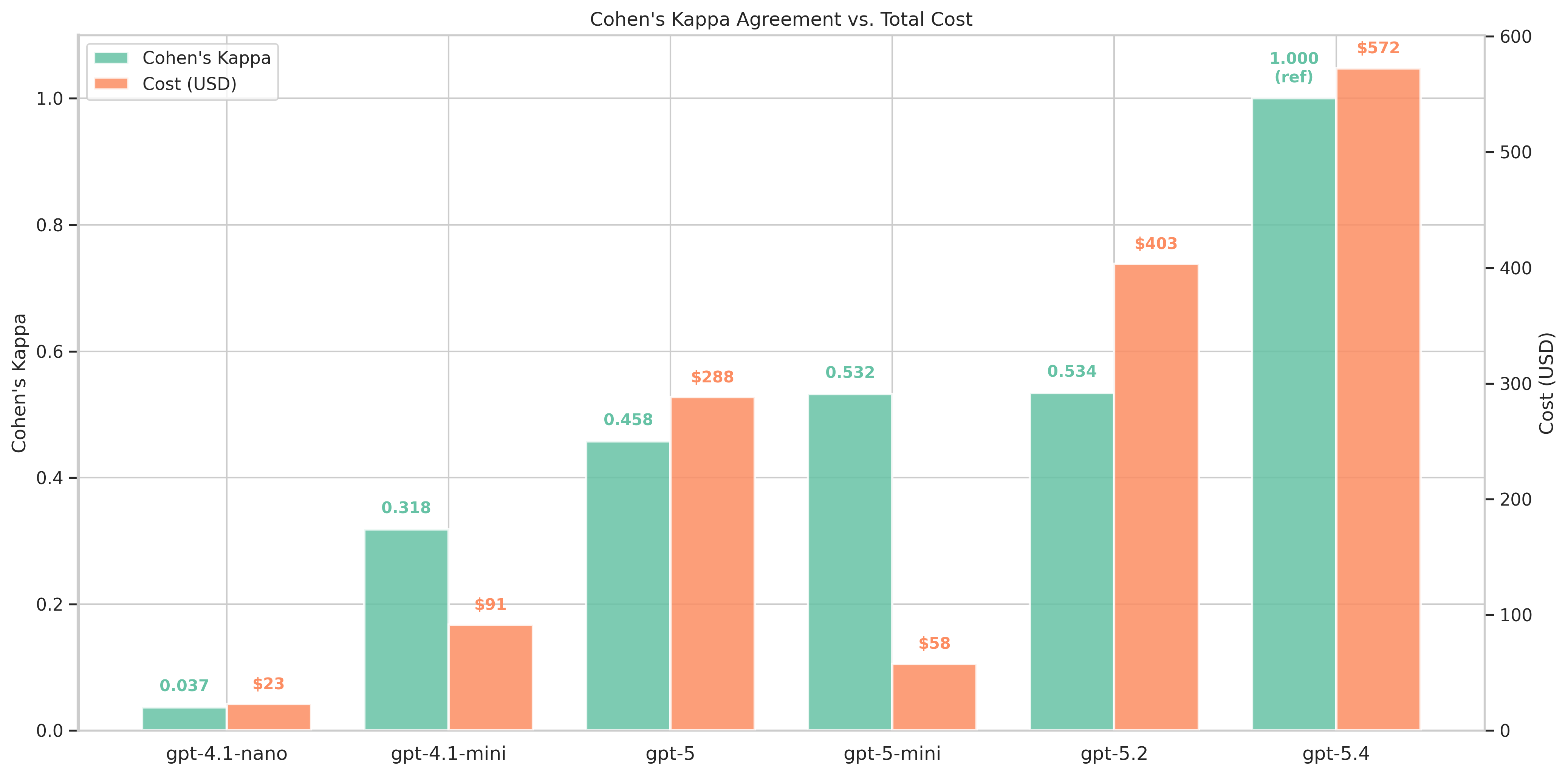}
    \caption{Cohen's $\kappa$ agreement with the reference judge (\textit{GPT-5.4}) and extrapolated total evaluation cost (USD) for six candidate judge models. Agreement is computed on a random subset of $100$ response pairs, and costs are extrapolated to the full evaluation pipeline. \textit{GPT-5-mini} achieves agreement comparable to substantially more expensive alternatives at a fraction of the cost.}
    \label{fig:kappa-cost}
\end{figure}

Figure~\ref{fig:kappa-cost} summarizes the results. Agreement with the reference model scales broadly with model capability, with the lower-capacity models exhibiting poor agreement and the higher-capacity candidates converging to comparable $\kappa$ values. The costs shown in Figure~\ref{fig:kappa-cost} are extrapolated from the $100$-pair subset to the full evaluation pipeline, reflecting the total expense each candidate would incur if deployed as the judge across all checkpoints and model scales. \textit{GPT-5-mini} achieves agreement on par with substantially more expensive alternatives while incurring the lowest extrapolated evaluation cost by a wide margin. We therefore selected \textit{GPT-5-mini} as the judge model for all preference ranking evaluations reported in this study, as it offers the most favorable trade-off between annotator agreement and computational cost.

\section{Adaptation Trajectories over Training Compute}
\label{sec:compute-trajectories}

\begin{figure*}[p]
    \centering
    \begin{subfigure}{0.49\linewidth}
        \centering
        \includegraphics[width=\linewidth]{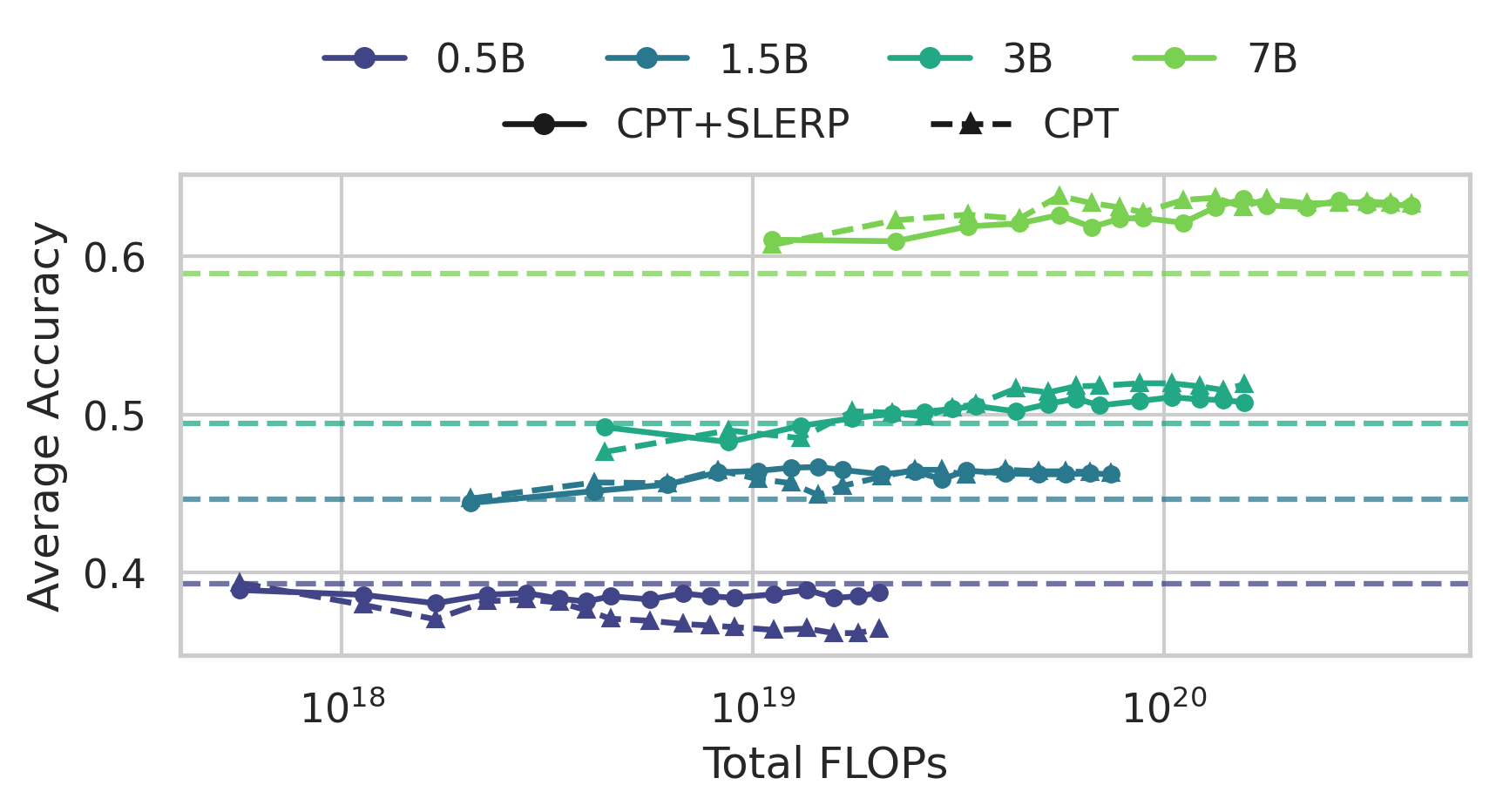}
        \caption{Qwen2.5 Model Family}
        \label{fig:qwen2.5-mcqa}
    \end{subfigure}
    \hfill
    \begin{subfigure}{0.49\linewidth}
        \centering
        \includegraphics[width=\linewidth]{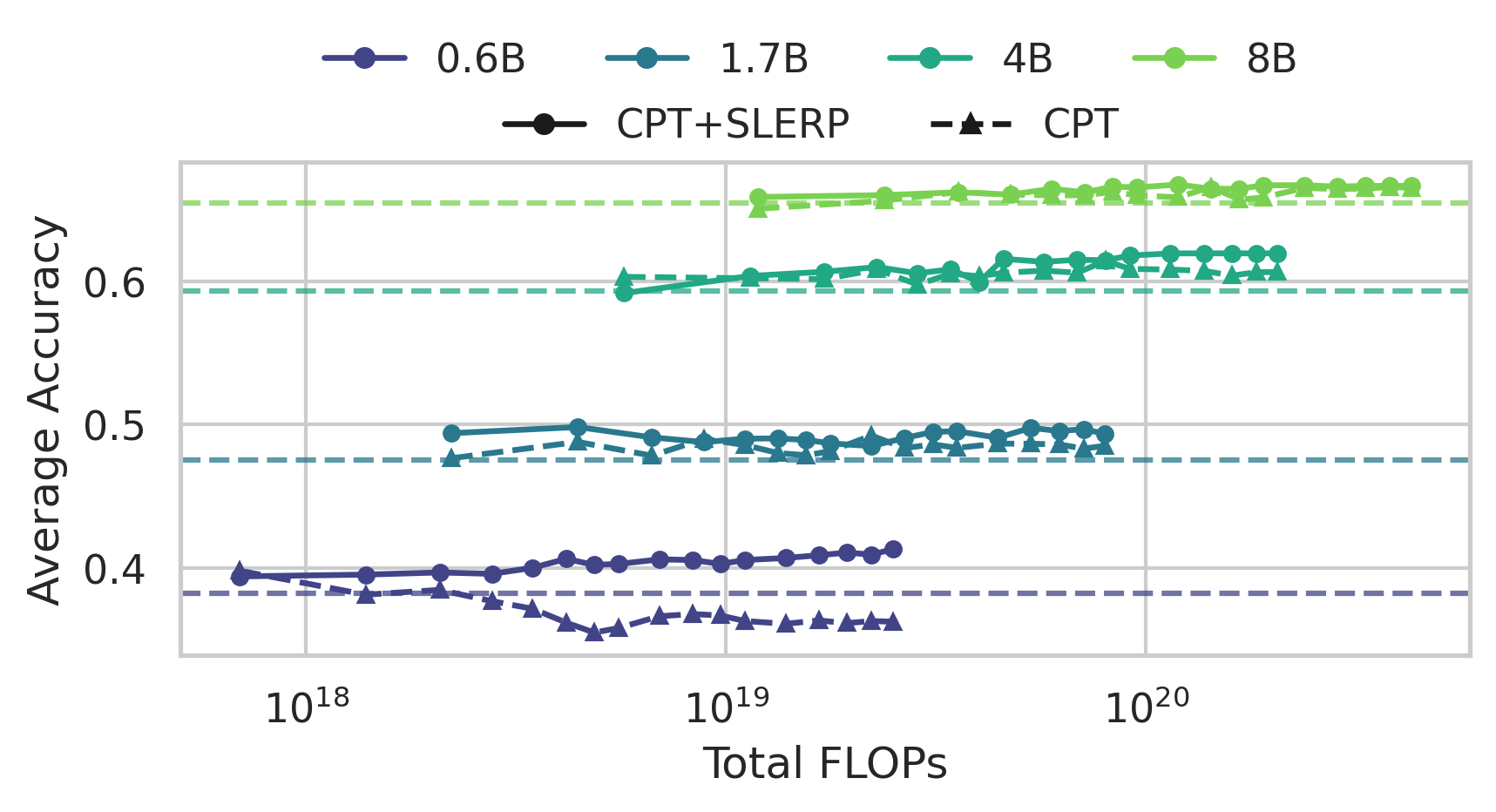}
        \caption{Qwen3 Model Family}
        \label{fig:qwen3-mcqa}
    \end{subfigure}
    \caption{Absolute performance trajectories on medical \textit{MCQA} benchmarks. Each plot illustrates the accuracy progression for the raw continued pre-trained weights ($CPT$) and the resulting model after weight-merging with the source model ($CPT+SLERP$).}
    \label{fig:combined-mcqa-trajectories}
\end{figure*}

Section~\ref{sec:trajectories} reports the adaptation trajectories as a function of training tokens. Figures~\ref{fig:qwen2.5-results-combined-flops} and~\ref{fig:qwen3-results-combined-flops} show the same trajectories as a function of cumulative training compute, which places models of different sizes on a common axis. Figure~\ref{fig:combined-mcqa-trajectories} reports the absolute \textit{MCQA} accuracy of each checkpoint before (\textit{CPT}) and after merging with the source model (\textit{CPT+SLERP}).

\section{HealthBench Results}
\label{sec:healthbench-results}

This section reports the full \textit{HealthBench} \cite{arora2025healthbench} trajectories used as a third evaluation axis (Section~\ref{sec:method:eval}). We first describe the data construction, then present the per-family trajectories.

\subsection{Data Construction}

\textit{HealthBench} is a multilingual benchmark of approximately $5{,}000$ multi-turn clinical conversations, each scored by physician-authored rubrics. We restrict the evaluation to German-language prompts to match the target domain of the study. The native German split contains $29$ samples, which is too small to support stable per-checkpoint estimates. We therefore extend it with $151$ machine-translated samples drawn through stratified subsampling over the original task categories, yielding $180$ prompts that approximate the per-category distribution of the full benchmark. The sample size is bounded by the cost of multi-turn rubric-based scoring across the $144$ checkpoint-level evaluations our trajectory analysis requires ($8$ models $\times$ $18$ checkpoints). Reported scores use \textit{HealthBench}'s default rubric-based metric. Two limitations follow from this construction. Most prompts are machine-translated rather than human-authored in German, and the $180$-prompt size is small enough that per-checkpoint scores might be noisy.

\subsection{Trajectories}

Absolute \textit{HealthBench} scores are low across all models, with the $8B$ source baselines reaching $\sim 0.16$ (\textit{Qwen2.5}) and $\sim 0.30$ (\textit{Qwen3}). Two factors contribute. First, the parameter range considered in this study ($0.5B$–$8B$) is below the regime for which \textit{HealthBench} was primarily designed. Second, our pipeline applies only continual pre-training and \textit{SLERP} merging without further instruction tuning or reinforcement-learning-based post-training, both of which are typically used to reinforce the safety, refusal, and careful-reasoning behaviours that \textit{HealthBench} rewards. The \textit{HealthBench} results should therefore be read as a directional signal alongside the \textit{MCQA} and preference-ranking evaluations rather than as a stand-alone clinical-capability measure.

\begin{figure*}[p]
    \centering
    \begin{subfigure}{0.49\linewidth}
        \centering
        \includegraphics[width=\linewidth]{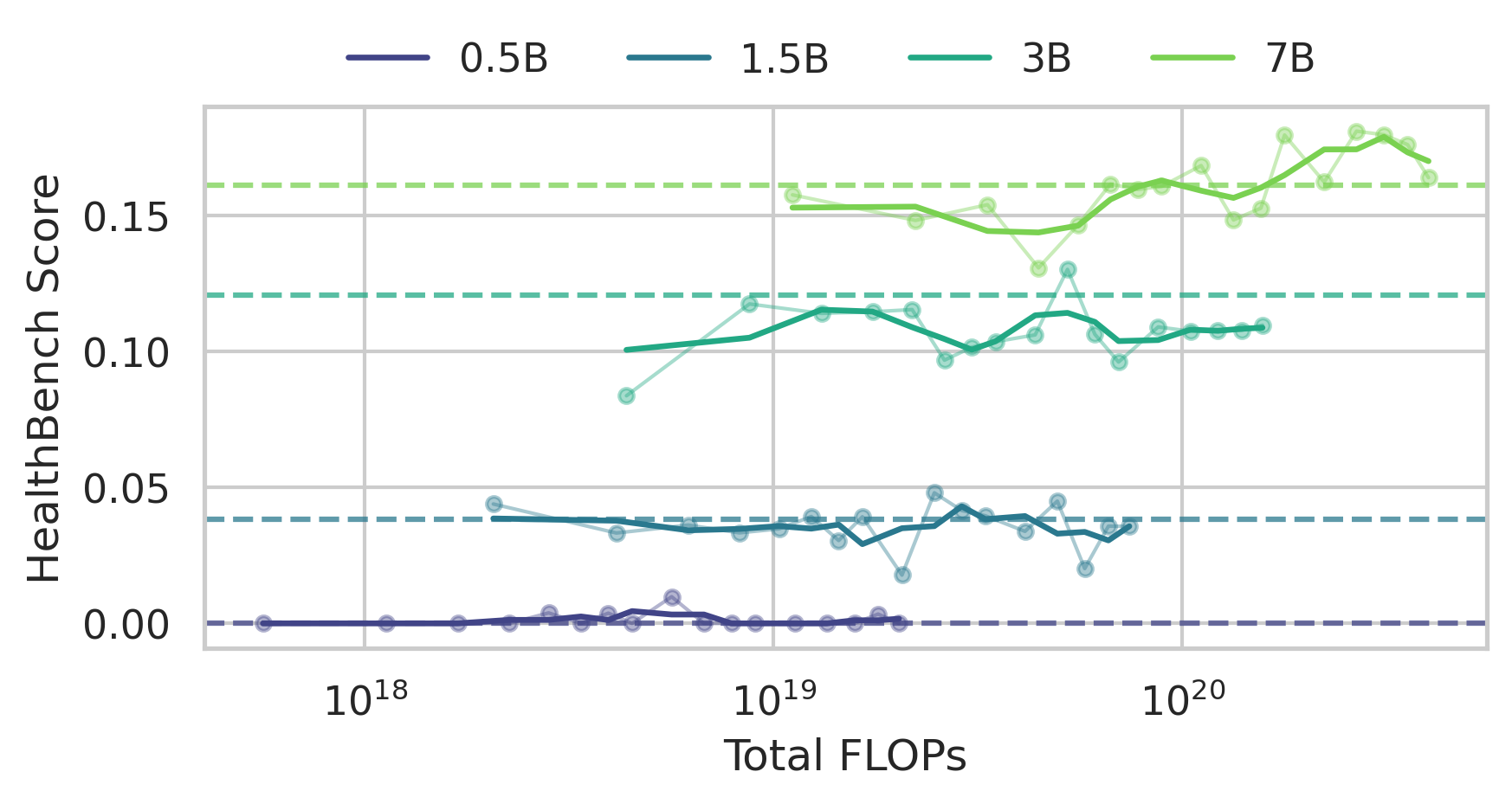}
        \caption{\textit{Qwen2.5} family}
        \label{fig:qwen2.5-healthbench-flops-smoothed}
    \end{subfigure}
    \hfill
    \begin{subfigure}{0.49\linewidth}
        \centering
        \includegraphics[width=\linewidth]{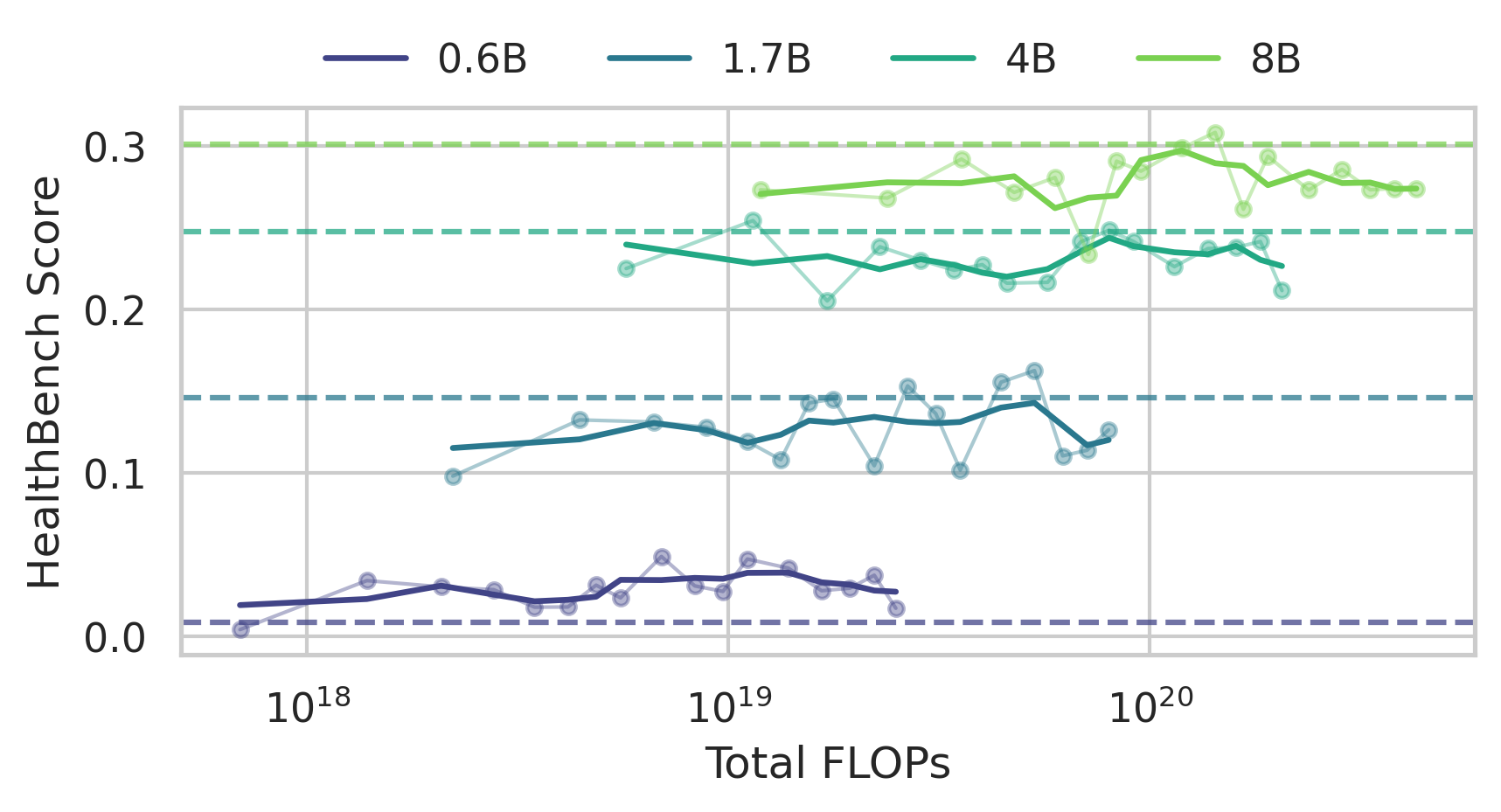}
        \caption{\textit{Qwen3} family}
        \label{fig:qwen3-healthbench-flops-smoothed}
    \end{subfigure}
    \caption{\textit{HealthBench} trajectories as a function of cumulative training compute (FLOPs) for the \textit{Qwen2.5} (left) and \textit{Qwen3} (right) families. Light markers show the raw per-checkpoint scores, and solid lines show a smoothed trend. Dashed lines indicate the source-model baseline for each parameter size.}
    \label{fig:healthbench-combined-flops}
\end{figure*}

Figure~\ref{fig:healthbench-combined-flops} shows the \textit{HealthBench} trajectories for both families as a function of cumulative training compute (FLOPs). Dashed lines denote the unadapted source-model baseline for each size.

The overall picture is that \textit{CPT} produces only modest movements in \textit{HealthBench} score for most models, and the direction of this movement differs systematically across model sizes and families. It is nonetheless broadly consistent with \textit{MCQA}, as the two models with the clearest \textit{MCQA} gains (\textit{Qwen2.5} $7B$, \textit{Qwen3} $0.6B$) are the only two that end above their \textit{HealthBench} source baselines.

\paragraph{Qwen2.5 family.} The $0.5B$ model is pinned near zero throughout training, matching its source baseline, because this size lacks the capacity to produce \textit{HealthBench}-scoring responses regardless of adaptation. The $1.5B$ and $3B$ models hover close to their source baselines, with the $3B$ ending slightly below source after early-checkpoint fluctuations. The $7B$ model is the one clear net positive in this family. It starts near its source baseline ($\sim 0.155$), exhibits a mid-training dip, and shows a late-stage rise that ends above source at $\sim 0.18$.

\paragraph{Qwen3 family.} The $0.6B$ model is the only \textit{Qwen3} size that consistently sits above its (very low) source baseline of $\sim 0.01$, with adapted scores between $0.02$ and $0.05$ across the training run. The $1.7B$, $4B$, and $8B$ models all end below their respective source baselines after a full training run, despite individual checkpoints occasionally touching or briefly surpassing the source line.

\section{Per-Aspect Response Quality}
\label{sec:per-aspect}

This appendix details the per-aspect quality scoring used as a diagnostic for the pairwise ranking trajectories (Section~\ref{sec:trajectories}). The goal of this analysis is not to introduce a new evaluation axis but to inspect which dimensions of the response account for the ranking gains observed during \textit{CPT}.

\subsection{Methodology}

The per-aspect scoring re-uses the responses already generated for the pairwise preference evaluation, using the same \textit{MedInstruct} prompts, the same model responses, and the same \textit{GPT-5-mini} judge. The judge is asked to rate each response on a $1$--$5$ scale along five aspects taken from the rubric prompt of \citet{lin2024the}, namely helpfulness, clarity, factuality, depth, and engagement. We omit the sixth aspect of that rubric, safety, because it is not the focus of this study. Because the five aspect scores are produced by the same judge from the same response, they are not statistically independent measurements. We therefore read the resulting trajectories as a decomposition of the aggregate response-quality signal rather than as five independent evaluations. A further limitation concerns the factuality aspect specifically. The judge's factuality scores rely on the judge model's own medical knowledge, so the factuality trajectory is at best as accurate as that knowledge. We partially mitigate this by triangulating the factuality aspect against the independent \textit{MCQA} accuracy curves (Section~\ref{sec:per-aspect:mcqa-correlation}), which is the same triangulation that motivates the dual-metric framing of the study.

\subsection{Per-Aspect Trajectories}

\begin{figure*}[p]
    \centering
    \includegraphics[width=\linewidth,height=0.88\textheight,keepaspectratio]{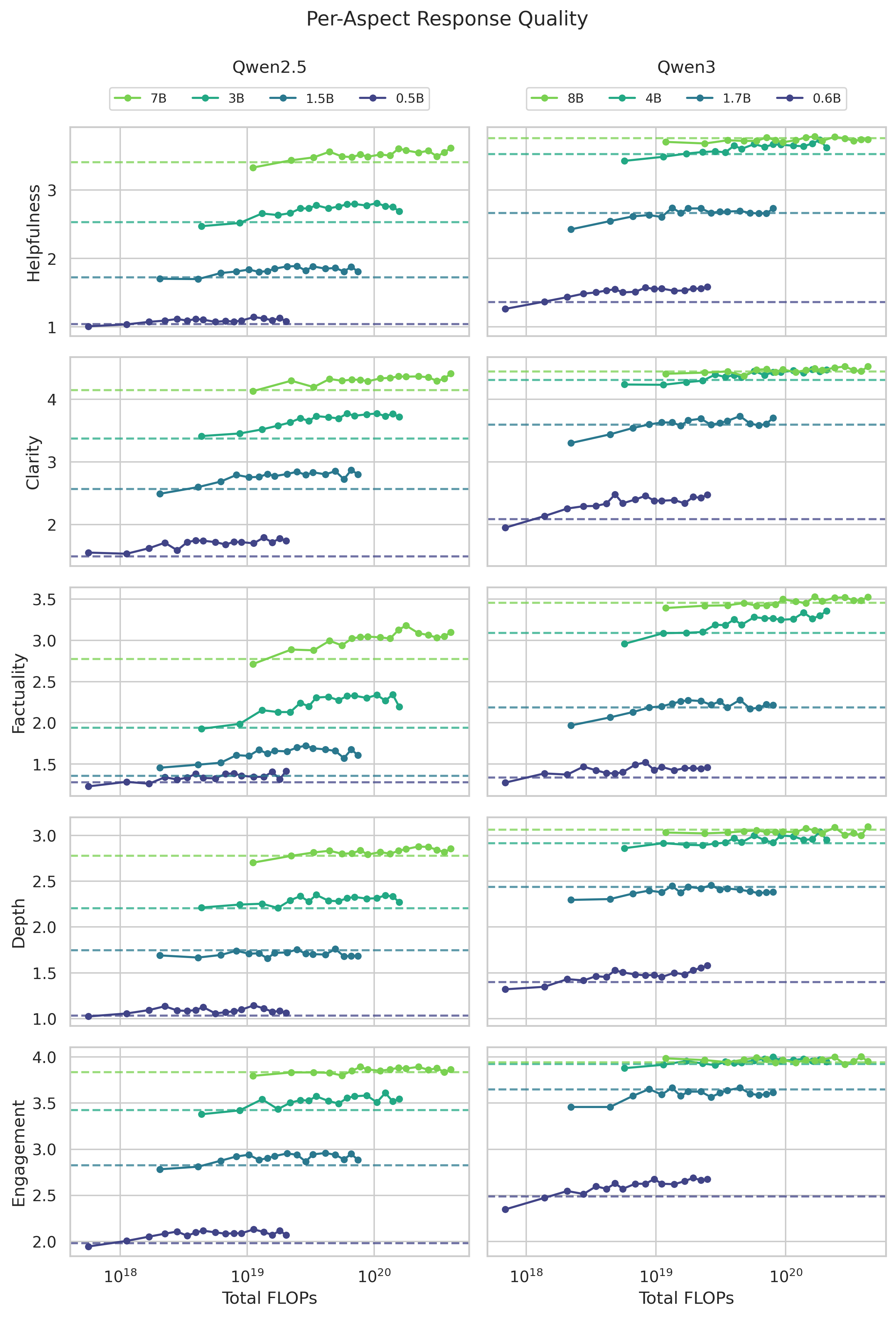}
    \caption{Per-aspect response quality during \textit{CPT}, scored on a $1$--$5$ scale by \textit{GPT-5-mini} following five aspects of the rubric of \citet{lin2024the}. Rows correspond to the five aspects (helpfulness, clarity, factuality, depth, engagement). The left column shows \textit{Qwen2.5} and the right column \textit{Qwen3}. Dashed lines indicate source-model baselines.}
    \label{fig:per-aspect-matrix}
\end{figure*}

Figure~\ref{fig:per-aspect-matrix} shows the per-aspect mean scores as a function of cumulative training compute for all eight adapted models. Dashed lines denote the corresponding source-model scores.

Two patterns are visible. First, in the \textit{Qwen2.5} family, the stylistic aspects (helpfulness, clarity, engagement, and to a lesser extent depth) rise quickly in the early portion of training and then plateau, while factuality continues to climb across the full training run. This is most pronounced in the $3B$ and $7B$ variants, where factuality is the aspect that gains the most absolute ground over the course of training. The $0.5B$ variant remains essentially flat across all aspects, consistent with the limited capacity discussed in Section~\ref{sec:trajectories}.

Second, in the \textit{Qwen3} family, all aspects move much less than in \textit{Qwen2.5}, reflecting the smaller adaptation headroom in this family. The $0.6B$ variant exhibits the largest relative aspect-level gains within \textit{Qwen3}.

\subsection{Correlation with MCQA}
\label{sec:per-aspect:mcqa-correlation}

\begin{figure*}[p]
    \centering

    \begin{subfigure}{\linewidth}
        \centering
        \begin{subfigure}{0.49\linewidth}
            \centering
            \includegraphics[width=\linewidth]{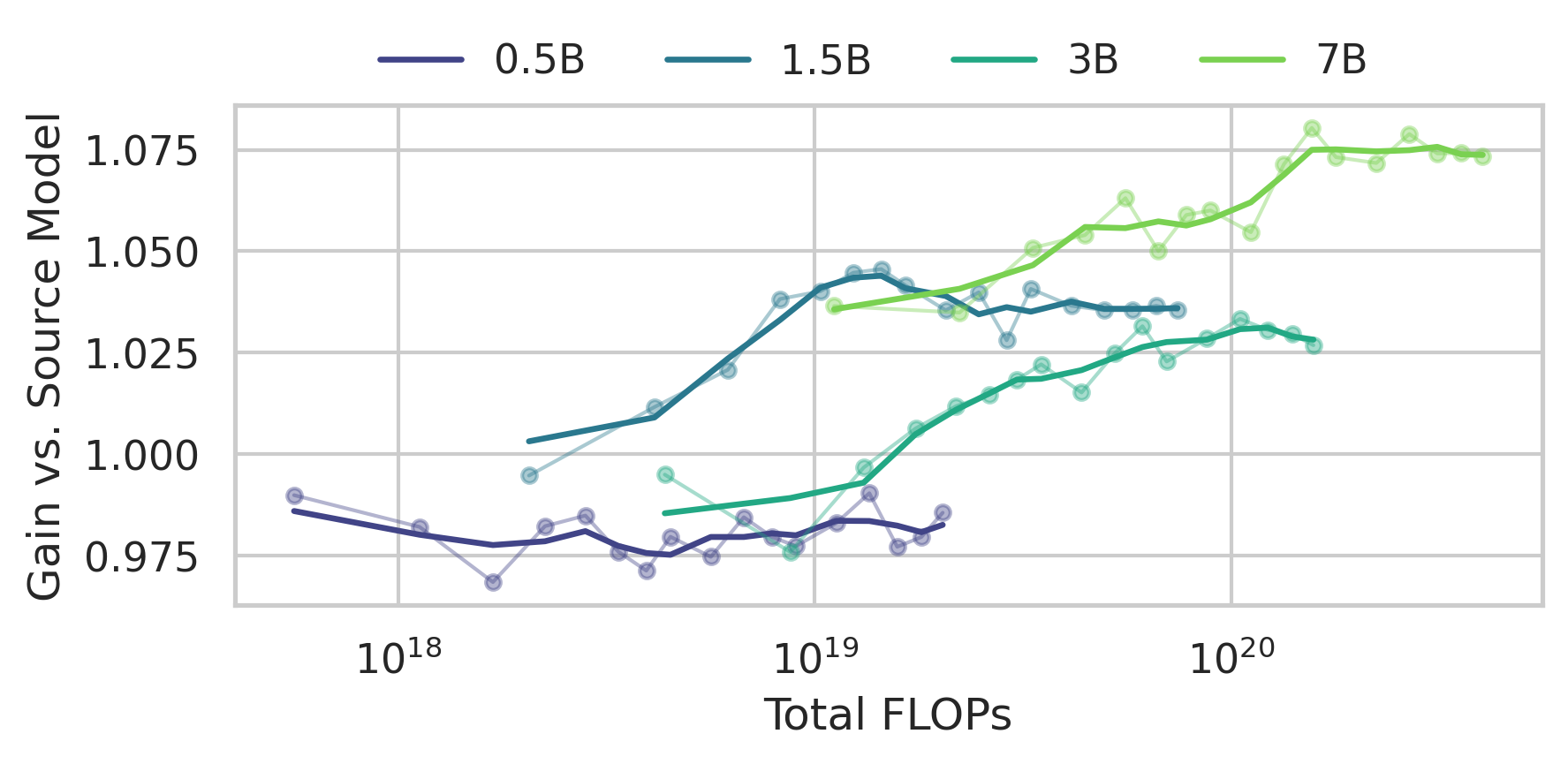}
            \caption{\textit{MCQA}: against source model}
            \label{fig:qwen2.5-mcqa-vs-source-flops}
        \end{subfigure}
        \hfill
        \begin{subfigure}{0.49\linewidth}
            \centering
            \includegraphics[width=\linewidth]{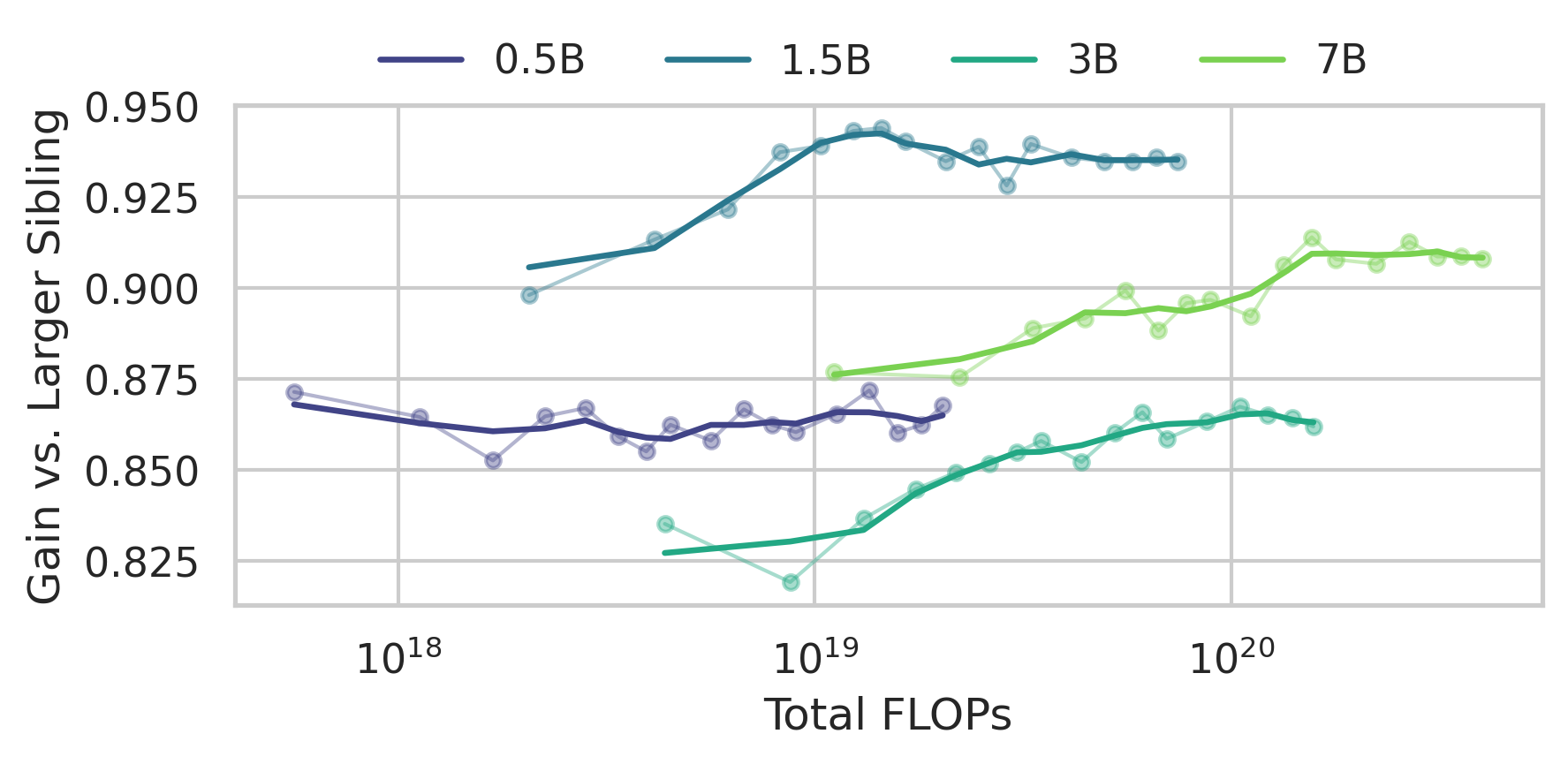}
            \caption{\textit{MCQA}: against next larger model}
            \label{fig:qwen2.5-mcqa-vs-larger-flops}
        \end{subfigure}
    \end{subfigure}

    \vspace{0.5em}

    \begin{subfigure}{\linewidth}
        \centering
        \begin{subfigure}{0.49\linewidth}
            \centering
            \includegraphics[width=\linewidth]{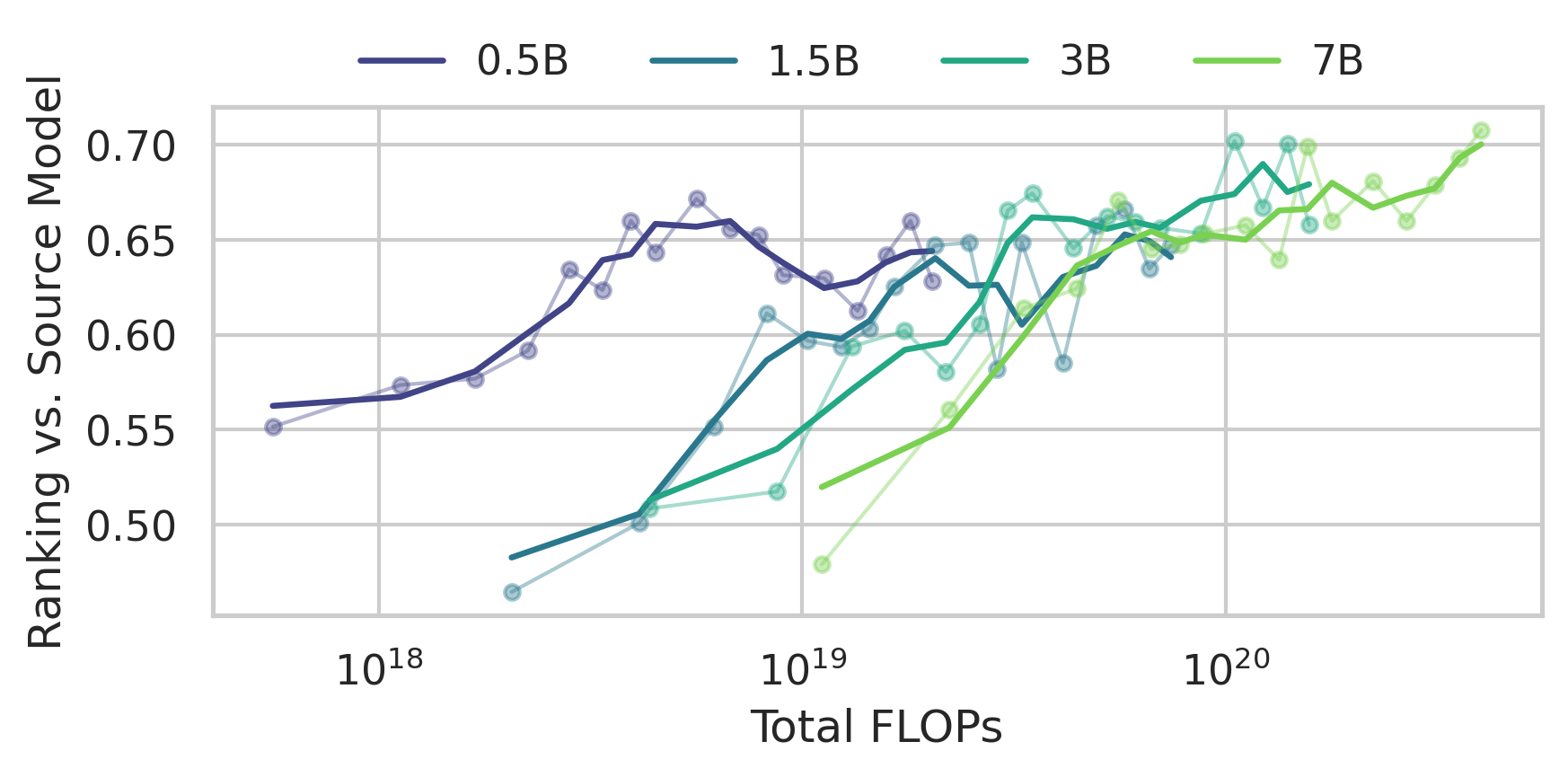}
            \caption{Generative rankings: against source model}
            \label{fig:qwen2.5-ranking-vs-source-flops}
        \end{subfigure}
        \hfill
        \begin{subfigure}{0.49\linewidth}
            \centering
            \includegraphics[width=\linewidth]{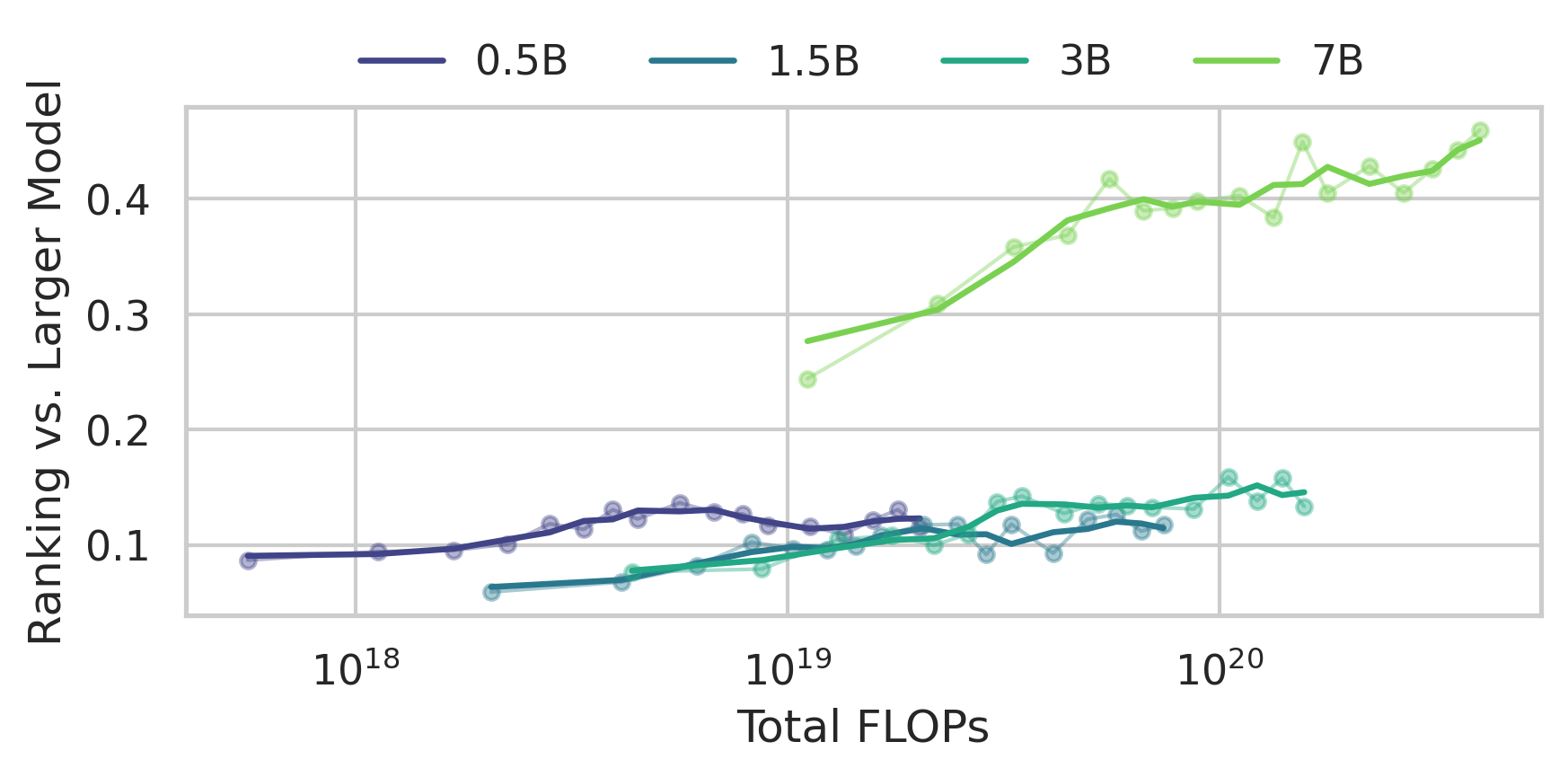}
            \caption{Generative rankings: against next larger model}
            \label{fig:qwen2.5-ranking-vs-larger-flops}
        \end{subfigure}
    \end{subfigure}

    \caption{
        Scaling trajectories for the \textit{Qwen2.5} family as a function of total compute. The top row tracks factual knowledge (\textit{MCQA}) and the bottom row tracks open-ended response quality via Davidson preference rankings. These plots illustrate adaptation dynamics relative to the original source models and relative parity achieved against the next-tier general-purpose variants (e.g., $7B$ vs. $14B$).
    }
    \label{fig:qwen2.5-results-combined-flops}
\end{figure*}

\begin{figure*}[p]
    \centering

    \begin{subfigure}{\linewidth}
        \centering
        \begin{subfigure}{0.49\linewidth}
            \centering
            \includegraphics[width=\linewidth]{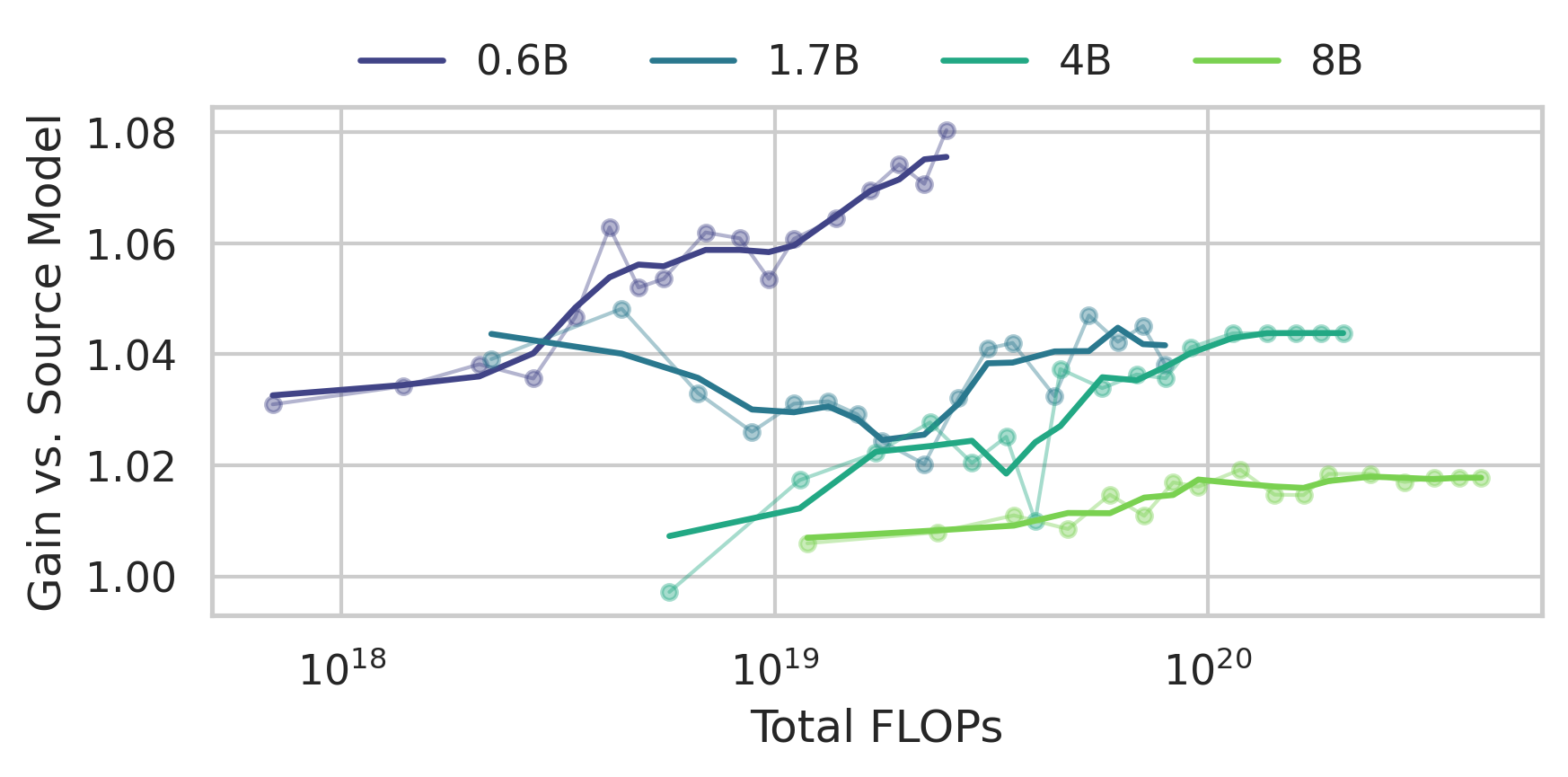}
            \caption{MCQA: against source model}
            \label{fig:qwen3-mcqa-vs-source-flops}
        \end{subfigure}
        \hfill
        \begin{subfigure}{0.49\linewidth}
            \centering
            \includegraphics[width=\linewidth]{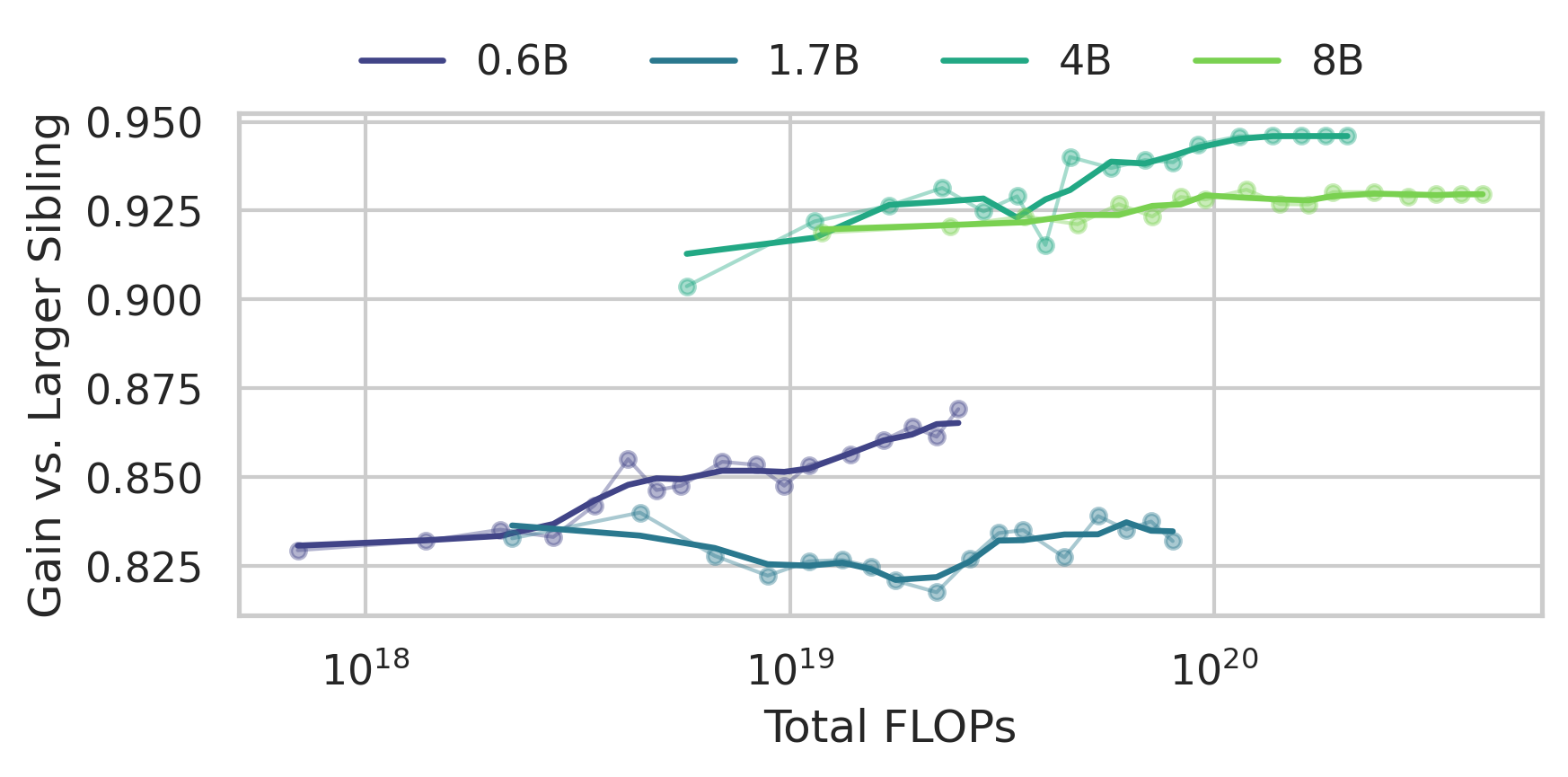}
            \caption{MCQA: against next larger model}
            \label{fig:qwen3-mcqa-vs-larger-flops}
        \end{subfigure}
    \end{subfigure}

    \begin{subfigure}{\linewidth}
        \centering
        \begin{subfigure}{0.49\linewidth}
            \centering
            \includegraphics[width=\linewidth]{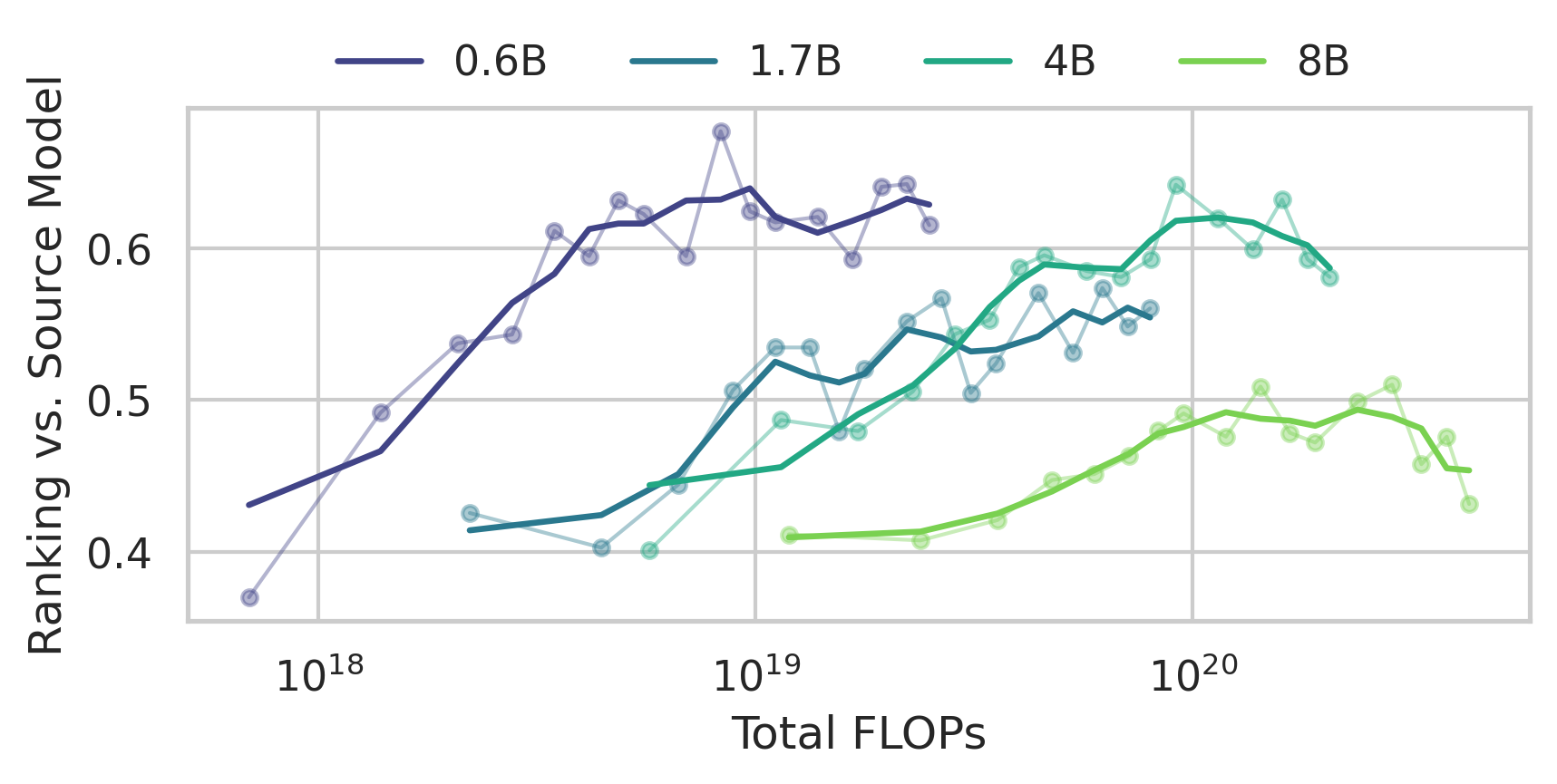}
            \caption{Generative rankings: against source model}
            \label{fig:qwen3-ranking-vs-source-flops}
        \end{subfigure}
        \hfill
        \begin{subfigure}{0.49\linewidth}
            \centering
            \includegraphics[width=\linewidth]{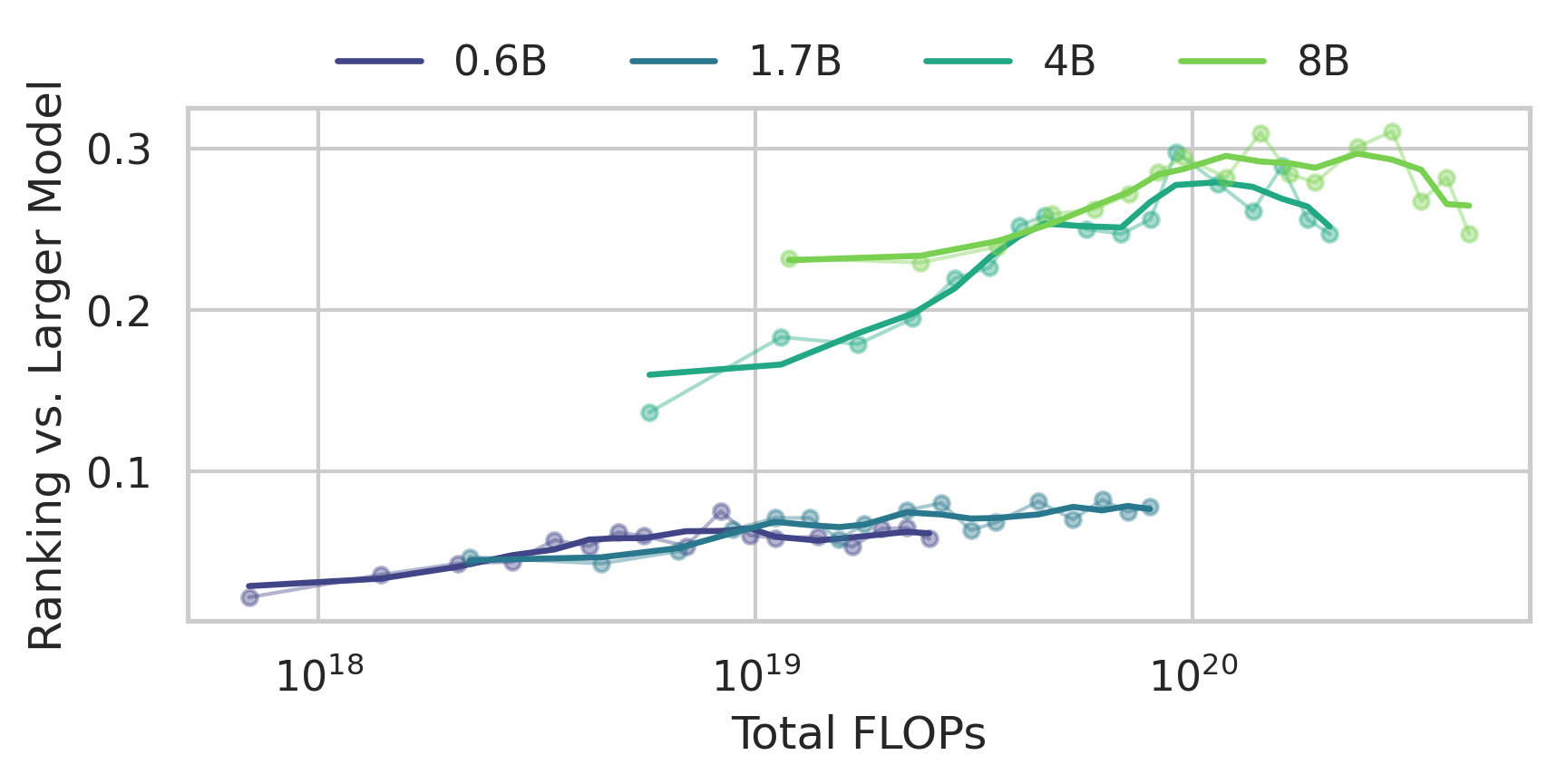}
            \caption{Generative rankings: against next larger model}
            \label{fig:qwen3-ranking-vs-larger-flops}
        \end{subfigure}
    \end{subfigure}

    \caption{
    Scaling trajectories for the \textit{Qwen3} family as a function of total compute, analogous to Figure~\ref{fig:qwen2.5-results-combined-flops}. The top row tracks factual knowledge while the bottom row tracks preference rankings, comparing adapted variants against their original source models and the next-tier general-purpose baselines (e.g., $8B$ vs. $14B$).
    }
    \label{fig:qwen3-results-combined-flops}
\end{figure*}

The factuality aspect provides an internal consistency check on the broader claim that knowledge and stylistic capabilities scale at different rates. Qualitatively, the factuality trajectories closely follow the \textit{MCQA} accuracy curves (Figures~\ref{fig:qwen2.5-mcqa-vs-source-flops}, \ref{fig:qwen3-mcqa-vs-source-flops}) for each size. The \textit{Qwen2.5} $7B$ variant, which shows the strongest sustained \textit{MCQA} gain in its family, also shows the largest sustained factuality gain across all aspects and sizes. The \textit{Qwen3} $8B$ variant, which is near-flat on \textit{MCQA}, is also near-flat on factuality. The \textit{Qwen3} $0.6B$ variant, which is the family's strongest \textit{MCQA} gainer, is also the family's strongest factuality gainer. Since \textit{MCQA} and the factuality aspect are produced by independent procedures, this alignment supports reading the factuality trajectories as a meaningful proxy for knowledge accumulation within the ranking signal.

\section{Output Token-Distribution Shift}
\label{sec:token-shift}

This appendix describes the judge-free token-distribution shift used in Section~\ref{sec:trajectories}. It measures how far the surface form of a checkpoint's outputs has moved away from its source model, without involving an LLM judge.

\paragraph{Data.} The measurement re-uses the free-text generations already produced for the preference-ranking evaluation (Section~\ref{sec:method:eval:ranking}), namely the same $216$ machine-translated \textit{MedInstruct} prompts and the same responses generated with \textit{vLLM} \cite{vllm}. For each model size we use the generations of every merged (\textit{CPT + SLERP}) checkpoint and of the unadapted instruction-tuned source model, which serves as the reference. Only the assistant-turn content is used, while prompt and system tokens are excluded.

\paragraph{Empirical token distribution.} For a model $m$ with responses $\mathcal{R}_m$, each response is tokenized with the tokenizer of the corresponding model family \cite{wolf-etal-2020-transformers}, without special tokens. Pooling token counts over all responses gives a unigram count vector over the vocabulary $V$, which is normalized to an empirical distribution
\begin{equation*}
P_m(t) = \frac{\sum_{r \in \mathcal{R}_m} \operatorname{count}(t, r)}{\sum_{t' \in V} \sum_{r \in \mathcal{R}_m} \operatorname{count}(t', r)}, \quad t \in V.
\end{equation*}
Each family is compared only against its own source model, so all distributions in a comparison share the same tokenizer and vocabulary.

\paragraph{Shift metric.} The shift of a checkpoint $c$ from its source model $b$ is the Jensen--Shannon divergence \cite{lin1991divergence}
\begin{equation*}
\mathrm{JSD}(P_c \,\|\, P_b) = \tfrac{1}{2} D_{\mathrm{KL}}(P_c \,\|\, M) + \tfrac{1}{2} D_{\mathrm{KL}}(P_b \,\|\, M),
\end{equation*}
with $M = \tfrac{1}{2}(P_c + P_b)$ and $D_{\mathrm{KL}}$ the Kullback--Leibler divergence \cite{kullback1951information}. Logarithms are taken in base $2$, so the divergence is reported in bits and bounded to $[0, 1]$. The JSD is symmetric and finite for any pair of distributions, so no smoothing is applied.

\begin{figure*}[p]
    \centering
    \begin{subfigure}{0.49\linewidth}
        \centering
        \includegraphics[width=\linewidth]{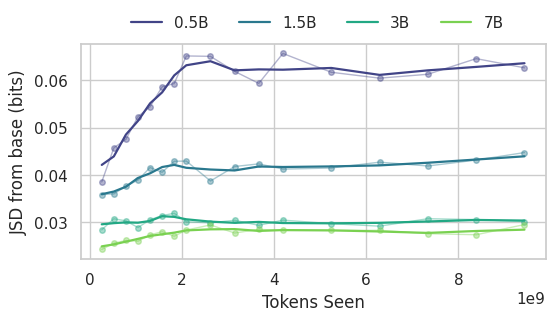}
        \caption{\textit{Qwen2.5} family}
        \label{fig:qwen2.5-token-shift}
    \end{subfigure}
    \hfill
    \begin{subfigure}{0.49\linewidth}
        \centering
        \includegraphics[width=\linewidth]{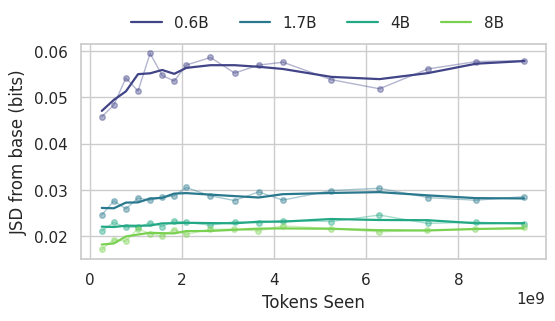}
        \caption{\textit{Qwen3} family}
        \label{fig:qwen3-token-shift}
    \end{subfigure}
    \caption{Jensen--Shannon divergence (bits) between the output token distribution of each merged checkpoint and that of its source model, as a function of training tokens seen. Light markers show per-checkpoint values, and solid lines show a centered running average over three checkpoints.}
    \label{fig:token-shift}
\end{figure*}

\paragraph{Results.} Figure~\ref{fig:token-shift} shows the trajectories for both families. At every model size, most of the shift occurs within the first $\sim 2B$ tokens, after which the divergence plateaus. The magnitude of the shift decreases with model size, with the smallest models moving furthest from their source distribution, but the front-loaded shape is shared across all sizes. Since a unigram distribution captures a lexical surface shift rather than style as such, we use this measure only as a judge-free indicator of how quickly the surface form of the outputs settles, not as a measure of response quality.

\section{Convergence Analysis}
\label{sec:convergence-analysis}

This appendix details the convergence fits summarized in Table~\ref{tab:c90-summary}.

\paragraph{Curve fit.} For every model and metric (\textit{MCQA} accuracy, preference ranking against the source model, and token-distribution shift), we fit the trajectory as a function of training tokens seen $x$ (in billions) with a saturating exponential
\begin{equation*}
y(x) = y_0 + (A - y_0)\left(1 - e^{-x/\tau}\right),
\end{equation*}
where $y_0$ is the starting level, $A$ the asymptote, and $\tau$ a characteristic token scale. All three parameters are fitted by nonlinear least squares, with $A$ bounded to $[\min y, \max y + 0.5 \cdot \mathrm{range}(y)]$ and $\tau$ to $[0.01, 100] \cdot \mathrm{range}(x)$. Among the candidate curve families we compared, the saturating exponential matched the observed shape substantially better than a logarithmic fit while retaining a single interpretable rate parameter.

\paragraph{Convergence score.} From each fit we derive
\begin{equation*}
c_{90} = \tau \ln 10,
\end{equation*}
the number of tokens needed to reach $90\%$ of the fitted total gain $A - y_0$. Lower values indicate faster convergence.

\paragraph{Fit-quality gate.} A low coefficient of determination indicates that a curve barely moves relative to checkpoint-level noise, in which case $c_{90}$ is not meaningful. We therefore retain only fits with $R^2 \geq 0.6$. The conclusions are unchanged when this threshold is moved between $0.6$ and $0.7$. Two \textit{MCQA} fits (\textit{Qwen3} $1.7B$ and \textit{Qwen2.5} $0.5B$) and three token-shift fits are excluded on this basis. All ranking fits are retained.

\paragraph{Uncertainty.} We attach $95\%$ confidence intervals to each $c_{90}$ using a residual bootstrap with $4{,}000$ resamples. Residuals of the fitted curve are resampled with replacement onto the fitted values, the curve is refitted, and the $2.5$th and $97.5$th percentiles of the resulting $c_{90}$ values form the interval. Resampling residuals rather than checkpoints keeps the token grid fixed, so every resample spans the full training run.

\paragraph{Results.} Table~\ref{tab:c90-per-model} reports the per-model estimates. Among the six models for which both the \textit{MCQA} and the ranking fit are retained, rankings converge earlier in five. The exception is \textit{Qwen2.5} $1.5B$, whose \textit{MCQA} curve saturates unusually early. The direction of the effect is consistent, but its magnitude is not precisely determined, as the \textit{MCQA} confidence intervals are typically as wide as the estimates themselves, and the paired difference between \textit{MCQA} and ranking $c_{90}$ excludes zero individually only for \textit{Qwen3} $0.6B$ and \textit{Qwen2.5} $7B$. Across the six models, the median paired difference is $1.5B$ tokens (hierarchical bootstrap $95\%$ CI $[-0.3, 3.9]$, Wilcoxon signed-rank $p = 0.06$, sign test $p = 0.11$). We therefore report a consistent tendency for preference rankings and the token-distribution shift to saturate earlier than \textit{MCQA} accuracy, rather than a precise speed-up factor.

\begin{table*}[t]
    \centering
    \small
    \begin{tabular}{llcccccc}
        \toprule
        & & \multicolumn{2}{c}{\textit{MCQA}} & \multicolumn{2}{c}{Preference ranking} & \multicolumn{2}{c}{Token shift} \\
        \cmidrule(lr){3-4} \cmidrule(lr){5-6} \cmidrule(lr){7-8}
        Family & Size & $c_{90}$ [95\% CI] & $R^2$ & $c_{90}$ [95\% CI] & $R^2$ & $c_{90}$ & $R^2$ \\
        \midrule
        \textit{Qwen2.5} & $0.5B$ & -- & $0.04$ & $1.60$ $[0.84, 3.04]$ & $0.74$ & $1.85$ & $0.92$ \\
                         & $1.5B$ & $0.86$ $[0.71, 1.32]$ & $0.85$ & $1.56$ $[0.94, 2.60]$ & $0.81$ & $2.08$ & $0.73$ \\
                         & $3B$   & $3.12$ $[2.01, 5.00]$ & $0.86$ & $2.37$ $[1.58, 3.68]$ & $0.87$ & -- & $0.21$ \\
                         & $7B$   & $3.88$ $[2.42, 6.92]$ & $0.85$ & $1.18$ $[0.81, 1.68]$ & $0.89$ & $1.85$ & $0.78$ \\
        \midrule
        \textit{Qwen3}   & $0.6B$ & $6.30$ $[3.79, 13.71]$ & $0.87$ & $1.18$ $[0.84, 1.60]$ & $0.91$ & $1.09$ & $0.64$ \\
                         & $1.7B$ & -- & $0.17$ & $2.25$ $[1.20, 4.38]$ & $0.75$ & -- & $0.57$ \\
                         & $4B$   & $4.38$ $[2.39, 9.09]$ & $0.79$ & $2.16$ $[1.55, 3.00]$ & $0.91$ & -- & $0.42$ \\
                         & $8B$   & $2.63$ $[1.58, 4.46]$ & $0.82$ & $1.89$ $[0.76, 5.31]$ & $0.61$ & $1.44$ & $0.79$ \\
        \midrule
        \multicolumn{2}{l}{Median (retained fits)} & $3.50$ & & $1.74$ & & $1.85$ & \\
        \bottomrule
    \end{tabular}
    \caption{Per-model convergence scores $c_{90}$ in billions of training tokens, with $95\%$ residual-bootstrap confidence intervals for \textit{MCQA} and preference rankings. Entries marked -- fail the $R^2 \geq 0.6$ fit-quality gate and are excluded from the medians.}
    \label{tab:c90-per-model}
\end{table*}

\section{Catastrophic Forgetting on General-Domain Benchmarks}
\label{sec:forgetting}

We evaluated the final checkpoints of our training pipeline on a number of german general-domain benchmarks to test whether our adapted models retain the general capabilities of their source models. For this evaluation, we selected several widely used benchmarks outside the medical domain. \textit{MMMLU} \citep{openaiMMMLU, hendryckstest2021} is the professionally human-translated multilingual version of MMLU and measures broad multi-task generalization across knowledge areas. \textit{ARC-Easy} and \textit{ARC-Challenge} \citep{clark2018arc} cover scientific and commonsense question answering, \textit{GSM8k} \citep{cobbe2021gsm8k} assesses mathematical reasoning, and \textit{HellaSwag} \citep{zellers2019hellaswag} evaluates contextual understanding and narrative completion. All benchmarks are evaluated in $1$-shot setting via the \textit{Eval-Harness} \cite{eval-harness} framework. For each model size and family, we compare the source model, the \textit{CPT}-only checkpoint, and the merged \textit{CPT + SLERP} checkpoint. Tables~\ref{tab:qwen25_de_results} and \ref{tab:qwen3_de_results} report the per-benchmark accuracies. Across all eight model size and family combinations, the \textit{CPT + SLERP} variant matches or exceeds the source-model average. We therefore observe no catastrophic forgetting, and in several cases a small net improvement on general-domain tasks.

\section{Win-Tie-Loss Distributions}
\label{sec:win-tie-loss}

For the smallest variants, the ``Ties'' category dominates the judge's pairwise outputs, suggesting that the judge often cannot distinguish between adapted and source responses. As model size increases, the tie share shrinks. This pattern likely reflects two factors that we are not able to disentangle within the scope of this study. First, weight interpolation is less effective in low-parameter regimes \cite{yadav2024, Lu2025}, which may leave the merged checkpoints close to the baseline. Second, smaller models exhibit lower general correctness, so both the adapted and source variants may produce factually incorrect outputs, in which case the judge has no meaningful basis for preference and defaults to a tie.

\clearpage
\begin{figure*}[p]
\centering
\begin{lstlisting}[
  basicstyle=\ttfamily\small,
  frame=single,
  breaklines=true
]
messages:
  - role: system
    content: |
      You are an AI assistant that evaluates pairs of responses
      to a given query. Your goal is to determine which response
      is better based on correctness, clarity, completeness,
      and relevance.

  - role: user
    content: |
      Compare the following two responses. Your answer needs
      to follow this JSON schema:

      ```json
      {
        "type": "object",
        "properties": {
          "winner": {
            "type": "string",
            "enum": ["A", "B", "0"],
            "description": "The winning response, i.e. 'A', 'B' or '0' indicating a tie."
          }
        },
      }
      ```

      {% if context %}
      Context:
      {% for msg in context %}
        {{ msg.role | capitalize }}:
          {{ msg.content | indent(width=8) }}
      {% endfor %}
      {% endif %}

      Query:
        {{ query | indent(width=8) }}

      Response A:
        {{ response_a | indent(width=8) }}

      Response B:
        {{ response_b | indent(width=8) }}

      Based on the query and context above,
      which response is better?
\end{lstlisting}
\caption{Prompt template for the pairwise Judge model. The structure facilitates head-to-head comparisons by providing the evaluator with query context and two candidate responses, requiring a structured JSON output to determine the winner based on clinical utility.}
\label{fig:judge-model-prompt-template}
\end{figure*}

\clearpage
\begin{table*}[p]
    \centering
    \small
    \setlength{\tabcolsep}{4pt}
    \begin{tabular}{llrrrrrr}
        \toprule
        Size & Variant            & GSM8K        & MMMLU        & ARC-C        & ARC-E        & HSwag        & Avg          \\
             &                    & acc          & acc (avg)    & acc\_norm    & acc\_norm    & acc\_norm    &              \\
        \midrule
        0.5B & source       & \underline{4.32 {\scriptsize $\pm$0.56}}  & 35.74 {\scriptsize $\pm$0.49}             & 24.91 {\scriptsize $\pm$1.26}             & 35.06 {\scriptsize $\pm$0.98}             & 31.88 {\scriptsize $\pm$0.47}             & 26.38 {\scriptsize $\pm$0.36}             \\
             & CPT          & 0.08 {\scriptsize $\pm$0.08}              & 34.71 {\scriptsize $\pm$0.48}             & 25.09 {\scriptsize $\pm$1.27}             & \underline{36.66 {\scriptsize $\pm$0.99}} & \underline{33.26 {\scriptsize $\pm$0.47}} & 25.96 {\scriptsize $\pm$0.35}             \\
             & CPT + SLERP  & 3.94 {\scriptsize $\pm$0.54}              & \underline{36.95 {\scriptsize $\pm$0.49}} & \underline{25.77 {\scriptsize $\pm$1.28}} & 36.53 {\scriptsize $\pm$0.99}             & 32.76 {\scriptsize $\pm$0.47}             & \underline{27.19 {\scriptsize $\pm$0.37}} \\
        \midrule
        1.5B & source       & 31.61 {\scriptsize $\pm$1.28}             & 49.08 {\scriptsize $\pm$0.50}             & \underline{33.36 {\scriptsize $\pm$1.38}} & 48.70 {\scriptsize $\pm$1.03}             & 43.21 {\scriptsize $\pm$0.50}             & 41.19 {\scriptsize $\pm$0.45}             \\
             & CPT          & 24.72 {\scriptsize $\pm$1.19}             & 49.18 {\scriptsize $\pm$0.50}             & 33.28 {\scriptsize $\pm$1.38}             & 49.62 {\scriptsize $\pm$1.03}             & 43.56 {\scriptsize $\pm$0.50}             & 40.07 {\scriptsize $\pm$0.44}             \\
             & CPT + SLERP  & \underline{32.68 {\scriptsize $\pm$1.29}} & \underline{49.52 {\scriptsize $\pm$0.50}} & 33.11 {\scriptsize $\pm$1.38}             & \underline{49.83 {\scriptsize $\pm$1.03}} & \underline{43.79 {\scriptsize $\pm$0.50}} & \underline{41.78 {\scriptsize $\pm$0.45}} \\
        \midrule
        3B   & source       & 44.50 {\scriptsize $\pm$0.43}              & 57.07 {\scriptsize $\pm$0.49}             & \underline{40.53 {\scriptsize $\pm$1.43}} & 56.14 {\scriptsize $\pm$1.02}             & \underline{52.54 {\scriptsize $\pm$0.50}} & 41.76 {\scriptsize $\pm$0.39}             \\
             & CPT          & 40.71 {\scriptsize $\pm$1.35}             & 57.54 {\scriptsize $\pm$0.49}             & 39.93 {\scriptsize $\pm$1.43}             & 57.95 {\scriptsize $\pm$1.01}             & 51.82 {\scriptsize $\pm$0.50}             & 49.59 {\scriptsize $\pm$0.46}             \\
             & CPT + SLERP  & \underline{45.49 {\scriptsize $\pm$1.37}} & \underline{58.38 {\scriptsize $\pm$0.49}} & 40.02 {\scriptsize $\pm$1.43}             & \underline{58.04 {\scriptsize $\pm$1.01}} & 52.38 {\scriptsize $\pm$0.50}             & \underline{50.86 {\scriptsize $\pm$0.47}} \\
        \midrule
        7B   & source       & 47.99 {\scriptsize $\pm$1.38}             & 65.53 {\scriptsize $\pm$0.47}             & 51.11 {\scriptsize $\pm$1.46}             & 68.86 {\scriptsize $\pm$0.95}             & 61.01 {\scriptsize $\pm$0.49}             & 58.90 {\scriptsize $\pm$0.46}             \\
             & CPT          & 62.77 {\scriptsize $\pm$1.33}             & 66.66 {\scriptsize $\pm$0.47}             & 50.17 {\scriptsize $\pm$1.46}             & 70.37 {\scriptsize $\pm$0.94}             & \textbf{61.39 {\scriptsize $\pm$0.49}}    & 62.27 {\scriptsize $\pm$0.46}             \\
             & CPT + SLERP  & \textbf{65.66 {\scriptsize $\pm$1.31}}    & \textbf{67.15 {\scriptsize $\pm$0.47}}    & \textbf{52.65 {\scriptsize $\pm$1.46}}    & \textbf{70.96 {\scriptsize $\pm$0.93}}    & 61.29 {\scriptsize $\pm$0.49}             & \textbf{63.54 {\scriptsize $\pm$0.45}}    \\
        \bottomrule
    \end{tabular}
    \caption{General-domain benchmark performance of the \textit{Qwen2.5} family, evaluated $1$-shot on the German-translated benchmarks at the final checkpoint of the full adaptation pipeline (\textit{CPT} and \textit{CPT + SLERP}) and on the unadapted source model. Values are accuracy in \% with $\pm$ standard error. \textbf{Bold} marks the global maximum per column, and \underline{underlined} marks the maximum within a size group.}
    \label{tab:qwen25_de_results}
\end{table*}

\begin{table*}[p]
    \centering
    \small
    \setlength{\tabcolsep}{4pt}
    \begin{tabular}{llrrrrrr}
        \toprule
        Size & Variant            & GSM8K        & MMMLU        & ARC-C        & ARC-E        & HSwag        & Avg          \\
             &                    & acc          & acc (avg)    & acc\_norm    & acc\_norm    & acc\_norm    &              \\
        \midrule
        0.6B & source       & 9.40 {\scriptsize $\pm$0.80}              & 40.05 {\scriptsize $\pm$0.50}             & \underline{28.16 {\scriptsize $\pm$1.31}} & 40.49 {\scriptsize $\pm$1.01}             & 33.39 {\scriptsize $\pm$0.47}             & 30.30 {\scriptsize $\pm$0.39}             \\
             & CPT          & 5.08 {\scriptsize $\pm$0.60}              & 37.18 {\scriptsize $\pm$0.49}             & 27.99 {\scriptsize $\pm$1.31}             & 43.81 {\scriptsize $\pm$1.02}             & \underline{35.07 {\scriptsize $\pm$0.48}} & 29.83 {\scriptsize $\pm$0.38}             \\
             & CPT + SLERP  & \underline{18.20 {\scriptsize $\pm$1.06}} & \underline{40.53 {\scriptsize $\pm$0.50}} & 27.99 {\scriptsize $\pm$1.31}             & \underline{44.87 {\scriptsize $\pm$1.02}} & 34.60 {\scriptsize $\pm$0.48}             & \underline{33.24 {\scriptsize $\pm$0.42}} \\
        \midrule
        1.7B & source       & 34.65 {\scriptsize $\pm$1.31}             & 51.15 {\scriptsize $\pm$0.50}             & 38.48 {\scriptsize $\pm$1.42}             & 57.83 {\scriptsize $\pm$1.01}             & 42.89 {\scriptsize $\pm$0.50}             & 45.00 {\scriptsize $\pm$0.46}             \\
             & CPT          & 32.30 {\scriptsize $\pm$1.29}             & 52.09 {\scriptsize $\pm$0.50}             & 38.05 {\scriptsize $\pm$1.42}             & 58.08 {\scriptsize $\pm$1.01}             & \underline{45.21 {\scriptsize $\pm$0.50}} & 45.15 {\scriptsize $\pm$0.46}             \\
             & CPT + SLERP  & \underline{47.54 {\scriptsize $\pm$1.38}} & \underline{53.37 {\scriptsize $\pm$0.50}} & \underline{39.51 {\scriptsize $\pm$1.43}} & \underline{58.59 {\scriptsize $\pm$1.01}} & 44.83 {\scriptsize $\pm$0.50}             & \underline{48.77 {\scriptsize $\pm$0.47}} \\
        \midrule
        4B   & source       & 71.19 {\scriptsize $\pm$1.25}             & 64.07 {\scriptsize $\pm$0.48}             & 48.12 {\scriptsize $\pm$1.46}             & 66.75 {\scriptsize $\pm$0.97}             & 51.42 {\scriptsize $\pm$0.50}             & 60.31 {\scriptsize $\pm$0.45}             \\
             & CPT          & 57.92 {\scriptsize $\pm$1.36}             & 65.87 {\scriptsize $\pm$0.47}             & \underline{49.06 {\scriptsize $\pm$1.46}} & \underline{70.29 {\scriptsize $\pm$0.94}} & \underline{55.24 {\scriptsize $\pm$0.50}} & 59.68 {\scriptsize $\pm$0.46}             \\
             & CPT + SLERP  & \underline{72.18 {\scriptsize $\pm$1.23}} & \underline{66.04 {\scriptsize $\pm$0.47}} & 48.81 {\scriptsize $\pm$1.46}             & 69.70 {\scriptsize $\pm$0.94}             & 55.21 {\scriptsize $\pm$0.50}             & \underline{62.39 {\scriptsize $\pm$0.45}} \\
        \midrule
        8B   & source       & \textbf{78.77 {\scriptsize $\pm$1.13}}    & 69.98 {\scriptsize $\pm$0.46}             & 50.43 {\scriptsize $\pm$1.46}             & 72.81 {\scriptsize $\pm$0.91}             & 58.55 {\scriptsize $\pm$0.49}             & 66.11 {\scriptsize $\pm$0.43}             \\
             & CPT          & 71.95 {\scriptsize $\pm$1.24}             & 70.80 {\scriptsize $\pm$0.45}             & \textbf{53.24 {\scriptsize $\pm$1.46}}    & \textbf{75.04 {\scriptsize $\pm$0.89}}    & \textbf{61.38 {\scriptsize $\pm$0.49}}    & 66.48 {\scriptsize $\pm$0.44}             \\
             & CPT + SLERP  & 76.19 {\scriptsize $\pm$1.17}             & \textbf{71.00 {\scriptsize $\pm$0.45}}    & 52.56 {\scriptsize $\pm$1.46}             & 74.62 {\scriptsize $\pm$0.89}             & 61.26 {\scriptsize $\pm$0.49}             & \textbf{67.13 {\scriptsize $\pm$0.44}}    \\
        \bottomrule
    \end{tabular}
    \caption{General-domain benchmark performance of the \textit{Qwen3} family, evaluated $1$-shot on the German-translated benchmarks at the final checkpoint of the full adaptation pipeline (\textit{CPT} and \textit{CPT + SLERP}) and on the unadapted source model. Values are accuracy in \% with $\pm$ standard error. \textbf{Bold} marks the global maximum per column, and \underline{underlined} marks the maximum within a size group.}
    \label{tab:qwen3_de_results}
\end{table*}

\clearpage

\clearpage

\clearpage

\clearpage
\begin{figure*}[p]
    \centering
    \includegraphics[width=\linewidth]{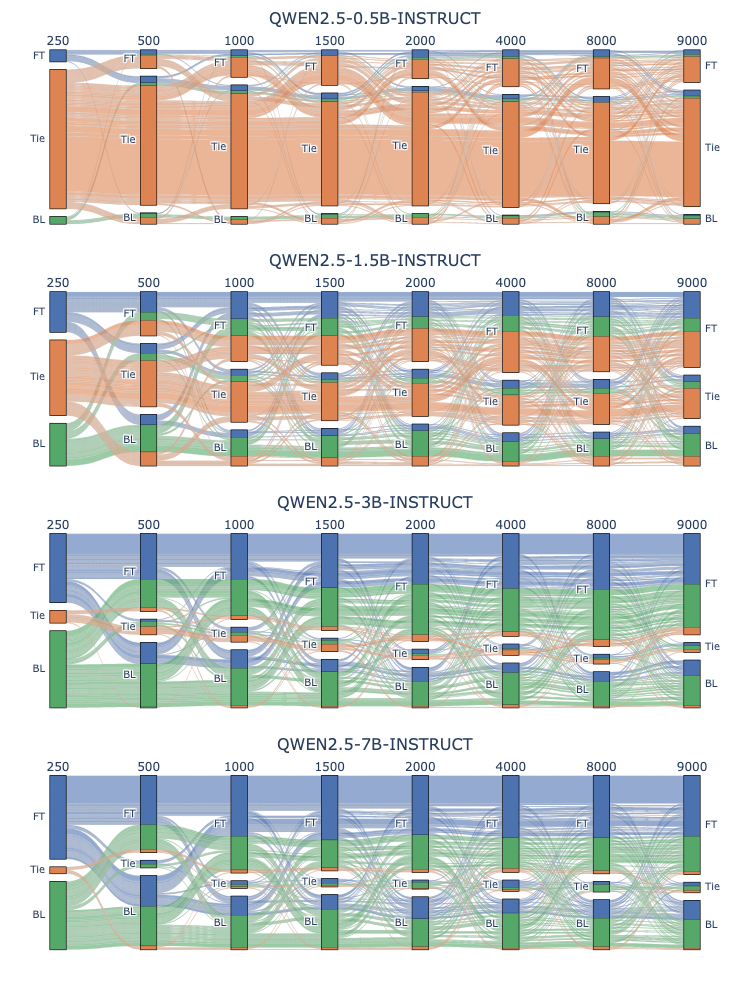}
    \caption{Parallel categories visualization of ``Win-Tie-Loss'' distributions for the \textit{Qwen2.5} family across different parameter scales. The plots track the flow of judge preferences between the source models (BL) and their domain-adapted counterparts (FT).}
    \label{fig:qwen2.5-win-rate-parcats}
\end{figure*}

\begin{figure*}[p]
    \centering
    \includegraphics[width=\linewidth]{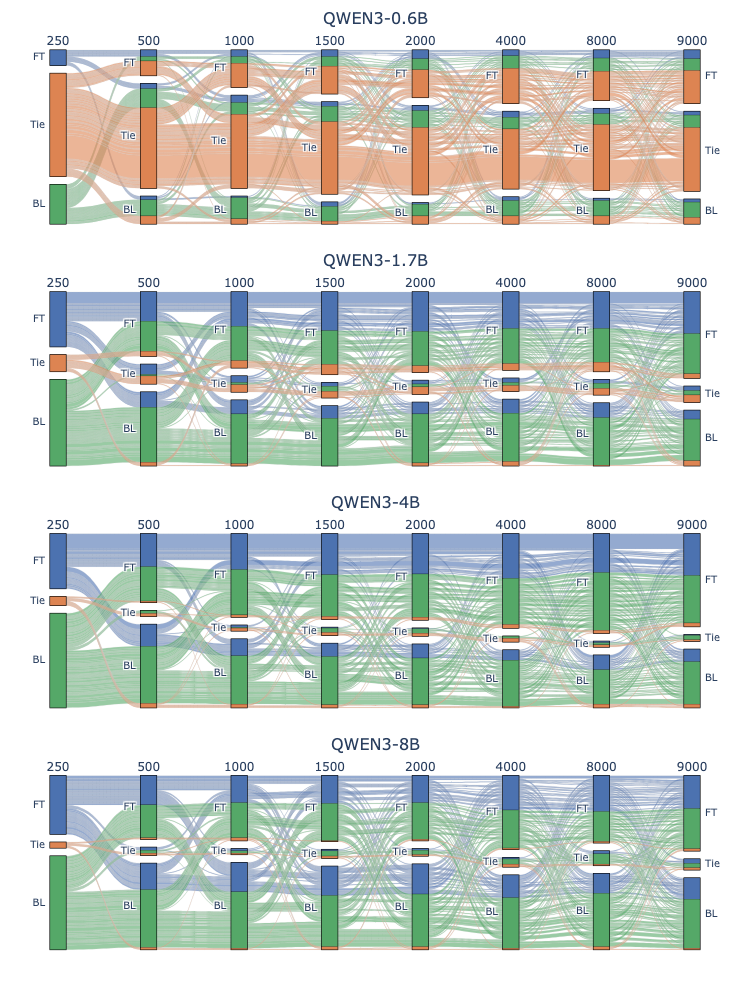}
    \caption{Parallel categories visualization of ``Win-Tie-Loss'' distributions for the \textit{Qwen3} family across different parameter scales. The plots track the flow of judge preferences between the source models (BL) and their domain-adapted counterparts (FT).}
    \label{fig:qwen3-win-rate-parcats}
\end{figure*}

\end{document}